\documentclass[lettersize,journal]{IEEEtran}
\usepackage{amsmath,amsfonts}
\usepackage{algorithmic}
\usepackage{algorithm}
\usepackage{array}
\usepackage[caption=false,font=normalsize,labelfont=sf,textfont=sf]{subfig}
\usepackage{textcomp}
\usepackage{stfloats}
\usepackage{url}
\usepackage{verbatim}
\usepackage{graphicx}
\usepackage{cite}

\usepackage{graphicx}

\usepackage{microtype}
\usepackage{graphicx}
\usepackage{subcaption}
\usepackage{booktabs} 
\usepackage{algorithm, algorithmic}
\usepackage{amsmath}
\usepackage{booktabs, makecell, multirow, array}
\usepackage{subcaption}
\usepackage{comment}
\usepackage{enumitem}
\usepackage{caption}
\usepackage{xcolor}
\usepackage{wrapfig}
\usepackage{hyperref}
\usepackage{multicol}
\usepackage{amsmath}
\usepackage{amssymb}
\usepackage{mathtools}
\usepackage{amsthm}
\usepackage{placeins}
\usepackage{tabularx}

\begin{document}

\title{Class Incremental Continual Learning with Self-Organizing Maps 
and Synthetic Replay}

\author{
Pujan~Thapa, Alexander~Ororbia, and Travis~Desell%
}



\maketitle

\begin{abstract}
This work introduces a novel generative continual learning framework based on self-organizing maps (SOMs), a brain-inspired natural computing model, extended with learned distributional statistics and encoder--decoder models for class incremental continual learning.  These extended SOMs enable exemplar-free replay with fixed-capacity statistical memory, eliminating the need to store raw data samples. For high-dimensional input spaces, the SOM operates over the latent space of the encoder--decoder, while for lower-dimensional inputs, the SOM operates in a standalone fashion. Our method stores a running mean, variance, and covariance for each SOM unit, from which synthetic samples are then generated during future learning iterations. For the encoder--decoder method, generated samples are fed through the decoder to be used in subsequent replay. Experimental results on standard class-incremental benchmarks show that our approach performs competitively with state-of-the-art memory-based methods and outperforms memory-free methods, notably improving over the best state-of-the-art single-class incremental performance without pretrained encoders on CIFAR-10 and CIFAR-100 by nearly $10$\% and $7$\%, respectively. We also achieve the best performance on single-class incremental CIFAR-100 using a foundational encoder-decoder and present the first baseline results for single-class incremental TinyImageNet. Our methodology facilitates easy visualization of the learning process and can also be utilized as a generative model post-training. Overall, the results demonstrate that SOM-based statistical replay offers a scalable, label-free, exemplar-free approach to class-incremental continual learning under standard task-boundary protocols.
\end{abstract}


\section{Introduction}
\label{sec:intro}

Computational systems deployed in real-world environments are often exposed to continuous streams of information, where the data distribution(s) that the system receives change over time. In such environments, machine learning models must adapt to new tasks sequentially, without revisiting previous data, while ensuring that they retain knowledge extracted from previous tasks. The ability of systems to learn new tasks while retaining knowledge of past experiences is referred to as continual learning (CL), or lifelong learning \cite{thrun1998lifelong}, and is central to building robust intelligent systems. 

The goal of continual learning is to integrate new information while preserving previously acquired knowledge. However, this is particularly challenging in neural networks because they have the tendency to overwrite old learned knowledge when trained on new data. This phenomenon is known as catastrophic forgetting \cite{cf_1, PARISI201954, ffe0793b43f842d2a50467d736a80c83} and has been the subject of extensive research in the field of machine learning. Catastrophic forgetting in artificial neural networks is rooted in the stability–plasticity dilemma ~\cite{abraham}, which means: a model needs plasticity to learn new information, but too much of it can overwrite older knowledge; meanwhile, if the model is too stable, it struggles to learn anything new. This trade-off between acquiring new knowledge and preserving prior learning remains a central challenge in the design of neural systems capable of robust, long-term, continuous learning.

Continual learning is typically categorized into three main scenarios: task-incremental learning (TIL), domain-incremental learning (DIL), and class incremental learning (CIL)~\cite{van2022three}. In TIL, the task identity is provided at inference, making it the easiest setting. DIL removes the task identity but keeps the same class set across domains. CIL is the most challenging, as the task identities are unknown and the model must discriminate among all classes seen so far while being exposed to only a subset of them at a time. The extreme case is single-class CIL, where data arrives one class at a time. Prior surveys on continual learning note that many benchmarks are defined in a multi-class-per-task fashion (e.g., Split-MNIST, Split-CIFAR), where each task introduces several new classes simultaneously \cite{survey1, survey2_ptm, survey3}. In contrast, this work focuses on the single-class-incremental learning setting, widely recognized as the most challenging protocol \cite{MALTONI201956}, since the model must incrementally separate classes without ever jointly observing them.

Zhou et al.\cite{10599804} presented a comprehensive survey that highlights CL approaches such as regularization, dynamic architectures, and memory-based strategies applied to vision, language, and reinforcement learning tasks. Strategies like elastic weight consolidation (EWC) ~\cite{schwarz2018progresscompressscalable, cf_1}, progressive neural networks ~\cite{rusu2022progressiveneuralnetworks}, and tree-CNNs ~\cite{roy2019treecnnhierarchicaldeepconvolutional} attempt to mitigate forgetting by freezing important weights, dynamically expanding architectures, or hierarchically partitioning data. More recent approaches extend these ideas: contrastive representation learning in Co$^2$L \cite{cha2021co2lcontrastivecontinuallearning}, rehearsal-based methods such as DER++ ~\cite{buzzega2020darkexperiencegeneralcontinual} and MEMO \cite{zhou2023model603exemplarsmemoryefficient}, and transformer-based expansions like DyTox \cite{douillard2022dytoxtransformerscontinuallearning}. While these approaches improve performance, they often require prior knowledge of task boundaries, access to labels, or an increased model size, which limits the scalability of the model.

The challenges of CL become particularly evident in the context of deep neural networks (DNNs). While DNNs have achieved remarkable success across vision, language, and reinforcement learning tasks~\cite{ Samek_2021, 8745428, ying2024enhancingdeepneuralnetwork}, they are highly susceptible to catastrophic forgetting~\cite{PARISI201954, ffe0793b43f842d2a50467d736a80c83, ororbia2022lifelong} when trained on sequential, non-i.i.d. data streams. Several DNN-based approaches have been developed in the field of CL~\cite{wang2024comprehensivesurveycontinuallearning}; however, most focus on supervised methods, which are often difficult to interpret due to model complexity. To better explore CL more clearly, some studies have turned to unsupervised methods ~\cite{9756660_uncl,madaan2022representationalcontinuityunsupervisedcontinual,10848884_uncl, ororbia2021continual}. These methods aim to learn evolving data distributions while maintaining previously acquired representations, often by leveraging latent space structure or clustering dynamics.

While extensive research has been done with DNNs in the area of CL~\cite{10599804}, self-organizing maps (SOMs) \cite{kohonen}, a class of unsupervised, topology-preserving neural models, have received relatively little attention despite their natural suitability for such settings. SOMs are a brain-inspired (natural computing) model that organizes data through topographic mapping. Additionally, SOMs are also well-suited for parallel implementations, as the computation of distances for Best Matching Unit (BMU) selection can be performed independently across units and their localized update mechanism reduces communication overhead, making them efficient for scalable deployment~\cite{pmlr-v95-liu18b, hsom_parallel, scalable_som}. SOMs consist of several unit vectors, where during training, only the best-matching unit and its neighbors are updated in response to input samples. This localized plasticity can help preserve previously learned representations and prevent forgetting, which has led to several studies of their use in CL \cite{ororbia2021continual}. Recent work has explored combining SOMs with neural architectures to improve continual learning -- SOMLP \cite{bashivan2019} utilizes a SOM layer to gate an MLP’s hidden units to reduce forgetting without requiring memory buffers or task labels. More recently, the dendritic SOM (DendSOM) \cite{pinitas2021} utilizes multiple localized SOMs to mimic dendritic processing, enabling sparse, task-specific learning. A related approach, the continual SOM (c-SOM)~\cite{hiteshcSOM}, introduces internal Gaussian replay in the input space, but the absence of a proper generative model limits its scalability and sample diversity.

Most continual learning methods rely on replay-based deep feature extractors, but usually do not explicitly organize the latent space. In our approach, the SOM acts as a layer that divides the latent space into localized regions. This allows (i) more memory-efficient replay by storing only simple statistics instead of full samples, (ii) localized updates that reduce interference between previously learned and new information, and (iii) more controlled sampling from specific regions rather than from a single global distribution. By introducing this structure, the SOM leads to more stable and interpretable behavior in class-incremental learning.

This paper proposes a novel family of unsupervised continual learning models that integrates extended Self-Organizing Maps (SOMs) with encoder--decoder architectures (for this work, convolutional VAEs \cite{kingma2013auto} and the foundational CLIP model \cite{radford2021learningtransferablevisualmodels}) to address catastrophic forgetting in class-incremental settings. The main contributions of this paper are: 
\begin{itemize}[noitemsep,nolistsep]
    \item A novel SOM-based CL framework that combines generative modeling with topology-preserving clustering for easily visualized unsupervised class-incremental learning.
    \item Three variants of the framework: 
    \textit{(i)} a SOM-only for low-dimensional data, 
    \textit{(ii)} a global encoder--decoder model jointly trained with the SOM to provide structured latent spaces, and  
    \textit{(iii)} per-BMU specialized encoder--decoder models for fine-grained generative replay.

    \item A generative replay scheme that stores only summary statistics (mean, variance, and covariance) per SOM unit, avoiding external buffers or replay data.

    \item Extensive evaluation in both non-pretrained and pretrained settings on MNIST, CIFAR-10, CIFAR-100, and TinyImageNet. Experiments demonstrate strong knowledge retention and scalability across increasing data complexity, showing state-of-the-art results in both single and multi-class incremental settings.
\end{itemize}

\section{Methodology}
\label{sec:methodology}

SOMs (also known as Kohonen maps)~\cite{kohonen} are a class of brain-inspired, or natural computing, models that mimic the topographic organization observed in biological neural systems. They offer a unique approach to unsupervised learning by mapping complex, multidimensional data into a two-dimensional grid. The strength of SOMs lies in their ability to capture the high-dimensional variance of data and represent it on a grid that is visually interpretable. SOMs contain a grid of unit vectors, and during training sample vectors are mapped to their best matching unit (BMU), which pulls the BMU and other unit vectors within a neighborhood radius towards the sample -- with unit vectors farther away from the BMU having smaller updates. Once the SOM is trained, an input vector can be assigned to its BMU, which serves as a representative anchor point for that input in the topological map. In the context of labeled datasets, each BMU can accumulate label distributions support downstream interpretability or weakly supervised clustering.

We propose a class-incremental continual learning framework that utilizes self-organizing maps (SOM) extended with learned per-unit distributions (running mean, variance and covariance), shown in Algorithm~\ref{alg:unified_all}. The SOM is trained with raw or embedding-level samples, and the per-unit summary statistics are utilized for generative replay -- either of raw data or by passing generated embeddings into a decoder -- allowing the SOM units to play a central role in learning by acting as generative memory units. This approach is capable of adapting to the complexity of the input data: simple grayscale images (e.g., MNIST) can be processed using raw data and the SOM alone, whereas higher-dimensional RGB images (e.g., CIFAR-10/100, TinyImageNet) can utilize embeddings from a larger scale model such as a variational autoencoder or foundation model for a more efficient and structured representation. Note this methodology is entirely unsupervised as class labels are not used to drive the weight updates in the SOM or embedding models. Class labels are tracked on BMU matches during training only so they can be used for testing the performance of using the SOM for inference. As labels are not required, this allows our methodology to be used for single class incremental learning, unlike many other continual learning models~\cite{survey1, wang2024comprehensivesurveycontinuallearning, survey3}. For evaluation, we use the final, trained SOM output as a classifier, where each unit is assigned a class label based on the most frequent label among the samples mapped to it during training, i.e., majority voting based on best matching unit (BMU) ``hit'' counts.

\subsection{Tracking SOM Unit Distribution Statistics}

For an $n \times n$ SOM grid and a momentum factor $\alpha$, each input $x$ is projected to its corresponding BMU, which is used to update the BMU's core properties: i), a running \textbf{mean vector}: $\mu_{ij} \leftarrow (1 - \alpha)\mu_{ij} + \alpha x$; ii), a running \textbf{variance vector} $\sigma^2_{ij}$, capturing per-dimension variability, calculated by: $\sigma^2_{ij} \leftarrow (1 - \alpha)\sigma^2_{ij} + \alpha (\mu_{ij} - x)^2$; and iii), a running \textbf{covariance matrix} $\Sigma_{ij}$, which helps in modeling inter-feature relationships, calculated by: $\Sigma_{ij} \leftarrow (1 - \alpha)\Sigma_{ij} + \alpha (x - \mu_{ij})(x - \mu_{ij})^\top$. Here, $x$ denotes a sample pattern vector when the SOM is standalone, otherwise, it is the latent vector encoding of the input, and is assigned to the BMU at position $(i,j)$ on the SOM grid. These running statistics characterize the local distribution of latent codes associated with each BMU. 

A practical issue with these running statistics is that they are biased toward their initialization in the early stages of training. For example, when updating the running mean of a BMU at location $(i,j)$ as $\mu_{ij,t} = \alpha x_t + (1-\alpha) \mu_{ij,t-1}$, when initialized as $\mu_{ij,0} = 0$, the estimate $\mu_{ij,t}$ underestimates the true mean since it is implicitly influenced by the zero initialization. This effect is especially problematic in continual learning settings, where some BMUs may receive very few samples early on. To mitigate this, we employ \emph{bias correction} in the style of the Adam optimizer~\cite{kingma2017adammethodstochasticoptimization}. Specifically, the corrected estimates for each BMU $(i,j)$ are:
\(
\begin{aligned}
\hat{\mu}_{ij,t} &= \frac{\mu_{ij,t}}{1 - \beta_\mu^t}, \quad
\hat{\sigma}^2_{ij,t} = \frac{\sigma^2_{ij,t}}{1 - \beta_\sigma^t}, \quad
\hat{\Sigma}_{ij,t} = \frac{\Sigma_{ij,t}}{1 - \beta_\Sigma^t}
\end{aligned}
\)
where $t$ denotes the number of update steps (BMU matches) received by the BMU, and $\beta_\mu, \beta_\sigma, \beta_\Sigma \in [0,1)$ are the statistic specific exponential decay rates. This bias correction ensures that BMU parameters $(\mu_{ij}, \sigma^2_{ij}, \Sigma_{ij})$ remain unbiased from the start, yielding stable local distribution estimates and well-conditioned covariances for synthetic sampling. The neighboring units are updated with decayed learning rates based on their distance to the BMU, preserving SOM topology. We utilize this methodology as opposed to an unbiased online estimator such as Welford's Algorithm as the BMU statistical distributions can change over time as new samples are mapped to the SOM.

\begin{figure*}[t]
    \centering
    \begin{minipage}{0.26\textwidth}
        \centering
        \includegraphics[width=\linewidth]{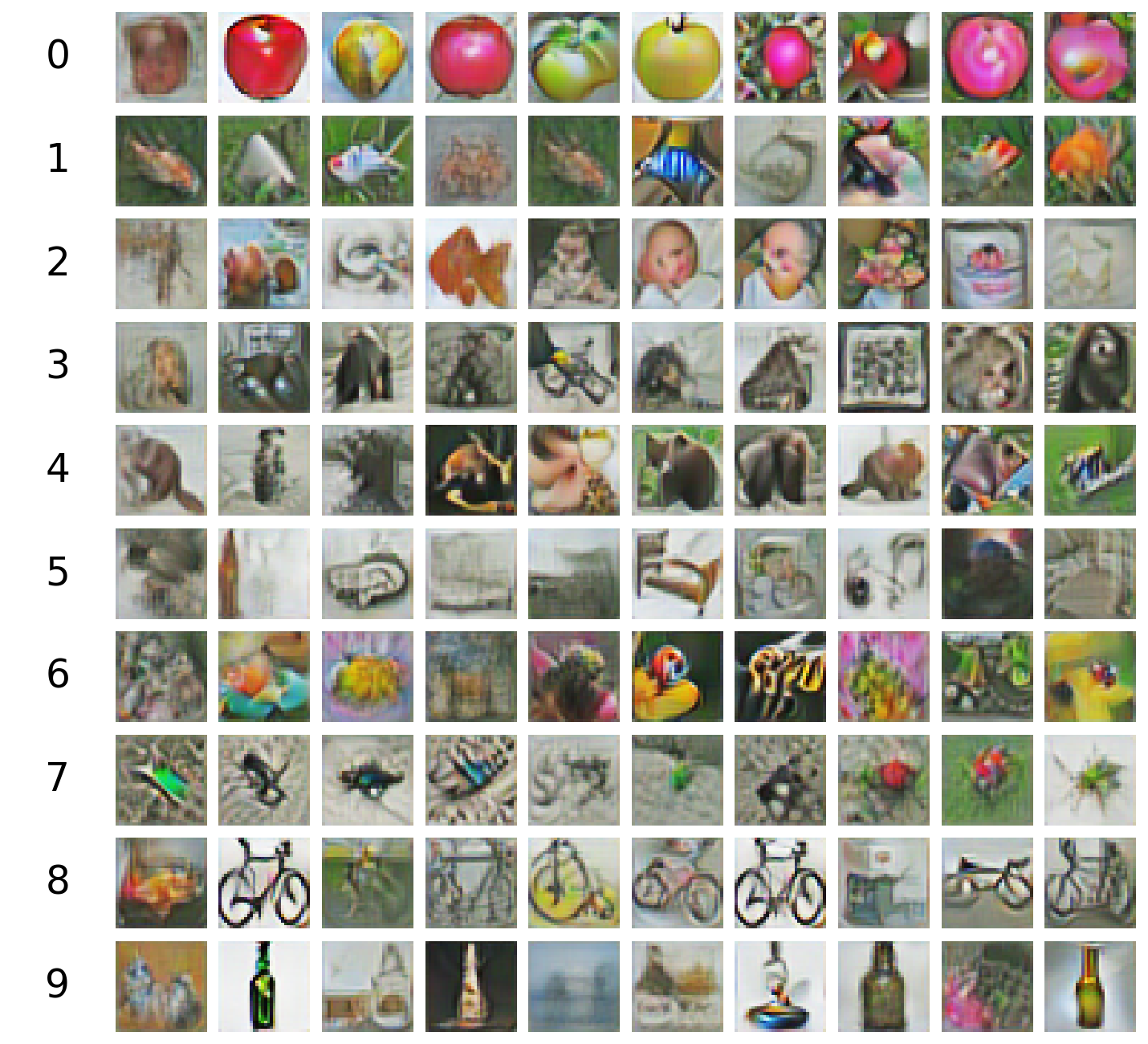}
    \end{minipage}
    \hspace{0.02\textwidth}
    \begin{minipage}{0.26\textwidth}
        \centering
        \includegraphics[width=\linewidth]{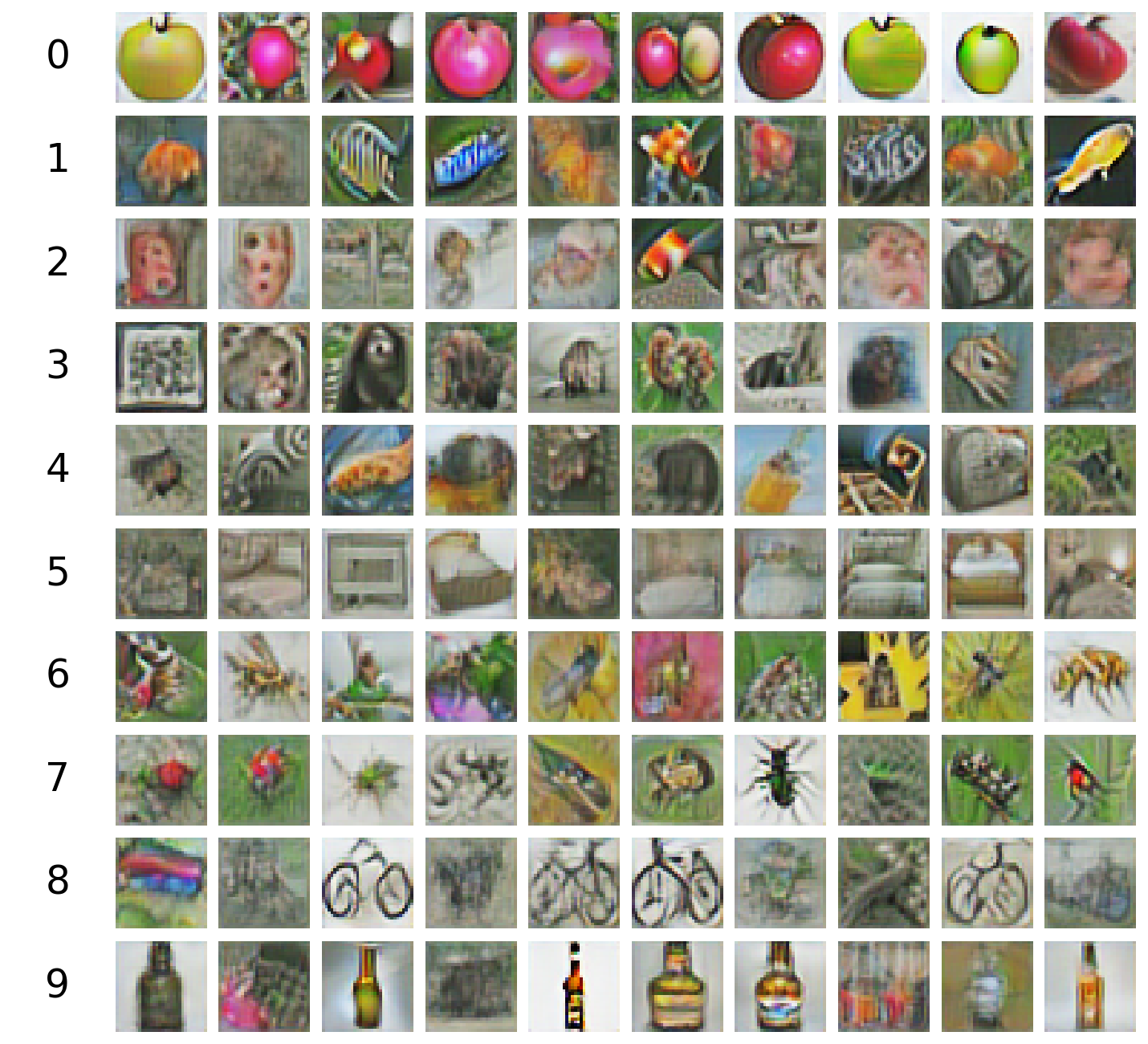}
    \end{minipage}
    \caption{
        Synthetic CIFAR-100 samples for the first $10$ classes generated by sampling latent vectors from an SOM trained on VAE latent space (left) and CLIP embeddings (right). Each row corresponds to one class ($0$–$9$).
    }
    \label{fig:som_synthetic_cifar100}
\end{figure*}

\subsection{Continual Learning and SOM Synthetic Replay}

To utilize this method for class incremental learning,  our model is trained sequentially as new classes arrive. At task $t$, only the data from class $C_t$ is available. To retain knowledge from previous classes $C_0, C_1, \dots, C_{t-1}$, our method generates synthetic samples using these distributional statistics, and replays them alongside new class data. For multi-class incremental learning, more than one class can be provided per task. The unit distribution statistics allow the SOM to serve as a class- and region-specific memory that accumulates information over time, mitigating forgetting.  When a new task $t$ consisting of a disjoint subset of classes is introduced, synthetic data from previous classes is generated by sampling from the stored Gaussian distributions of their corresponding BMUs:
\(
\begin{aligned}
\tilde{x} &\sim \mathcal{N}(\hat{\mu}_{ij}, \hat{\sigma}^2_{ij})
\quad \text{(MNIST)} \\
\tilde{z} &\sim \mathcal{N}(\hat{\mu}_{ij}, \hat{\Sigma}_{ij}) \quad \text{(CIFAR-10 / CIFAR-100 / TinyImageNet)} \\
\end{aligned}
\)

It should be noted that when using the covariance method to generate the samples from the distribution, we need to apply eigenvalue regularization to ensure numerically stable sampling (Algorithm~\ref{alg:cifar10-synth}, Appendix~\ref{sec:experiment_settings}). This procedure eliminates negative or near-zero variance directions that may otherwise lead to instability during sampling and ensures that the multivariate normal distribution remains valid and well-conditioned.

While we use boundaries to organize the training phases (for experimental simulation) and trigger the generation of synthetic samples from previously seen classes, the model itself is trained without access to task IDs or class labels. Replay is scheduled externally, but the CLIP, VAE and SOM modules update solely based on input data, making no distinction between real and replayed samples. This places our method in a more challenging CL context, with similarities to task-free CL \cite{Lee2020A,ororbia2021continual}, where explicit supervision about task transitions is not available during model updates. While the learning dynamics are label-free, this work assumes access to task boundary information for replay scheduling; therefore, it should not be considered fully task-free continual learning, and the overall framework remains task-aware.

\begin{algorithm*}[tb]
\scriptsize
\caption{Unified Algorithm for Class-Incremental Learning with SOM and Optional Encoder/Decoder Models (e.g., VAE, CLIP)}
\label{alg:unified_all}
\textbf{Input}: Dataset $\mathcal{D}=\{\mathcal{D}_c\}_{c=0}^{C-1}$; flags \texttt{GLOBAL}, \texttt{BMU\_HEADS}; replay samples per BMU $K$\\
\textbf{Output}: SOM, (optional) global encoder/decoder, (optional) BMU-specific final heads $\mathcal{H}$
\begin{multicols}{2}
\begin{algorithmic}[1]
\STATE Initialize SOM; initialize replay buffer $\mathcal{R}\leftarrow \emptyset$
\STATE \textbf{if} \texttt{GLOBAL} \textbf{then} initialize shared encoder/decoder trunk and global heads
\STATE \textbf{if} \texttt{BMU\_HEADS} \textbf{then} initialize BMU-head dictionary $\mathcal{H}\leftarrow \emptyset$

\FOR{$c=0$ to $C{-}1$}
   \STATE $\mathcal{T}_c \leftarrow \mathcal{D}_c$ if $c{=}0$ else $\mathcal{D}_c \cup \mathcal{R}$

  \vspace{0.25em}
  \STATE \textbf{// Encode features for SOM update}
  \IF{\texttt{GLOBAL}}
    \STATE Train/update shared encoder/decoder on $\mathcal{T}_c$
    \STATE $\mathcal{Z}_c \leftarrow \text{Enc}_{\text{global}}(\mathcal{T}_c)$
    \STATE Update SOM with $\mathcal{Z}_c$ (including neighborhood updates)
  \ELSE
    \STATE Update SOM with $\mathcal{T}_c$ (including neighborhood updates)
  \ENDIF

  \vspace{0.25em}
  \STATE \textbf{// Update BMU-specific heads after routing}
\IF{\texttt{BMU\_HEADS}}
  \STATE Freeze $E_{\theta}^{\mathrm{trunk}}$ and $D_{\phi}^{\mathrm{trunk}}$
  \STATE Assign each $x \in \mathcal{T}_c$ to BMU $b=(i,j)$ using shared trunk; collect subsets $\{\mathcal{S}_b\}$
  \FOR{each active BMU $b$ with assigned subset $\mathcal{S}_b$}
    \STATE $\mathcal{H}_b \leftarrow \textsc{GetHeads}(\mathcal{H}, b)$
    \STATE $h_b \leftarrow E_{\theta}^{\mathrm{trunk}}(\mathcal{S}_b)$
    \STATE $z_b \leftarrow q_b(h_b;\mathcal{H}_b)$ \COMMENT{BMU-specific latent/projection head}
    \STATE $u_b \leftarrow D_{\phi}^{\mathrm{trunk}}(z_b)$
    \STATE $\hat{x}_b \leftarrow o_b(u_b;\mathcal{H}_b)$ \COMMENT{BMU-specific output head}
    \STATE $\mathcal{L}_b \leftarrow \mathcal{L}(\hat{x}_b, \mathcal{S}_b)$
    \STATE Update only $\mathcal{H}_b$ using $\nabla_{\mathcal{H}_b}\mathcal{L}_b$
  \ENDFOR
\ENDIF

  \vspace{0.25em}
  \STATE \textbf{// Build replay for next class}
  \STATE $\mathcal{R}_c \leftarrow \emptyset$
  \FOR{each BMU $(i,j)$}
    \STATE Obtain BMU stats $(\mu_{ij}, \Sigma_{ij})$ from running latent/feature stats
    \STATE $(\hat{\mu}_{ij}, \hat{\sigma}^2_{ij}, \hat{\Sigma}_{ij}) \leftarrow$
    \STATE\textsc{BiasCorrect}$(\mu_{ij}, \sigma^2_{ij}, \Sigma_{ij}, \beta, t_{ij}, \lambda)$

    \FOR{$k=1$ to $K$}
      \STATE $\tilde{z}\!\sim\!\mathcal{N}(\hat\mu_{ij},\hat\Sigma_{ij})$
      \STATE $\tilde{x}\!\leftarrow\!\begin{cases}
      \text{Dec}_{g}^{\text{trunk}}(\tilde{z};\mathrm{head}_{ij}), 
      & \texttt{BMU\_HEADS}\ \&\ (i,j)\!\in\!\mathcal{H}\\
      \text{Dec}_{g}(\tilde{z}), & \texttt{GLOBAL}\\
      \tilde{z}, & \text{SOM-only}
      \end{cases}$
      \STATE Append $\tilde{x}$ to $\mathcal{R}_c$
    \ENDFOR
  \ENDFOR

  \STATE Set $\mathcal{R}\leftarrow \mathcal{R}_c$
\ENDFOR
\STATE \textbf{return} SOM, (global encoder/decoder if used), (BMU heads $\mathcal{H}$ if used)
\end{algorithmic}
\end{multicols}
\end{algorithm*}

Note that, unlike classical buffer-based replay, our SOM system design leverages only these statistical summaries stored at each BMU. This enables highly compact memory usage as it scales with respect to the fixed SOM grid size as opposed to the dataset size. Figure~\ref{fig:som_synthetic_cifar100} shows synthetic samples generated from an SOM fully trained, class-incrementally, on CIFAR-100, where each latent vector was sampled from the full-covariance Gaussian distribution of a
BMU associated with the target class and subsequently decoded by the VAE.
For experiments that use CLIP embeddings, we employ a decoder trained on CLIP
embeddings to reconstruct images from the sampled latent vectors.  Appendix~\ref{app:generative_model} provides similar visualizations of images generated from the SOM unit vectors for other datasets. Over all datasets, we find that the SOM systems generate feasible variations of their learned class.

Our methodology also provides interpretability via the SOM grid's topological visualization of class structure and supports modularity, since synthetic replay depends solely on local BMU statistics. This makes it possible to easily view the progress of the SOM by decoding unit vectors after each task to visually see how the model is performing (see Appendix~\ref{app:som_representations}). If the method is performing well, classes have well formed clusters within the SOM.



\subsection{High-Dimensional Replay Challenges and the Role of Latent Compression}

For low-dimensional image datasets such as MNIST, synthetic sample generation using SOMs is generally efficient and effective due to the low dimensionality of the feature space ($28 \times 28$ grayscale pixels), where the variance along each independent dimension is sufficient to model the data distribution. However, this mean-variance sampling strategy is insufficient when applied to high-dimensional datasets, e.g., CIFAR-10 and CIFAR-100; and with preliminary results for sample generation using the mean-variance method above were of poor quality and degraded classification performance. Additionally, using the mean covariance method in pixel space introduces memory storage issues, e.g., CIFAR-10/100 requires storing a full $3072 \times 3072$ covariance matrix for each BMU.

To address this, we allow the use of embeddings from any generative model, which serve to compress high-dimensional images into a compact latent space. In this work, we investigate two instantiations: a non-pretrained convolutional VAE~\cite{kingma2013auto,Kingma_2019} and a pretrained foundation model, CLIP (ViT-B/32)~\cite{radford2021learningtransferablevisualmodels}. This significantly reduces the dimensionality of each SOM BMU; for example, the CLIP encoder produces a $512$-dimensional embedding, while our VAE configuration uses a $d$-dimensional latent space (e.g., $d=128$). Such compression makes it feasible to model and store full covariance matrices. By sampling from the full multivariate Gaussian $\mathcal{N}(\mu, \Sigma)$ in this latent space, we generate high-quality synthetic representations by feeding these synthetic samples through the models decoder, replaying them in subsequent training phases. This hybrid approach ensures scalable, efficient, and expressive sample replay for complex image distributions in CL settings. It is also flexible in that different encoder/decoder models can be substituted depending on the dataset and experimental requirements. This setup emphasizes that our contribution lies not in the specific encoder--decoder, but in the replay framework itself. Whether the latent codes or embeddings originate from a VAE or a large pretrained transformer, the SOM provides the same modular structure for clustering, exemplar-free replay with fixed-capacity statistical memory, and continual adaptation.

\textbf{VAE Instantiation.} The VAE encoder maps inputs into a structured latent distribution parameterized by mean $\mu$ and log-variance $\log \sigma^2$, with sampling performed via the reparameterization trick: $z = \mu + \sigma \odot \epsilon,\ \epsilon \sim \mathcal{N}(0,I)$. The VAE decoder reconstructs the image from $z$, enabling generative replay by reconstructing synthetic samples drawn from SOM statistics (means and covariances). The encoder and decoder are implemented using residual downsampling and upsampling blocks~\cite{he2016deep}, providing stable and expressive nonlinear transformations. Depending on the configuration, the VAE compresses images into latent vectors of dimension $d$ (e.g., $d=128$), achieving over 90\% dimensionality reduction compared to raw pixel space. 

\noindent 
\textbf{CLIP Instantiation.} To demonstrate the generality of the framework, we also evaluate our replay strategy using CLIP (ViT-B/32)~\cite{radford2021learningtransferablevisualmodels}, a large vision--language foundation model. Instead of training an encoder from scratch, we extract $512$-dimensional embeddings from CLIP’s penultimate transformer block. The CLIP encoder is either kept frozen or fine-tuned, depending on the experimental setting. Once the embeddings are extracted, they are fed directly into the SOM, and the replay mechanism is identical to the VAE case, where the SOM maintains class-specific statistics over the embeddings and uses them to generate synthetic embeddings. These sampled embeddings are then passed through a decoder trained on CLIP embeddings to reconstruct images for replay. The full decoder architecture used for CLIP-based reconstruction is provided in 
Appendix \ref{app:enc_dec_arch}.

The encoder-decoder framework, as described above, offers several advantages in CL by reducing input dimensionality, facilitating efficient memory usage for SOM BMU statistics, and enabling fast and realistic synthetic replay for SOMs. It acts as a front-end compression module, transforming high-dimensional images into a structured, low-dimensional latent space. This ultimately makes the overall SOM-based system training faster and allows for reliable Gaussian-based sample generation to induce replay.

\subsection{BMU-Specific Terminal-Head Specialization}

As an extension to the global encoder--decoder framework, we introduce a lightweight specialization strategy in which the encoder and decoder trunks are shared across all SOM units, and only the terminal heads are BMU-specific. This is equivalent to performing a lightweight fine-tuning per BMU: the shared trunk acts as a frozen backbone, and each BMU maintains its own locally fine-tuned terminal heads trained on the samples routed to it.

Each input is first embedded by the shared encoder trunk, and BMU assignment is performed in this common representation space. Once a sample is assigned to BMU $b=(i,j)$, the shared trunk representation $h$ is passed through a BMU-specific encoding head $q_b$ to produce the local representation $z_b = q_b(h)$. For VAE-based models, $q_b$ outputs local mean and log-variance parameters $(\mu_b, \log\sigma_b^2)$. For deterministic encoders such as CLIP or ResNet, $q_b$ is a lightweight linear projection $W_b h + a_b$. On the decoder side, $z_b$ is passed through the shared decoder trunk followed by a BMU-specific output head $o_b$. During training, the shared trunks are frozen and only the terminal heads $(q_b, o_b)$ are updated.

This design is motivated by two practical concerns. First, \textbf{scalability}: a full per-BMU encoder--decoder would require memory $O(|\mathcal{B}|P_{\mathrm{model}})$, which is prohibitive for large grids or pretrained backbones such as CLIP. The terminal-head variant reduces this to $O(P_{\mathrm{shared}} + |\mathcal{B}|P_{\mathrm{head}})$, with $P_{\mathrm{head}} \ll P_{\mathrm{model}}$. Second, \textbf{data sparsity}: each BMU receives only a fraction of the training data, which is insufficient to train a full encoder from scratch. By sharing the trunk, the encoder sees all samples; only the lightweight head is trained on the sparse per-BMU subset, making local specialization tractable.

\section{Results}
\label{sec:results}

Our methodology was evaluated across widely-adopted CL benchmarks: MNIST, CIFAR-10, CIFAR-100 and TinyImageNet, first in the more challenging single class incremental setting, where classes are presented one at a time (that is, class 0, then class 1, etc.). We also evaluated our methodology on standard split versions of these datasets, set up in a task-incremental fashion, where each task contains $N$ disjoint classes, e.g., for a task size of two the first task would have classes 0-1, the second 2-3, etc. All results are averaged over 20 random task orderings to reduce ordering bias, and the reported accuracies correspond to the mean performance across these runs. Results utilize the best found training and initialization hyperparameters, which were taken after significant ablation studies (see Appendices~\ref{sec:experiment_settings} and~\ref{sec:app_som_vae_dimensions}).

\begin{table*}[t]
\centering
\scriptsize
\setlength{\tabcolsep}{2pt}   
\renewcommand{\arraystretch}{0.9}
\caption{Final classification accuracy for single class incremental learning.}
\label{tab:cil_pretrained_and_single}
\begin{tabular}{lcccccc}
\toprule
\textbf{Method} 
  & \multicolumn{3}{c}{\textbf{w/o pretraining}} 
  & \multicolumn{2}{c}{\textbf{w/ pretraining}}
  & \textbf{TinyImageNet}\\
\cmidrule(lr){2-4} \cmidrule(lr){5-6}
  & \textbf{MNIST} & \textbf{CIFAR10} & \textbf{CIFAR100} 
  & \textbf{CIFAR10} & \textbf{CIFAR100} & \\
\midrule
EWC       & 9.91  & 10.01 & 1.03  & 10.21 & 2.93 & --\\
LwF       & 19.96 & 10.05 & 2.13  & 19.39 & 6.25 & --\\
IMM       & 29.16 & 10.25 & 1.21  & 51.22 & 12.58 & --\\
PGMA      & 71.36 & 20.08 & 1.86  & 56.22 & 12.37 & --\\
RPSNet    & 40.29 & 16.31 & 1.96  & 55.54 & 4.13 & --\\
OWM       & 94.46 & 19.63 & 3.67  & 83.03 & 63.26 & --\\
PCL       & 95.75 & 31.58 & 5.58  & \textbf{84.93} & 63.61 & --\\
DisCOIL   & \textbf{96.69} & 44.54 & --    & --    & -- & --\\
PCL-L2    & --    & --    & --    & 77.95 & 54.83 & --\\
\midrule
\multicolumn{7}{l}{\textbf{Ours (SOM-based w/o bias)}} \\
\midrule
SOM only & 94.29 & -- & -- & -- & -- & -- \\
VAE (FT global) & 92.38 & 53.74 & 11.98 & -- & -- & 6.92\\
VAE (BMU Specific) & 91.96 & 45.52 & 11.63 & -- & -- & 6.12\\
CLIP (global) & -- & -- & -- & 77.36 & 60.71 & 40.21 \\
CLIP (FT global) & -- & -- & -- & 80.41 & 62.44 & 40.89 \\
CLIP (BMU Specific) & -- & -- & -- & 82.37 & \textbf{63.02} & 42.48 \\
\midrule
\multicolumn{7}{l}{\textbf{Ours (SOM-based w bias)}} \\
\midrule
SOM only & 94.96 & -- & -- & -- & -- & -- \\
VAE (FT global) & 92.41 & \textbf{53.92} & \textbf{12.08} & -- & -- & 7.02\\
VAE (BMU Specific) & 92.05 & 48.51 & 11.69 & -- & -- & 6.24\\
CLIP (global) & -- & -- & -- & 78.31 & 61.88 & 41.72 \\
CLIP (FT global) & -- & -- & -- & 80.55 & 62.54 & 42.61 \\
CLIP (BMU Specific) & -- & -- & -- & 82.74 & \textbf{63.98} & \textbf{45.11} \\
\bottomrule
\end{tabular}
\end{table*}

\begin{table*}[t]
\centering
\scriptsize
\setlength{\tabcolsep}{3.4pt}   
\renewcommand{\arraystretch}{0.9}
\caption{Final classification accuracy for multi class-incremental learning, non-pretrained models.}
\label{tab:cil_results_benchmark}
\begin{tabular}{lp{1.8cm}p{1.8cm}p{1.8cm}p{1.8cm}}
\toprule
\textbf{Method} & \textbf{MNIST} & \textbf{CIFAR-10} & \textbf{CIFAR-100}  & \textbf{TinyImageNet}  \\
& (5 tasks) & (5 tasks) & (10 tasks) & (10 tasks) \\
\midrule
\multicolumn{5}{l}{\textbf{Baseline}} \\
iid-offline   & 95.82 $\pm$ 0.33 & 80.54 $\pm$ 0.63 & 48.09 $\pm$ 0.90 & 59.99 $\pm$ 0.34\\
Fine-Tune     & 19.68 $\pm$ 0.02 & 19.19 $\pm$ 0.06 & 8.32 $\pm$ 0.23  & 7.92 $\pm$ 0.05 \\
\midrule
\multicolumn{5}{l}{\textbf{Memory-free}} \\
EWC           & 19.92 $\pm$ 0.35 & 16.18 $\pm$ 1.37 & 4.41 $\pm$ 0.37  & 7.58 $\pm$ 0.76 \\
SI            & 19.76 $\pm$ 0.01 & 17.27 $\pm$ 0.87 & 5.87 $\pm$ 0.21  & 6.58 $\pm$ 0.14 \\
LwF           & 20.54 $\pm$ 0.64 & 18.53 $\pm$ 0.12 & 6.93 $\pm$ 0.32  & 8.46 $\pm$ 0.46\\
\midrule
\multicolumn{5}{l}{\textbf{Memory-based}} \\
GEM           & 48.57 $\pm$ 5.26 & 25.54 $\pm$ 0.19 & 6.18 $\pm$ 0.20  & -- \\
iCaRL         & 72.55 $\pm$ 0.45 & 35.88 $\pm$ 1.43 & 15.76 $\pm$ 0.15 & 7.53 $\pm$ 0.79 \\
GSS           & 54.14 $\pm$ 4.68 & 49.22 $\pm$ 1.71 & 11.33 $\pm$ 0.40 & -- \\
ER-MIR        & 86.60 $\pm$ 1.60 & 37.80 $\pm$ 1.80 & 9.20 $\pm$ 0.40  & -- \\
CN-DPM        & \textbf{93.81 $\pm$ 0.07} & 47.05 $\pm$ 0.62 & 16.13 $\pm$ 0.14 & -- \\
DER++         & 92.21 $\pm$ 0.54 & 52.01 $\pm$ 3.06 & 15.04 $\pm$ 1.04 & 10.96 $\pm$ 0.17 \\
ER-ACE        & 82.98 $\pm$ 1.79 & 35.16 $\pm$ 1.34 & 8.92 $\pm$ 0.25  & 12.11 $\pm$ 0.06 \\
\midrule
\multicolumn{5}{l}{\textbf{Biologically Inspired}} \\
NNA-CIL (INEL+MNIST)   & 77.25 $\pm$ 1.02 & 45.95 $\pm$ 0.90 & \textbf{25.56 $\pm$ 0.69} & -- \\
\midrule
\multicolumn{5}{l}{\textbf{Our Method (w/o bias)}} \\
SOM-only & 91.86 $\pm$ 1.02 & -- & -- & -- \\
VAE (FT global) & 91.02 $\pm$ 1.18 & 52.63 $\pm$ 1.07 & 14.41 $\pm$ 0.19 & 12.61 $\pm$ 0.71 \\
VAE (BMU specific) & 89.54 $\pm$ 1.42 & 45.92 $\pm$ 1.26 & 13.04 $\pm$ 0.29 & 12.03 $\pm$ 0.28 \\
\midrule
\multicolumn{5}{l}{\textbf{Our Method (w bias)}} \\
SOM-only & 92.11 $\pm$ 1.18 & -- & -- & -- \\
VAE (FT global) & 91.23 $\pm$ 1.21 & \textbf{53.01 $\pm$ 1.18} & 14.63 $\pm$ 0.33 & \textbf{12.74 $\pm$ 0.84} \\
VAE (BMU specific) & 90.48 $\pm$ 1.44 & 46.58 $\pm$ 1.21 & 13.39 $\pm$ 0.41 & 12.15 $\pm$ 0.24 \\
\bottomrule
\end{tabular}
\end{table*}

\subsection{One Class Per Task Incremental Learning}

Table~\ref{tab:cil_pretrained_and_single} compares our methodology with several well-known CL methods for the single class incremental setting (MNIST and CIFAR 10 have 10 tasks, CIFAR-100 has 100 tasks, and TinyImageNet has 200 tasks; one class per task for each). We compare pretrained and non-pretrained methodologies, bias and non-bias correction, and our three methods for incorporating generative models -- a global fixed model (frozen, global), a global model that is trained concurrently with the SOM (FT global), and models concurrently trained for each SOM unit (FT BMU specific). For MNIST, our SOM-only approach achieves an accuracy of $94.29$\% without the bias correction, closely matching the best performing rehearsal-based methods, including DisCOIL ($96.69$\%) and PCL ($95.75$\%). With bias correction, the accuracy increased to $94.96$\%. Traditional regularization-based approaches such as EWC and LwF perform poorly on CIFAR-10 ($10.01$\%, $10.05$\%) and CIFAR-100 ($1.03$\%, $2.13$\%), indicating significant forgetting in the one-class stream. More advanced strategies like PGMA~\cite{pgma}, RPSNet~\cite{rpsnet}, and OWM~\cite{owm} achieve lower gains, with CIFAR-100 scores ranging from $1.86$\% to $3.67$\%. The memory-based approaches, such as PCL~\cite{pcl} and DisCOIL~\cite{discoil}, perform significantly better (up to $44.54$\% in CIFAR-10), although they often rely on external memory/labels. In contrast, our method (with bias correction) achieves $53.92$\% in CIFAR-10 and $12.08$\% in CIFAR-100 without access to external exemplars or task identifiers, outperforming most baselines by large margins. Notably, the VAE (FT BMU specific) variant surpasses previous methods on CIFAR-100 ($11.98$\% without bias correction; $12.08$\% with bias correction), highlighting the effectiveness of our generative replay strategy based on structured topological organization. These results also highlight an important distinction between different families of continual learning methods. Regularization-based approaches (e.g., EWC, SI, and LwF) attempt to preserve previous knowledge only through parameter constraints, which becomes increasingly difficult when each task contains only a single class. In contrast, replay-based methods explicitly revisit previously observed distributions. Our framework follows this principle but differs from conventional rehearsal methods by reconstructing replay samples directly from the learned SOM representation rather than storing original training examples. Consequently, the proposed approach achieves competitive or superior performance without requiring an external exemplar memory or task labels during replay.

With CLIP embeddings, SOM replay again delivers strong gains: frozen global features achieve $78.31$\% on CIFAR-10, while fine-tuned global reaches $80.55$\% and BMU-specific fine-tuning achieves $82.74$\%, with bias correction. On CIFAR-100, the BMU-specific configuration reaches $63.02$\%, well above rehearsal-based PCL ($54.83$\%) and other baselines, even without bias correction. With bias correction, the accuracy increases to $63.98$\%. This demonstrates that while per-BMU replay struggles with non-pretrained VAEs due to limited data, it is highly effective when paired with pretrained encoders like CLIP. The performance difference between the VAE and CLIP variants also provides insight into the behavior of the proposed framework. When training a VAE from scratch, replay quality is limited by the amount of data available to each BMU, particularly for high-dimensional datasets such as CIFAR-100 and TinyImageNet. In contrast, CLIP already provides a well-structured feature space learned from large-scale image-text supervision. Under these conditions, the SOM primarily models feature organization rather than representation learning, allowing localized BMU-specific adaptation to generate substantially higher-quality replay samples. This explains why BMU-specific method becomes increasingly beneficial when pretrained representations are available.

\textbf{Benchmark: Single Class Incremental TinyImageNet.} To the best of our knowledge, no prior work reports single class-incremental results on TinyImageNet. Existing CIL studies that include TinyImageNet use multiple classes per task (e.g., 2, 5, or 10), as shown in Table~\ref{tab:cil_results_benchmark}. We observe that training VAEs with SOM (bias-corrected) from scratch on TinyImageNet in the single class incremental setting yields very low accuracies (6.92\% for the global model and 6.12\% for the per-BMU variant), similar to CIFAR-100. This reflects the difficulty of scaling generative replay on complex datasets from limited data when both the encoder and decoder must be learned jointly from scratch. In contrast, leveraging pretrained CLIP embeddings with our SOM methodology substantially boosts performance. The frozen global variant without bias correction achieves 40.21\%, and fine-tuning further improves results to 40.89\% (w/o bias). The per-BMU with bias correction configuration achieves the best score
(45.11\%, with bias), highlighting that localized adaptation on top of a strong pretrained backbone can provide effective replay signals even under the strict one-class stream. All reported numbers are averaged over 20 independent reordered runs to account for variability. To our knowledge, these results are the first presented for single class incremental TinyImageNet.

\subsection{Standard Multi Class Incremental Benchmarks}
To further highlight the applicability of our methodology, we evaluate it on multi-class incremental methods on split versions of the benchmark datasets. Tables \ref{tab:cil_results_benchmark} and ~\ref{tab:cil_pretrained_split} present classification accuracy (mean $\pm$ standard deviation) over 20 independent reordered runs, comparing our method with baseline, memory-free, memory buffer, and biologically-inspired CL approaches for non-pretrained and pretrained methodologies. MNIST and CIFAR-10 have 2 classes per task, CIFAR-100 has 10 classes per task, and TinyImageNet has 20 classes per task. The IID-offline model represents an upper reference rather than a continual learning baseline, since it is trained jointly on all classes simultaneously with unrestricted access to the complete dataset. Confusion matrices for each data set are shown in the Appendix~\ref{app:confusion_matrices}, highlighting that our methodology retains accuracy across all tasks. Unlike the single-class setting, each incremental task now contains multiple semantically related classes, providing richer information for both representation learning and replay generation. This setting, therefore, allows us to examine whether the proposed SOM-based replay framework remains competitive under more conventional continual learning benchmarks while maintaining the advantage of not storing real replay exemplars.

On Split-MNIST, our method achieves $91.86 \pm 1.02$ \% when using SOM-based replay with Gaussian sampling without bias correction. This outperforms all memory-free approaches (i.e., EWC: $19.92$\% ~\cite{cf_1}, SI: $19.76$\% ~\cite{zenke2017continuallearningsynapticintelligence}, LwF: $20.54$\% ~\cite{8107520}), bio-inspired NNA-CIL ($77.25$\%) ~\cite{madireddy2023improvingperformancecontinuallearning}, PMI($43.23$\%)~\cite{pmi} and R-DFCIL ($59.41$\%)~\cite{R-DFCIL}). In particular, it performs comparably to the best memory-based methods like CN-DPM ($93.81$\%) ~\cite{Lee2020A} and DER++ ($92.21$\%). With bias correction, accuracy improves further to $92.11 \pm 1.18$\% , approaching strong memory-based methods such as CN-DPM ($93.81$\%) and DER++ ($92.21$\%), despite not storing any replay exemplars. On Split-CIFAR-10, our VAE (FT global) variant reaches an accuracy of $52.63 \pm 1.07$\% without bias correction, outperforming DER++ ($52.01$\%), ER-MIR ($37.80$\%) ~\cite{aljundi2019onlinecontinuallearningmaximally}, GSS ($49.22$\%) ~\cite{aljundi2019gradientbasedsampleselection}, and NNA-CIL ($52.55$\%). With bias correction, the result improves to $53.01 \pm 1.12$\%. For the more complex Split-CIFAR-100, where forgetting is more severe, our method demonstrates competitive performance. VAE (FT global) achieves $14.41 \pm 0.19$\%, exceeding most memory-free schemes and approaching the performance of CN-DPM ($16.13$\%) and DER++ ($15.04$\%). In contrast, the VAE (BMU specific) variant achieves $13.04 \pm 0.12\%$, despite being computationally more complex.  


\begin{table}[t]
\scriptsize
\centering
\caption{Final classification accuracy for multi class-incremental learning, pre-trained models.}
\label{tab:cil_pretrained_split}
\setlength{\tabcolsep}{3pt}   
\renewcommand{\arraystretch}{0.9}
\begin{tabular}{p{3.3cm}p{1.1cm}p{1.3cm}p{1.3cm}}
\toprule
\textbf{Method} & \textbf{CIFAR-10} & \textbf{CIFAR-100}  & \textbf{TinyImageNet} \\
& (5 tasks) & (10 tasks) & (10 tasks) \\
\midrule
EWC                & 31.82 & --    & 7.53 \\
SI                  & 27.43 & --    & 6.58 \\
LwF                 & 21.43 & 43.39 & 8.46 \\
iCaRL               & 71.15 & 50.74 & 23.22 \\
PGMA                & 74.31 & 17.47 & -- \\ 
RPSNet              & 83.37 & 25.27 & --   \\
OWM                 & 83.36 & 57.70   & 40.29 \\
PCL                 & \textbf{85.78} & 63.72 & 39.19 \\
DisCOIL             & 77.35 & --    & 19.75 \\
DyTox               & --    & 51.68 & 47.23 \\
NNA-CIL (INEL+CIFAR10) & 52.55 & 18.87 & -- \\
Continual-CLIP      & --    & 66.72 & \textbf{66.43} \\
DDGR      & --    & 63.40 & -- \\
R-DFCIL      & --    & 59.41  & -- \\
\midrule
\multicolumn{4}{l}{\textbf{Ours (SOM+CLIP w/o bias)}} \\
\midrule
SOM+CLIP (frozen global) & 77.61 & 64.82 & 58.53 \\
SOM+CLIP (FT global) & 79.84 & 66.41 & 59.21 \\
SOM+CLIP (BMU specific) & 83.37 & \textbf{67.89} & 61.94 \\
\midrule
\multicolumn{4}{l}{\textbf{Ours (SOM+CLIP w bias)}} \\
\midrule
SOM+CLIP (frozen global) & 78.44 & 65.21 & 59.55 \\
SOM+CLIP (FT global) & 81.24 & 67.19 & 60.13 \\
SOM+CLIP (BMU specific) & 83.94 & \textbf{69.12} & 62.71 \\
\bottomrule
\end{tabular}
\end{table}

On Split-TinyImageNet, the most complex benchmark due to its larger class set and higher intra-class variability, our methodology outperforms most other methods. VAE (FT global) with bias correction achieves $12.74 \pm 0.84$\%, outperforming memory-free baselines such as EWC ($7.53$\%), SI ($6.92$\%), and LwF ($8.41$\%). While memory-based strategies like DER++ ($11.09$\%) and ER-ACE ($12.11$\%) also perform well, our approach surpasses them without relying on a replay buffer. The BMU-specific variant achieves $12.15 \pm 0.24$\%, confirming that even under higher visual complexity, SOM-driven generative replay offers robustness and adaptability. When extending to CLIP embeddings (ViT-B/32), SOM replay yields substantial additional gains (Table~\ref{tab:cil_pretrained_split}). In the split CIFAR-10 benchmark, SOM+CLIP achieves $77.61$\% with frozen features, $79.84$\% when fine-tuned globally, and $83.37$\% with BMU-specific method. On CIFAR-100, the BMU-specific variant reaches $67.89$\%, outperforming PCL ($63.72$\%) and DyTox ($51.68$\%) and even Continual-Clip ($66.72$\%) \cite{thengane2022clipmodelefficientcontinual}, and further increases with bias correction to $69.12\%$. On TinyImageNet, SOM+CLIP attains $61.94$\% and increases to $62.71\%$ with bias correction, far above PCL ($39.19$\%) and DyTox ($47.23$\%), and competitive with continual-CLIP ($66.43$\%). Compared to DDGR~\cite{pmlr-v202-gao23e}, a recent diffusion-based generative replay method, our SOM+CLIP variants consistently achieve higher accuracy on CIFAR-100, highlighting the effectiveness of our lightweight replay strategy. These results highlight that while VAE-based SOM replay is competitive with replay-buffer baselines, combining our extended SOM methodology with foundation models like CLIP scales the framework to large, complex datasets. Interestingly, while BMU-specific VAEs underperform in the generative-from-scratch setting due to sparse training data per unit, the same design yields strong gains when applied to pretrained encoders such as CLIP, surpassing global fine-tuning.

These results suggest different trade-offs across continual learning paradigms. Memory-buffer methods often achieve strong performance by explicitly storing representative samples for rehearsal, as in GEM~\cite{gem}, iCaRL~\cite{rebuffi2017icarlincrementalclassifierrepresentation}, GSS~\cite{aljundi2019gradientbasedsampleselection}, ER-MIR~\cite{aljundi2019onlinecontinuallearningmaximally}, DER++~\cite{buzzega2020darkexperiencegeneralcontinual}, and ER-ACE~\cite{caccia2022newinsightsreducingabrupt}; however, their memory requirements increase with the replay buffer and may conflict with privacy or storage constraints. Regularization-based methods such as EWC~\cite{cf_1}, SI~\cite{zenke2017continuallearningsynapticintelligence}, and LwF~\cite{8107520} avoid storing data but often struggle as task interference increases. Our framework occupies an intermediate point by maintaining a compact SOM representation together with generative replay models, enabling replay without preserving original training samples while remaining competitive with strong rehearsal-based approaches.

Overall, these experiments demonstrate that the proposed SOM-based replay framework consistently benefits from preserving the topological organization of the latent space. The framework is particularly effective when combined with pretrained representations, where the SOM focuses on modeling feature organization rather than learning the representation itself. Although fully supervised IID-offline training remains the upper performance bound, the proposed approach offers a practical alternative for continual learning scenarios that prohibit retaining previous data, achieving competitive accuracy without storing original exemplars or requiring task identities during inference.

\section{Conclusion and Future Work}
\label{sec:conclusion}

This work introduces an incremental continual learning framework that integrates extended self-organizing maps (SOMs) with encoder--decoder models to enable exemplar-free replay with fixed-capacity statistical memory.  SOM units learn distribution statistics (running mean, variance and covariance) which can be utilized to generate synthetic samples to prevent forgetting. This methodology provides a number of benefits, including eliminating the need to store raw data in a memory buffer, easy visualization of progress, the ability to plug in any type of encoder/decoder (including foundation) models, and, additionally, the trained model can serve as a generative model of feasible examples of classes.

Experimental results across datasets, including complex ones such as  CIFAR10, CIFAR100, and TinyImageNet show that the extended SOM acts as an effective memory (sub)system capable of mitigating forgetting. The standalone SOM method is effective for low-dimensional datasets whereas the hybrid variants (non-pretrained VAEs or pretrained CLIP) scale well to high-dimensional ones. Our method outperforms or competes with state-of-the-art memory-based and memory-less architectures. Notably, our framework outperforms previous best methods on CIFAR-10 and CIFAR-100 single class incremental learning by nearly $10$\% and $7$\%, respectively, yielding significant improvement, and we also provide the first baseline results for single class incremental TinyImageNet.


Although our work shows an effective form of continual learning, there are still areas to improve. One drawback is that the BMU-specific VAE variant suffers from limited training data, reducing its effectiveness despite offering modularity and interpretability. In addition, the reliance on Gaussian statistics for replay may not fully capture complex class distributions, leading to reduced generative capability. Alternatively, the Gaussian distributions could potentially match multiple classes, reducing inference accuracy. We also note that while our experiments simulate class boundaries to pace replay, future work will evaluate boundary-free scheduling (e.g., fixed-period replay or drift detection) to more fully align with task-free continual learning. Future work will involve the use of a dynamic or growing SOMs that can adapt more effectively to varying class complexities, potentially leading to improved accuracy in higher-dimensional and higher-class cases. Moreover, this work can be extended with a deeper study of robust sampling strategies to improve the synthetic sample generation. Our methodology can also be extended to non-vision domains such as language models, reinforcement learning or time series prediction, which would further validate its applicability.

\appendices

\section{Experimental Settings}
\label{sec:experiment_settings}

Tables~\ref{tab:som_hyperparameters} and~\ref{tab:vae_som_hyperparams} summarize the hyperparameters used for SOM and VAE across different datasets. In all setups, the SOM unit vectors were initialized using random samples from the input data or latent space of the first class seen during training (i.e. class 0), in line with the CIL setup. We also experimented with other initialization strategies such as uniform random initialization over the fixed range of [0,1], and initialization with the global mean of input data, but these methods consistently underperformed compared to the random sampling method. For each SOM unit, the respective statistics were initialized as follows: the mean vector was set to zero, the variance vector to ones and the covariance matrix to zeroes. We also explored other initialization methods for the statistics such as using diagonal covariance matrices, fixed weight initializations and statistics derived from latent vectors from VAE (when VAE is used). However, these alternatives performed less accurately than the basic initialization of zero mean, unit variance and zero covariance. In all experiments, exponential moving averages were maintained for the mean, variance, and covariance of SOM unit activations. The following hyperparameter settings were explored for the bias correction:

\begin{table}[H]
\footnotesize
\centering
\caption{Bias-correction hyperparameters with best-performing values and additional settings tested.}
\label{tab:beta_values}
\begin{tabular}{lccc}
\toprule
\textbf{Parameter} & \textbf{Symbol} & \textbf{Best value} & \textbf{Other values tested} \\
\midrule
Mean               & $\beta_\mu$     & $0.99$              & $0.98$, $0.95$ \\
Variance / Cov.    & $\beta_\sigma, \beta_\Sigma$ & $0.95$ & $0.91$, $0.99$, $0.90$ \\
\bottomrule
\end{tabular}
\end{table}

To identify the BMUs in the SOMs, Euclidean distance between the input (latent) vector and the SOM unit weights was used. Euclidean distance produced the most accurate results across the datasets. While we experimented with alternative distance metrics including cosine similarity, Manhattan (L1) distance and Mahalanobis distance~\cite{ALY2014239}, these methods did not yield any significant improvements in the experiments.

The VAE was trained using the Adam optimizer, which provided consistent reconstruction quality across all datasets. We also evaluated other optimizers such as Stochastic Gradient Descent (SGD) with momentum and Root Mean Square Propagation (RMSprop), but these optimizers resulted in less stable reconstructions and as a result the SOMs replay quality was affected.

\begin{algorithm}
\scriptsize
\caption{Synthetic Sample Generation using Mean and Covariance (CIFAR-10)}
\begin{algorithmic}[1]
\REQUIRE Trained SOM, BMU coordinates $(i,j)$, number of samples $n$, regularization constant $\epsilon$
\STATE $\mu \leftarrow \text{SOM}.\text{running\_mean}[i][j]$
\STATE $\Sigma \leftarrow \text{SOM}.\text{running\_cov}[i][j]$
\STATE $\Sigma \leftarrow \Sigma + \epsilon I$ \hfill // Add diagonal regularization
\STATE Compute eigen-decomposition: $Q \Lambda Q^\top = \Sigma$
\STATE $\Lambda \leftarrow \max(\Lambda, \epsilon)$ \hfill // Clamp eigenvalues
\STATE $\Sigma_{\text{pd}} \leftarrow Q \cdot \text{diag}(\Lambda) \cdot Q^\top$
\STATE Sample $\tilde{z} \sim \mathcal{N}(\mu, \Sigma_{\text{pd}})$
\RETURN $n$ samples $\{\tilde{z}_1, ..., \tilde{z}_n\}$
\end{algorithmic}
\label{alg:cifar10-synth}
\end{algorithm}

In the CLIP-based experiments (Table~\ref{tab:clip_som_hyperparams}), we used the pretrained CLIP ViT-B/32 model with weights from \texttt{openai}. Unless otherwise noted, the encoder was kept frozen to retain the pretrained visual representation, though in some experiments, we optionally fine-tuned only the last transformer block to adapt to the continual learning setup. All images (CIFAR-10/100 at $32 \times 32$ and TinyImageNet at $64 \times 64$) were internally upsampled to $224 \times 224$ to match CLIP’s expected resolution. The pooled CLIP embeddings are 512-dimensional, which were then fed into a lightweight transposed-convolutional decoder that reconstructed images at their dataset-native resolution ($32 \times 32$ for CIFAR-10/100 and $64 \times 64$ for TinyImageNet).

\begin{table}[!h]
\centering
\small
\caption{SOM Hyperparameters for MNIST and Fashion-MNIST}
\label{tab:som_hyperparameters}
\begin{tabular}{lc}
\toprule
\textbf{Setting} & \textbf{Value} \\
\midrule
Datasets              & MNIST, Fashion-MNIST \\
Image Size              & $28 \times 28$ \\
SOM Sizes ($n \times n$)        & $10$, $12$, $20$, $30$, $35$ \\
SOM Epochs  & 10,20,30,50 \\
Sigma                 & 0.95 \\
Neighborhood Function & Gaussian \\
Distance Metric       & Euclidean \\
SOM Learning Rate & 0.5 \\
\bottomrule
\end{tabular}
\end{table}

\begin{table}[!t]
\centering
\caption{VAE and SOM hyperparameters for CIFAR-10, CIFAR-100, and TinyImageNet}
\label{tab:vae_som_hyperparams}
\small
\setlength{\tabcolsep}{3pt}
\renewcommand{\arraystretch}{1.05}

\begin{tabularx}{\columnwidth}{p{0.34\columnwidth} X}
\toprule
\textbf{Component} & \textbf{Hyperparameter Value} \\
\midrule
\multicolumn{2}{l}{\textbf{VAE Parameters}} \\
\midrule
Image Size & $32 \times 32 \times 3$; $64 \times 64 \times 3$ for TinyImageNet \\
Latent Dim. ($n \times 2 \times 2$) & 32, 64, 128 \\
VAE Epochs & 200 \\
Batch Size & 128, 64 \\
Learning Rate & $1 \times 10^{-4}$, $1 \times 10^{-5}$, $5 \times 10^{-5}$ \\
Optimizer & Adam \\
KL Loss Scale & 1.0 \\
Feature Loss Scale & 1.0 \\
Norm Type & BatchNorm \\
Residual Blocks & 1 bottleneck block \\
Channel Multiplier & 32 \\
Block Widths & [1, 2, 4, 8] \\
\midrule
\multicolumn{2}{l}{\textbf{SOM Parameters}} \\
\midrule
SOM Grid Sizes & 25, 30, 35, 40 \\
SOM Epochs & 150 \\
SOM Learning Rate & 0.5, 0.495, 0.45 \\
Sigma & 0.95, 0.94, 0.96 \\
Neighbor Function & Gaussian \\
Distance Metric & Euclidean \\
Covariance Regularization & $\lambda = 1 \times 10^{-4}$ \\
Replay Samples per BMU & $K = 1$ by default; this value can be increased \\
\bottomrule
\end{tabularx}
\end{table}

\begin{table}[!t]
\centering
\caption{CLIP ViT-B/32 and SOM hyperparameters for CIFAR-10, CIFAR-100, and TinyImageNet}
\label{tab:clip_som_hyperparams}
\small
\setlength{\tabcolsep}{3pt}
\renewcommand{\arraystretch}{1.05}

\begin{tabularx}{\columnwidth}{p{0.34\columnwidth} X}
\toprule
\textbf{Component} & \textbf{Hyperparameter Value} \\
\midrule
\multicolumn{2}{l}{\textbf{CLIP Encoder / Decoder}} \\
\midrule
Encoder Model & CLIP ViT-B/32 \\
Pretrained Weights & \texttt{openai} \\
Fine-tuning & Frozen by default; optional fine-tuning of the last transformer block \\
Embedding Dim. & 512 pooled embedding \\
Input Resolution to CLIP & $224 \times 224$, upsampled inside the wrapper \\
Decoder Target Size & CIFAR-10/100: $32 \times 32 \times 3$; TinyImageNet: $64 \times 64 \times 3$ \\
Decoder Type & Transposed-convolution upsampler from $4 \times 4$ to the target size \\
Decoder Base Channels & 1024 (XL); attention at $16^2$ and $32^2$ \\
Decoder Norm / Activations & GroupNorm(32), SiLU; final \texttt{tanh} with range $[-1,1]$ \\
\midrule
\multicolumn{2}{l}{\textbf{Training / Optimization}} \\
\midrule
Batch Size & 128 \\
Epochs per Class & 200 with cosine learning-rate schedule and warmup \\
Optimizer & Adam, with $\beta_1 = 0.9$ and $\beta_2 = 0.999$ \\
Learning Rate & $1 \times 10^{-4}$, $1 \times 10^{-5}$, $5 \times 10^{-5}$ \\
LR Schedule & 500 warmup steps followed by cosine decay \\
\midrule
\multicolumn{2}{l}{\textbf{SOM Parameters}} \\
\midrule
SOM Grid Size & 25, 30, 35, 40 \\
SOM Iterations per Class & 150 \\
Learning Rate & 0.5, 0.495, 0.45 \\
Sigma & 0.95, 0.94, 0.96 \\
Neighbor Function & Gaussian \\
Distance Metric & Euclidean \\
Covariance Regularization & $\lambda = 1 \times 10^{-4}$ \\
Replay Samples per BMU & $K = 1$ by default; this value can be increased \\
\bottomrule
\end{tabularx}
\end{table}

\begin{table}
\centering
\scriptsize
\caption{Last Accuracy on MNIST and Fashion-MNIST Datasets (10 tasks) on different SOM configurations}
\label{tab:mnist_fmnist_combined}
\begin{tabular}{@{}cccp{1.6cm}p{1.6cm}@{}}
\toprule
\textbf{SOM Size} & \textbf{Epochs} & \textbf{Sigma} & \textbf{MNIST} & \textbf{FMNIST} \\
\midrule
$10 \times 10$  & 10  & 0.95 & 87.95 & 72.79 \\
$10 \times 10$  & 50  & 0.94 & 87.97 & 73.83 \\
$12 \times 12$  & 20  & 0.95 & 90.67 & 76.24 \\
$20 \times 20$  & 20  & 0.95 & 93.01 & 79.42 \\
$30 \times 30$  & 30  & 0.95 & 94.28 & 81.08 \\
$35 \times 35$  & 50  & 0.95 & \textbf{95.16} & \textbf{81.91} \\
\bottomrule
\end{tabular}
\end{table}

\begin{table*}
\centering
\scriptsize
\caption{SOM+VAE (Global) Configuration and Last Accuracy on CIFAR-10 (10 tasks), CIFAR-100 (100 tasks), and TinyImageNet (10 tasks).}
\label{tab:som_vae_combined}
\begin{tabular}{@{}cccccc@{}}
\toprule
\textbf{Latent Dim} & \textbf{SOM Size} & \textbf{VAE+ SOM Epochs} & \textbf{CIFAR-10 } & \textbf{CIFAR-100} & \textbf{TinyImageNet} \\
\midrule
$32 \times 2 \times 2$  & $30 \times 30$    & 200+150 & 52.32            & 12.18 & 6.88 \\
  & $35 \times 35$    & 200+150 & 54.06            & 12.20 & 7.04 \\
  & $40 \times 40$    & 200+150 & \textbf{54.16}   & \textbf{12.41} & \textbf{7.14} \\
\midrule
$64 \times 2 \times 2$  & $25 \times 25$    & 200+150 & 45.18            & 11.83 & 6.89 \\
  & $30 \times 30$    & 200+150 & 46.53            & 11.88 & 6.57 \\
  & $35 \times 35$    & 200+150 & 37.72            & 11.01 & 6.44 \\
\midrule
$128 \times 2 \times 2$ & $25 \times 25$    & 200+150 & 44.21            & 11.13 & 6.66 \\
 & $30 \times 30$    & 200+150 & 41.77            & 11.10 & 6.12\\
\bottomrule
\end{tabular}
\end{table*}

\begin{table*}
\centering
\scriptsize
\caption{VAE (BMU specific) Configuration and Last Accuracy on CIFAR-10 (10 tasks), CIFAR-100 (100 tasks), and TinyImageNet (10 tasks).}
\label{tab:per_bmu_vae_som_combined}
\begin{tabular}{@{}cccccc@{}}
\toprule
\textbf{Latent Dim} & \textbf{SOM Size} & \textbf{VAE+ SOM Epochs} & \textbf{CIFAR-10} & \textbf{CIFAR-100} & \textbf{TinyImageNet} \\
\midrule
$32 \times 2 \times 2$  & $30 \times 30$ & 200+150 & 44.80 & 11.90 & 6.11 \\
 & $35 \times 35$ & 200+150 & 45.60 & 11.20 & 6.21 \\
 & $40 \times 40$ & 200+150 & \textbf{46.10} & \textbf{12.15} & \textbf{6.45} \\
\midrule
$64 \times 2 \times 2$  & $25 \times 25$ & 200+150 & 42.55 & 10.82 & 6.32 \\
 & $30 \times 30$ & 200+150 & 41.88 & 10.63 & 6.35 \\
 & $35 \times 35$ & 200+150 & 43.40 & 11.18 & 6.11 \\
\midrule
$128 \times 2 \times 2$ & $25 \times 25$ & 200+150 & 40.85 & 11.95 & 6.21 \\
 & $30 \times 30$ & 200+150 & 40.10 & 11.88 & 5.99 \\
\bottomrule
\end{tabular}
\end{table*}

\section{Effect of SOM Resolution and VAE Capacity on Incremental Learning}
\label{sec:app_som_vae_dimensions}

Table \ref{tab:mnist_fmnist_combined} reports the accuracy trends for MNIST and Fashion-MNIST under different SOM configurations. We observe that larger SOM resolutions significantly improve performance, with the test accuracy rising with the size of the grid. In MNIST, the precision increases from 87.95\% at $10 \times 10$ to 95.16\% at
$35 \times 35$, approaching the performance of the offline i.i.d. upper bound (95.82\%). In FashionMNIST, the performance improves from 72.79\% to 81.91\% under the same grid expansion of the SOM. These results demonstrate the model's capacity to retain and learn new knowledge using only synthetic replay generated from SOM statistics. It should also be noted that the accuracy is achieved without any generative model or memory buffer, relying solely on the internal topology and statistics of the SOM for stable representation learning and replay.

Table \ref{tab:som_vae_combined} presents the results for CIFAR-10 and CIFAR-100 under the one-class-at-a-time method using a VAE (FT global) hybrid model. We investigated both the size of the SOM and the latent dimensionality of the VAE. For CIFAR-10, the best configuration ($32 \times 32 \times 2 \times 2$ latent, $40 \times 40$ SOM) achieves an accuracy of 54.16\%, even surpassing the performance of standard split-CIFAR-10 (53.01\%). This demonstrates the robustness of our method to finer-grained tasks. In the CIFAR-100 stream, our model achieves an accuracy of 12.41\% with the same configuration. 

We further evaluate on our VAE (FT BMU specific) method (Table~\ref{tab:per_bmu_vae_som_combined}), where a separate VAE distribution is associated with each BMU. On CIFAR-10, this method achieved a peak accuracy of 46.10\%, lower than the shared VAE (FT global) approach. This might suggest that dividing the latent space too finely could hurt the generalization in the BMUs. A similar trend is observed on CIFAR-100, where the best accuracy is 12.15\%. Even though VAE (FT BMU specific) is more complex, it performs worse than the shared latent distribution model. This suggests that using a single shared representation might be more effective for stable memory replay in high-dimensional data.

We extend our evaluation to TinyImageNet, a very complex benchmark consisting of 200 classes, which we structure into a 10-task stream with 20 classes per task. This setting is considerably more challenging due to both the fine-grained nature of the categories and the larger classification space. The global VAE (FT global) achieves a peak accuracy of 7.14\%, higher than the BMU-specific variant at 6.45\%. These results further confirm the scalability of the shared VAE (FT global) representation, which maintains more stable replay dynamics than its BMU-specific counterpart even under highly complex continual learning conditions.

\subsubsection{Remarks:}We find that larger SOM sizes consistently lead to better performance across datasets (e.g. $35 \times 35 > 30 \times 30 > 25 \times 25$), which aligns with trends observed on all datasets. However, increasing the SOM size also leads to longer training times and higher computational complexity, as the number of BMUs (neurons) grows quadratically. Similarly, increasing the latent dimensionality in the VAE beyond $32  \times 2 \times 2$ does not improve the accuracy; it often degrades it. For example, using $128 \times 2 \times 2$ results in a drop in precision to 38.94\% on CIFAR-10. Larger latent spaces are more expensive to train and may introduce sparsity in the SOM map, which might make it harder for the SOM to organize meaningful structures. These results highlight that compact, well-structured latent representations and moderate SOM sizes strike the best balance between performance and efficiency.

\section{Class-Level Analysis (Confusion Matrices)}
\label{app:confusion_matrices}

\begin{figure*}[t]
    \centering
    \setlength{\tabcolsep}{2pt}
    \renewcommand{\arraystretch}{0.9}

    \begin{tabular}{cccc}
        \includegraphics[width=0.24\textwidth]{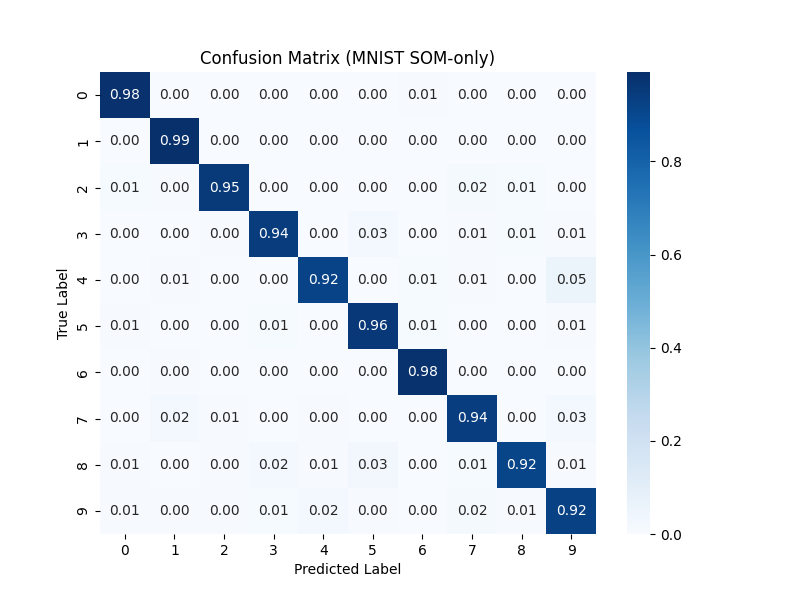} &
        \includegraphics[width=0.24\textwidth]{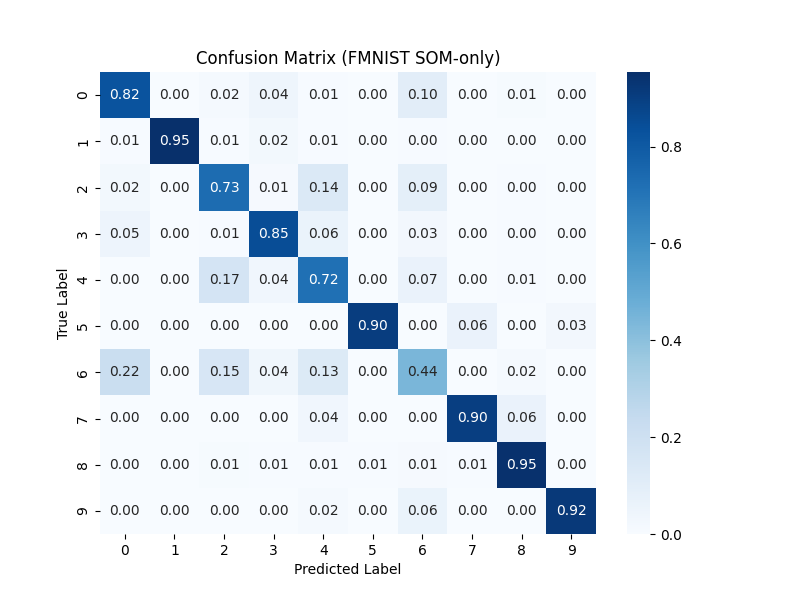} &
        \includegraphics[width=0.24\textwidth]{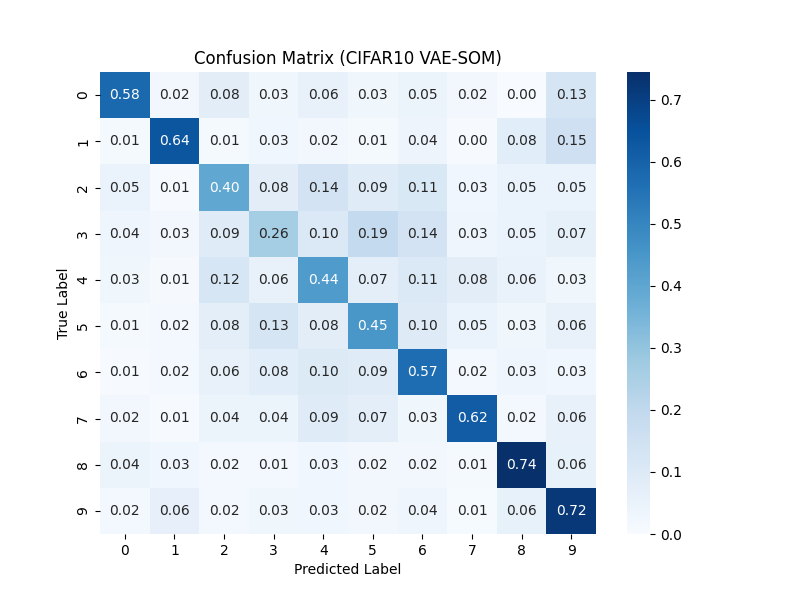} &
        \includegraphics[width=0.24\textwidth]{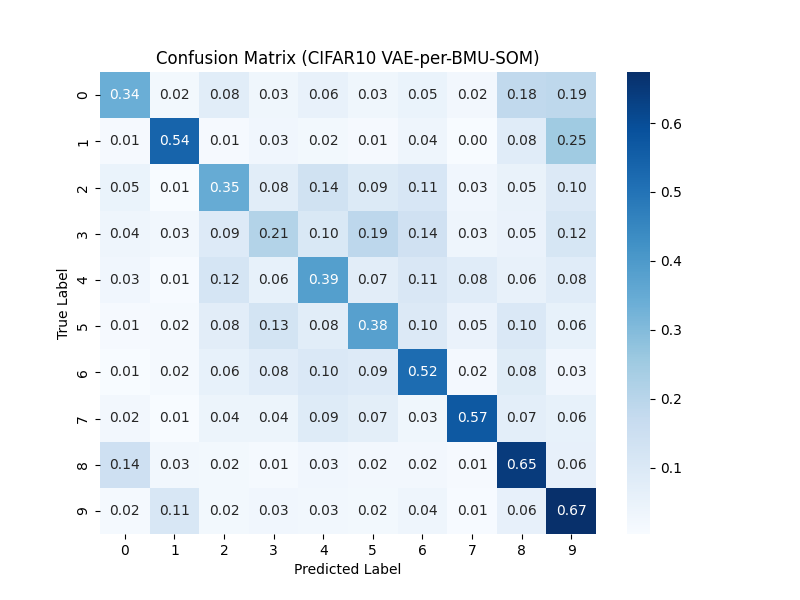} \\[-1mm]

        \footnotesize (a) MNIST &
        \footnotesize (b) Fashion-MNIST &
        \footnotesize (c) CIFAR-10 global &
        \footnotesize (d) CIFAR-10 BMU-specific \\[2mm]

        \includegraphics[width=0.24\textwidth]{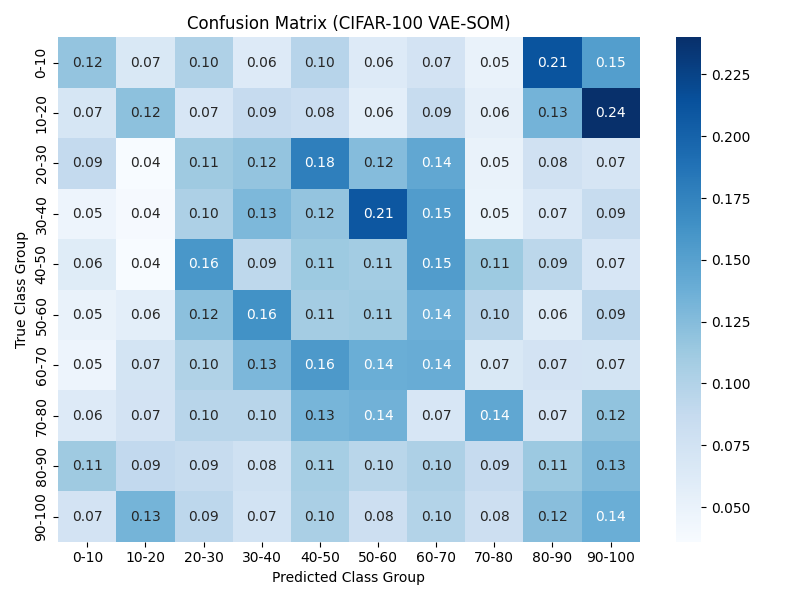} &
        \includegraphics[width=0.24\textwidth]{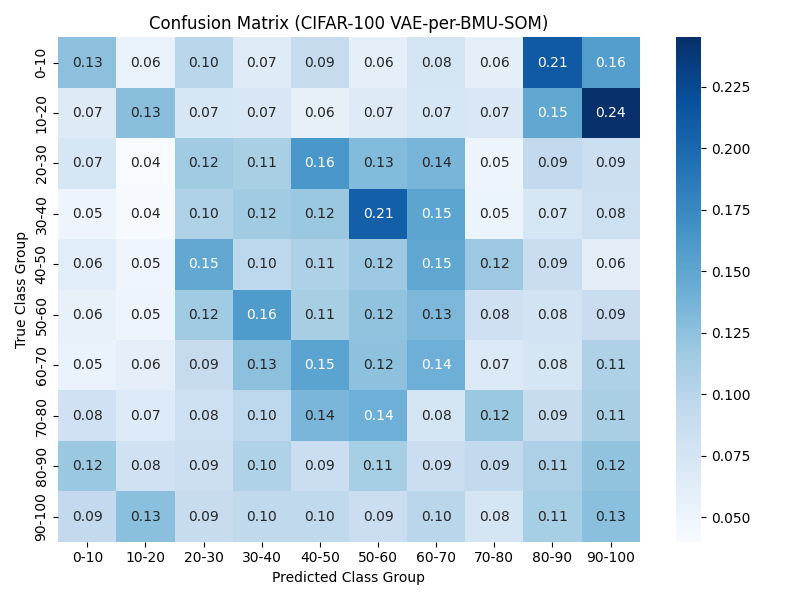} &
        \includegraphics[width=0.24\textwidth]{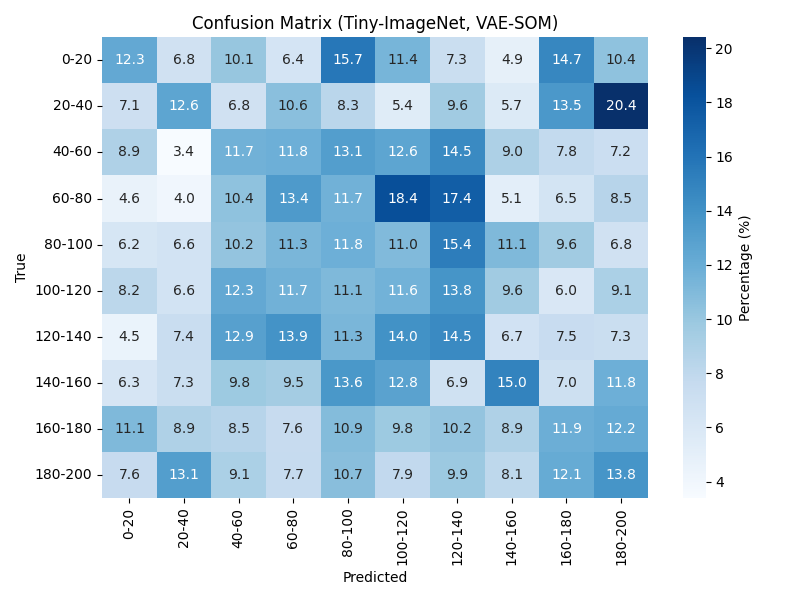} &
        \includegraphics[width=0.24\textwidth]{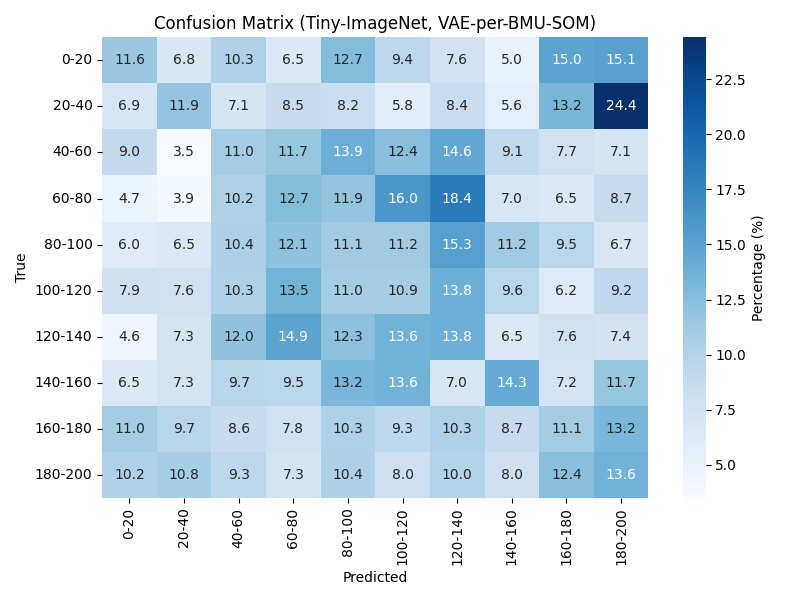} \\[-1mm]

        \footnotesize (e) CIFAR-100 global &
        \footnotesize (f) CIFAR-100 BMU-specific &
        \footnotesize (g) TinyImageNet global &
        \footnotesize (h) TinyImageNet BMU-specific
    \end{tabular}

    \caption{Confusion matrices for MNIST, Fashion-MNIST, CIFAR-10, CIFAR-100, and TinyImageNet. For CIFAR-10, CIFAR-100, and TinyImageNet, the global and BMU-specific VAE variants are shown separately.}
    \label{fig:all_heatmaps}
\end{figure*}

To complement the aggregate accuracy results presented in the main paper, we include full normalized confusion matrices for all datasets under different variants of our method. Each matrix is row-normalized, allowing us to visualize the distribution of predictions per true class and to highlight which classes are most affected by forgetting or confusion over the course of continual learning. Figure ~\ref{fig:all_heatmaps} shows the confusion matrices for the evaluated datasets. For MNIST and Fashion-MNIST, we evaluate the SOM-only variant without auxiliary generative replay. The resulting matrices exhibit strong diagonal dominance, with MNIST showing nearly perfect separation across all classes. FMNIST, while slightly noisier, retains clear diagonals with only minor confusion among semantically similar clothing items (e.g., shirts vs. coats). These results confirm that SOM-only replay is sufficient for simple, low-variance datasets.

On CIFAR-10 , we compare the VAE (FT BMU specific) and the global VAE (FT global) variants. Both maintain clear diagonal structures, but the per-BMU approach exhibits broader off-diagonal activations, especially among visually similar vehicle categories. By contrast, the global variant produces slightly better diagonals than the global VAE and fewer cross-class confusions, suggesting that aggregating statistics across the full latent space provides a more stable basis for replay.

For CIFAR-100, where classes are grouped into 10 superclasses, both the per-BMU and global VAE (FT global) variants preserve diagonal structure but also exhibit leakage into neighboring groups. The per-BMU variant shows more dispersion, while the global variant achieves slightly cleaner diagonals and reduced cross-group mixing, suggesting modest but consistent benefits from pooling statistics at the global level. On TinyImageNet, which is the most challenging benchmark with 200 fine-grained classes, both variants again preserve recognizable diagonals but suffer from substantial leakage across adjacent groups due to strong visual overlap between classes. The global VAE (FT global) provides a slight improvement over the per-BMU variant, reinforcing its relative robustness as dataset complexity increases.

Overall, these confusion matrix visualizations (Figure~\ref{fig:all_heatmaps}) reinforce our main results: SOM-based replay maintains per-class retention across diverse benchmarks, and the global VAE (FT global) and the per-BMU variant consistently improve stability and scalability as dataset complexity increases.





\section{Analysis of MNIST, Fashion-MNIST, CIFAR10 and CIFAR100 Results}
Tables~\ref{tab:R-matrix-mnist-35}--\ref{tab:metrics-all} report the full
per-task accuracy matrices and forgetting statistics for the best-performing SOM model from Table \ref{tab:mnist_fmnist_combined} and \ref{tab:som_vae_combined} for datasets MNIST, Fashion MNIST and CIFAR10. We evaluated the metrices: forward transfer (FWT), backward transfer (BWT), and forgetting
using the standard formulation of \cite{lopez}. In all datasets, the model
achieves perfect performance on the first task and maintains high accuracy on
most subsequent tasks despite the absence of an external replay buffer. The
structure of the $R$ matrices reveals smooth degradation along the rows,
consistent with mild but controlled catastrophic forgetting. 
Since the SOM is trained in a single-head class-incremental setting, the model
has no prior ability on unseen classes, yielding $R_{t-1,t}\approx 0$ before
learning task $t$. As a result, FWT is consistently negative and equal in
magnitude to the per-class baseline accuracy, indicating that no zero-shot
transfer occurs across tasks. BWT values for early tasks are slightly
negative, reflecting mild forgetting after subsequent updates, while later
tasks show near-zero or slightly positive BWT, consistent with the reduced
number of interfering updates. Forgetting follows the same trend: early
classes exhibit small positive forgetting, whereas the last task exhibits
large negative forgetting, as its accuracy improves from zero before training
to its final post-training value.

\begin{table*}
\centering
\scriptsize
\caption{Per-task class accuracy matrix $R[t,j]$ for SOM (35$\times$35) on MNIST (10 tasks, one class at a time).}
\label{tab:R-matrix-mnist-35}
\begin{tabular}{c|cccccccccc}
\toprule
\textbf{t $\backslash$ j} &
\textbf{0} & \textbf{1} & \textbf{2} & \textbf{3} &
\textbf{4} & \textbf{5} & \textbf{6} & \textbf{7} &
\textbf{8} & \textbf{9} \\
\midrule
0 & 1.0000 & -- & -- & -- & -- & -- & -- & -- & -- & -- \\
1 & 0.9997 & 0.9995 & -- & -- & -- & -- & -- & -- & -- & -- \\
2 & 0.9985 & 0.9978 & 0.9875 & -- & -- & -- & -- & -- & -- & -- \\
3 & 0.9972 & 0.9980 & 0.9836 & 0.9894 & -- & -- & -- & -- & -- & -- \\
4 & 0.9960 & 0.9972 & 0.9821 & 0.9866 & 0.9902 & -- & -- & -- & -- & -- \\
5 & 0.9940 & 0.9975 & 0.9784 & 0.9698 & 0.9852 & 0.9554 & -- & -- & -- & -- \\
6 & 0.9908 & 0.9960 & 0.9752 & 0.9674 & 0.9835 & 0.9496 & 0.9742 & -- & -- & -- \\
7 & 0.9894 & 0.9952 & 0.9706 & 0.9648 & 0.9820 & 0.9462 & 0.9738 & 0.9730 & -- & -- \\
8 & 0.9878 & 0.9940 & 0.9662 & 0.9610 & 0.9786 & 0.9440 & 0.9712 & 0.9704 & 0.9310 & -- \\
9 & 0.9862 & 0.9922 & 0.9618 & 0.9584 & 0.9736 & 0.9406 & 0.9704 & 0.9672 & 0.9282 & 0.9620 \\
\bottomrule
\end{tabular}
\end{table*}

\begin{table*}
\centering
\scriptsize
\caption{Per-task class accuracy matrix $R[t,j]$ for SOM (35$\times$35) on Fashion-MNIST (10 tasks, one class at a time).}
\label{tab:R-matrix-fmnist-35}
\begin{tabular}{c|cccccccccc}
\toprule
\textbf{t $\backslash$ j} &
0 & 1 & 2 & 3 & 4 & 5 & 6 & 7 & 8 & 9 \\
\midrule
0 & 1.0000 & -- & -- & -- & -- & -- & -- & -- & -- & -- \\
1 & 0.9962 & 0.9870 & -- & -- & -- & -- & -- & -- & -- & -- \\
2 & 0.9680 & 0.9804 & 0.9640 & -- & -- & -- & -- & -- & -- & -- \\
3 & 0.9402 & 0.9576 & 0.9504 & 0.9200 & -- & -- & -- & -- & -- & -- \\
4 & 0.9496 & 0.9642 & 0.7710 & 0.8554 & 0.7702 & -- & -- & -- & -- & -- \\
5 & 0.9360 & 0.9618 & 0.7868 & 0.8662 & 0.7344 & 0.9984 & -- & -- & -- & -- \\
6 & 0.8524 & 0.9606 & 0.7324 & 0.8520 & 0.7230 & 0.9978 & 0.4708 & -- & -- & -- \\
7 & 0.8726 & 0.9594 & 0.7122 & 0.8550 & 0.7124 & 0.9402 & 0.4580 & 0.9338 & -- & -- \\
8 & 0.8720 & 0.9586 & 0.7074 & 0.8496 & 0.7090 & 0.9460 & 0.4188 & 0.9312 & 0.9520 & -- \\
9 & 0.8602 & 0.9508 & 0.7100 & 0.8324 & 0.6880 & 0.9318 & 0.4360 & 0.8710 & 0.9390 & 0.8920 \\
\bottomrule
\end{tabular}
\end{table*}

\begin{table*}
\centering
\scriptsize
\caption{Per-task class accuracy matrix $R$ for CIFAR-10 across 10 sequential tasks for a 40x40 SOM with a global VAE with latent dimension 32x2x2 configuration.}
\label{tab:R-matrix-cifar10}\begin{tabular}{c|cccccccccc}
\hline
\textbf{t $\backslash$ j} & 0 & 1 & 2 & 3 & 4 & 5 & 6 & 7 & 8 & 9 \\
\hline
0 & 1.000 & -- & -- & -- & -- & -- & -- & -- & -- & -- \\
1 & 0.863 & 0.819 & -- & -- & -- & -- & -- & -- & -- & -- \\
2 & 0.718 & 0.780 & 0.768 & -- & -- & -- & -- & -- & -- & -- \\
3 & 0.680 & 0.765 & 0.583 & 0.610 & -- & -- & -- & -- & -- & -- \\
4 & 0.653 & 0.764 & 0.453 & 0.558 & 0.528 & -- & -- & -- & -- & -- \\
5 & 0.650 & 0.762 & 0.425 & 0.420 & 0.512 & 0.445 & -- & -- & -- & -- \\
6 & 0.648 & 0.759 & 0.404 & 0.355 & 0.461 & 0.441 & 0.569 & -- & -- & -- \\
7 & 0.637 & 0.740 & 0.399 & 0.350 & 0.415 & 0.424 & 0.559 & 0.619 & -- & -- \\
8 & 0.506 & 0.723 & 0.380 & 0.349 & 0.410 & 0.420 & 0.564 & 0.616 & 0.734 & -- \\
9 & 0.580 & 0.640 & 0.400 & 0.260 & 0.440 & 0.450 & 0.570 & 0.620 & 0.740 & 0.720 \\
\hline
\end{tabular}

\end{table*}

\begin{table*}
\centering
\scriptsize
\caption{Task accuracy matrix $R$ for CIFAR-100 across 10 sequential tasks for a 40x40 SOM with a global VAE with latent dimension 32x2x2 configuration.}
\label{tab:R-matrix-cifar100}\begin{tabular}{c|cccccccccc}
\hline
\textbf{t $\backslash$ j} &
0 & 1 & 2 & 3 & 4 & 5 & 6 & 7 & 8 & 9 \\
\hline
0 & 1.000 & -- & -- & -- & -- & -- & -- & -- & -- & -- \\
1 & 0.812 & 0.603 & -- & -- & -- & -- & -- & -- & -- & -- \\
2 & 0.402 & 0.381 & 0.364 & -- & -- & -- & -- & -- & -- & -- \\
3 & 0.361 & 0.372 & 0.350 & 0.342 & -- & -- & -- & -- & -- & -- \\
4 & 0.329 & 0.345 & 0.322 & 0.337 & 0.310 & -- & -- & -- & -- & -- \\
5 & 0.301 & 0.321 & 0.298 & 0.314 & 0.296 & 0.284 & -- & -- & -- & -- \\
6 & 0.277 & 0.298 & 0.284 & 0.296 & 0.281 & 0.269 & 0.259 & -- & -- & -- \\
7 & 0.253 & 0.276 & 0.264 & 0.279 & 0.268 & 0.257 & 0.247 & 0.236 & -- & -- \\
8 & 0.192 & 0.209 & 0.204 & 0.216 & 0.211 & 0.203 & 0.198 & 0.191 & 0.187 & -- \\
9 & 0.120 & 0.120 & 0.110 & 0.210 & 0.110 & 0.140 & 0.140 & 0.140 & 0.110 & 0.140 \\

\hline
\end{tabular}
\end{table*}

\section{SOM Architecture and Encoder–Decoder Integration}
\label{app:enc_dec_arch}
For datasets trained from scratch, we employed the residual VAE in Table~\ref{tab:vae_architecture}, which compresses CIFAR images into latent codes of size $32 \times 2 \times 2$, $64 \times 2 \times 2$, or $128 \times 2 \times 2$, reconstructed symmetrically with ResUp blocks (12.6M parameters for the $32$-dimensional variant). For pretrained settings, the encoder was replaced with the frozen CLIP ViT-B/32 visual tower, while the decoder (Table~\ref{tab:clip_decoder}) upsampled the 512-D embedding through residual and attention blocks to match dataset resolution ($32 \times 32$ for CIFAR-10/100, $64 \times 64$ for TinyImageNet).  In the fine-tune setting, only the final transformer block of CLIP was unfrozen. Latent vectors from either encoder were projected onto SOM grids between $20 \times 20$ and $40 \times 40$ units, with $40 \times 40$ providing the best balance of capacity, stability and, computation. Each SOM unit stored running statistics (mean, variance, covariance), which were used to generate replay samples via the respective decoder.

Each SOM unit stored running statistics (mean, variance, covariance), which were used to generate replay samples via the respective decoder. During replay, latent samples were drawn from the distributions associated with previously populated SOM units and decoded back into input space using our decoder setup. This design enables the SOM to act as a compact statistical memory while preserving the topological organization of the learned latent space.

\begin{table*}[!t]
\renewcommand{\arraystretch}{1.15}
\centering
\small
\caption{Per-task Forward Transfer (FWT), Backward Transfer (BWT), and
Forgetting across MNIST, Fashion-MNIST, CIFAR-10, and CIFAR-100.}
\label{tab:metrics-all}

\begin{tabular}{c|ccc|ccc|ccc|ccc}
\hline
& \multicolumn{3}{c|}{\textbf{MNIST}}
& \multicolumn{3}{c|}{\textbf{Fashion-MNIST}}
& \multicolumn{3}{c|}{\textbf{CIFAR-10}}
& \multicolumn{3}{c}{\textbf{CIFAR-100}} \\
\cline{2-4}\cline{5-7}\cline{8-10}\cline{11-13}

\textbf{Task}
& \textbf{FWT} & \textbf{BWT} & \textbf{F}
& \textbf{FWT} & \textbf{BWT} & \textbf{F}
& \textbf{FWT} & \textbf{BWT} & \textbf{F}
& \textbf{FWT} & \textbf{BWT} & \textbf{F} \\
\hline

0 & -.7133 & -- & --
  & -.6900 & -- & --
  & -.323 & -- & --
  & -.228 & -- & -- \\

1 & -.8943 & -.0073 & .0073
  & -.6810 & -.0362 & .0362
  & -.203 & -.179 & .179
  & -.102 & -.483 & .483 \\

2 & -.4002 & -.0257 & .0257
  & -.5130 & -.2540 & .2540
  & -.161 & -.368 & .368
  & -.089 & -.254 & .254 \\

3 & -.4723 & -.0310 & .0310
  & -.6860 & -.0876 & .0876
  & -.065 & -.350 & .350
  & -.082 & -.132 & .132 \\

4 & -.3595 & -.0166 & .0166
  & -.4900 & -.0822 & .0822
  & -.160 & -.088 & .088
  & -.079 & -.200 & .200 \\

5 & -.3901 & -.0148 & .0148
  & -.5540 & -.0666 & .0666
  & -.192 & .005 & -.005
  & -.075 & -.144 & .144 \\

6 & -.7589 & -.0038 & .0038
  & -.2680 & -.0348 & .0348
  & -.212 & .001 & -.001
  & -.073 & -.119 & .119 \\

7 & -.2977 & -.0058 & .0058
  & -.4010 & -.0628 & .0628
  & -.223 & .001 & -.001
  & -.071 & -.096 & .096 \\

8 & -.3203 & -.0028 & .0028
  & -.6150 & -.0130 & .0130
  & -.329 & .006 & -.006
  & -.069 & -.077 & .077 \\

9 & -.4767 & .0000 & --
  & -.8130 & .0000 & --
  & -.261 & .000 & --
  & -.067 & .000 & -.140 \\

\hline
\end{tabular}
\end{table*}

\begin{table*}[!t]
\centering
\caption{Architecture of the VAE model used for CIFAR-10/100 (input size: $3 \times 32 \times 32$). ResDown and ResUp denote residual blocks with downsampling and upsampling, respectively. All activations are ELU except for the output layer, which uses Tanh. \textbf{Notation:} Type = kernel size or layer type, Str. = stride or upsampling scale, Act. = activation function, Skip = presence of skip connection.}

\label{tab:vae_architecture}
\scriptsize
\begin{tabular}{ccccccc}
\toprule
\textbf{Layer} & 
\makecell{\textbf{Kernel}\\\textbf{/ Type}} & 
\makecell{\textbf{Stride}\\\textbf{/ Scale}} & 
\textbf{Act.} & 
\textbf{Skip} & 
\makecell{\textbf{Output}\\\textbf{Shape}} & 
\textbf{Params} \\
\midrule
\multicolumn{7}{c}{\textit{Encoder}} \\
\midrule
Conv2d (Input)   & $3 \times 3$ / Conv     & 1           & –     & No  & $32 \times 32 \times 32$  & 896 \\
ResDown Block 1  & $3 \times 3$ / ×2       & 2 + 1       & ELU   & Yes & $64 \times 16 \times 16$  & 46.3K \\
ResDown Block 2  & $3 \times 3$ / ×2       & 2 + 1       & ELU   & Yes & $128 \times 8 \times 8$   & 184.6K \\
ResDown Block 3  & $3 \times 3$ / ×2       & 2 + 1       & ELU   & Yes & $256 \times 4 \times 4$   & 737.9K \\
ResDown Block 4  & $3 \times 3$ / ×2       & 2 + 1       & ELU   & Yes & $512 \times 2 \times 2$   & 2.36M \\
ResBlock         & $3 \times 3$ / ×2       & 1 + 1       & ELU   & Yes & $512 \times 2 \times 2$   & 2.36M \\
Conv2d ($\mu$)   & $1 \times 1$ / Conv     & 1           & –     & No  & $32 \times 2 \times 2$    & 16.4K \\
Conv2d (log$\sigma^2$)   & $1 \times 1$ / Conv     & 1           & –     & No  & $32 \times 2 \times 2$    & 16.4K \\
\midrule
\multicolumn{7}{c}{\textit{Decoder}} \\
\midrule
Conv2d (Latent)  & $1 \times 1$ / Conv     & 1           & ELU   & No  & $512 \times 2 \times 2$   & 16.9K \\
ResBlock         & $3 \times 3$ / ×2       & 1 + 1       & ELU   & Yes & $512 \times 2 \times 2$   & 2.36M \\
ResUp Block 1    & $3 \times 3$ / ×2       & Upsample×2  & ELU   & Yes & $256 \times 4 \times 4$   & 627.4K \\
ResUp Block 2    & $3 \times 3$ / ×2       & Upsample×2  & ELU   & Yes & $128 \times 8 \times 8$   & 295.2K \\
ResUp Block 3    & $3 \times 3$ / ×2       & Upsample×2  & ELU   & Yes & $64 \times 16 \times 16$  & 184.5K \\
ResUp Block 4    & $3 \times 3$ / ×2       & Upsample×2  & ELU   & Yes & $32 \times 32 \times 32$  & 82.9K \\
Conv2d (Output)  & $5 \times 5$ / Conv     & 1           & Tanh  & No  & $3 \times 32 \times 32$   & 2.4K \\
\midrule
\textbf{Total Parameters} & & & & & & \textbf{12.6M} \\
\bottomrule
\end{tabular}
\end{table*}

\begin{table*}[!t]
\centering
\scriptsize
\caption{Decoder Architecture used for CLIP+SOM experiments. The encoder is the CLIP ViT-B/32 visual tower (pretrained) and is \emph{frozen} by default; when \texttt{finetune\_last} is enabled, only the last transformer block of ViT-B/32 is unfrozen. The decoder upsamples a single CLIP embedding to the target RGB image in $[-1,1]$. GN denotes GroupNorm(32). SelfAttention2d uses $4$ heads with GN(32). Decoder output target size is set to match dataset resolution ($32\times32$ for CIFAR-10/100, $64\times64$ for TinyImageNet).}
\label{tab:clip_decoder}
\begin{tabular}{cccccc}
\toprule
\textbf{Layer} & \textbf{Operation / Details} & \textbf{Blocks} & \textbf{Activation} & \textbf{GN} & \textbf{Output Shape} \\
\midrule
Linear (embed$\rightarrow$4$\times$4) & FC($D\!\rightarrow\!$base\_ch$\cdot4\cdot4$) & -- & --   & Yes & $B\times$1024$\times4\times4$ \\
ResBlock                               & Conv $3{\times}3$ (GN+SiLU)                 & $1{+}1$ & SiLU & Yes & $B\times$1024$\times4\times4$ \\
Upsample                               & ConvTranspose2d $4{\times}4$                 & $\times2$ & -- & Yes & $B\times$1024$\times8\times8$ \\
ResBlock (ch$\downarrow$)              & Conv $3{\times}3$ (GN+SiLU)                 & $1{+}1$ & SiLU & Yes & $B\times$512$\times8\times8$ \\
Upsample                               & ConvTranspose2d $4{\times}4$                 & $\times2$ & -- & Yes & $B\times$512$\times16\times16$ \\
ResBlock (+Attn)                       & Conv $3{\times}3$ + SelfAttn2d               & $1{+}1$ & SiLU & Yes & $B\times$512$\times16\times16$ \\
ResBlock (ch$\downarrow$)              & Conv $3{\times}3$ (GN+SiLU)                 & $1{+}1$ & SiLU & Yes & $B\times$256$\times16\times16$ \\
Upsample                               & ConvTranspose2d $4{\times}4$                 & $\times2$ & -- & Yes & $B\times$256$\times32\times32$ \\
ResBlock (+Attn)                       & Conv $3{\times}3$ + SelfAttn2d               & $1{+}1$ & SiLU & Yes & $B\times$256$\times32\times32$ \\
ResBlock (ch$\downarrow$)              & Conv $3{\times}3$ (GN+SiLU)                 & $1{+}1$ & SiLU & Yes & $B\times$128$\times32\times32$ \\
Upsample                               & ConvTranspose2d $4{\times}4$                 & $\times2$ & -- & Yes & $B\times$128$\times64\times64$ \\
ResBlock                               & Conv $3{\times}3$ (GN+SiLU)                 & $1{+}1$ & SiLU & Yes & $B\times$128$\times\text{target}\times\text{target}$ \\
Output head                            & GN $\rightarrow$ SiLU $\rightarrow$ $3{\times}3$ $\rightarrow$ Tanh & $1$ & Tanh & Yes & $B\times3\times\text{target}\times\text{target}$ \\
\bottomrule
\end{tabular}
\end{table*}

\bibliographystyle{IEEEtran}
\bibliography{references}

\clearpage

\section*{Supplementary Material}

\section*{Generative Capability of the Proposed Method}
\label{app:generative_model}

\begin{figure}[h]
    \centering
    \subfloat[MNIST]{%
        \includegraphics[width=0.45\linewidth]{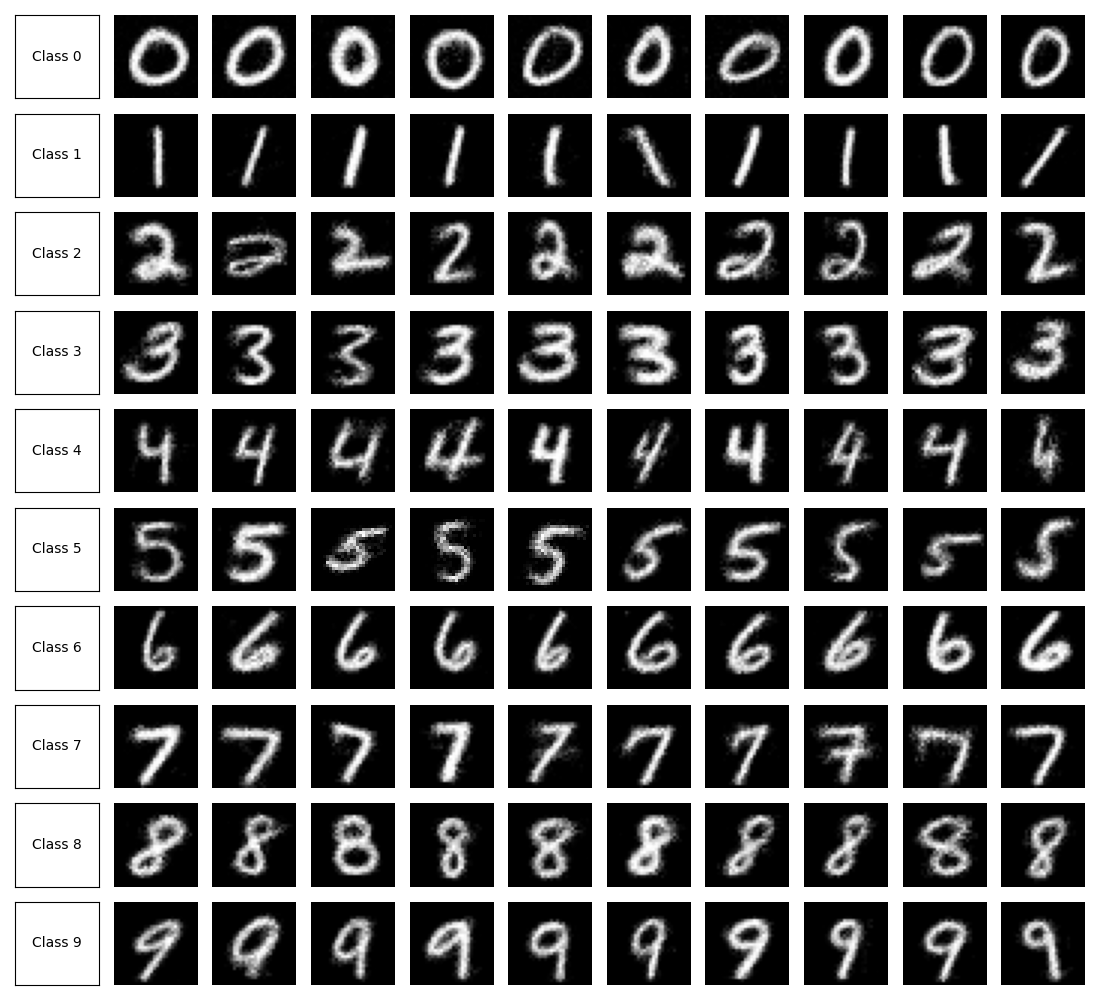}
        \label{fig:som_synthetic_mnist}
    }
    \hfill
    \subfloat[Fashion-MNIST]{%
        \includegraphics[width=0.45\linewidth]{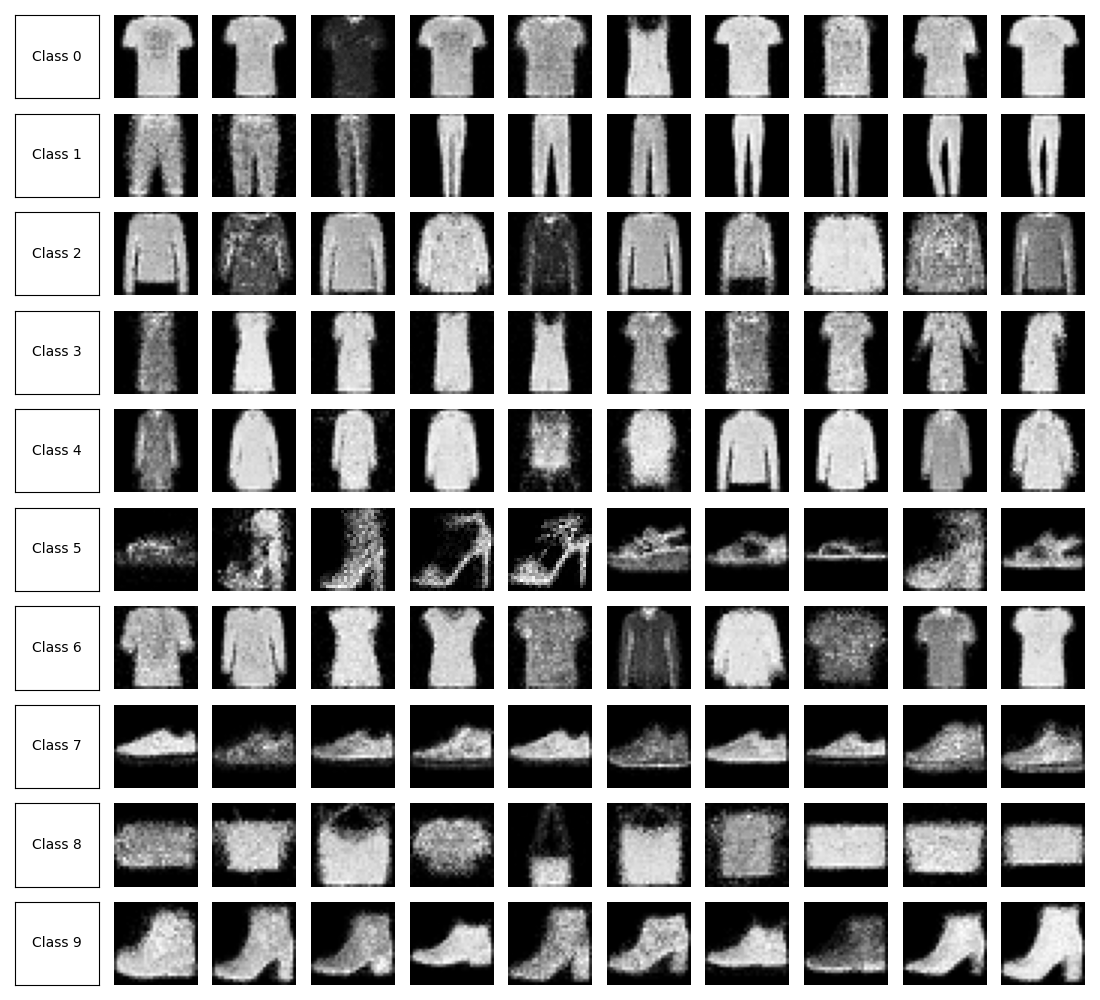}
        \label{fig:som_synthetic_fmnist}
    }
    \caption{Synthetic samples generated using Gaussian sampling from the trained SOM. Each row corresponds to one class (0–9), and each row contains 10 synthetic samples generated from BMUs labeled with that class.}
    \label{fig:som_synthetic_samples}
\end{figure}

Our proposed model can also be used as a generative model, which is capable of synthesizing samples without storing original data. For MNIST and FashionMNIST, as we only use the SOM for training, image samples are generated using Gaussian sampling as described in the paper. After training, each SOM unit stores the specific statistics in the form of mean and variance, and they are used to generate sample images. Figure~\ref{fig:som_synthetic_samples} illustrates some of the samples generated for each class using this method.

\begin{figure}[h]
    \centering
    \includegraphics[width=0.24\textwidth]{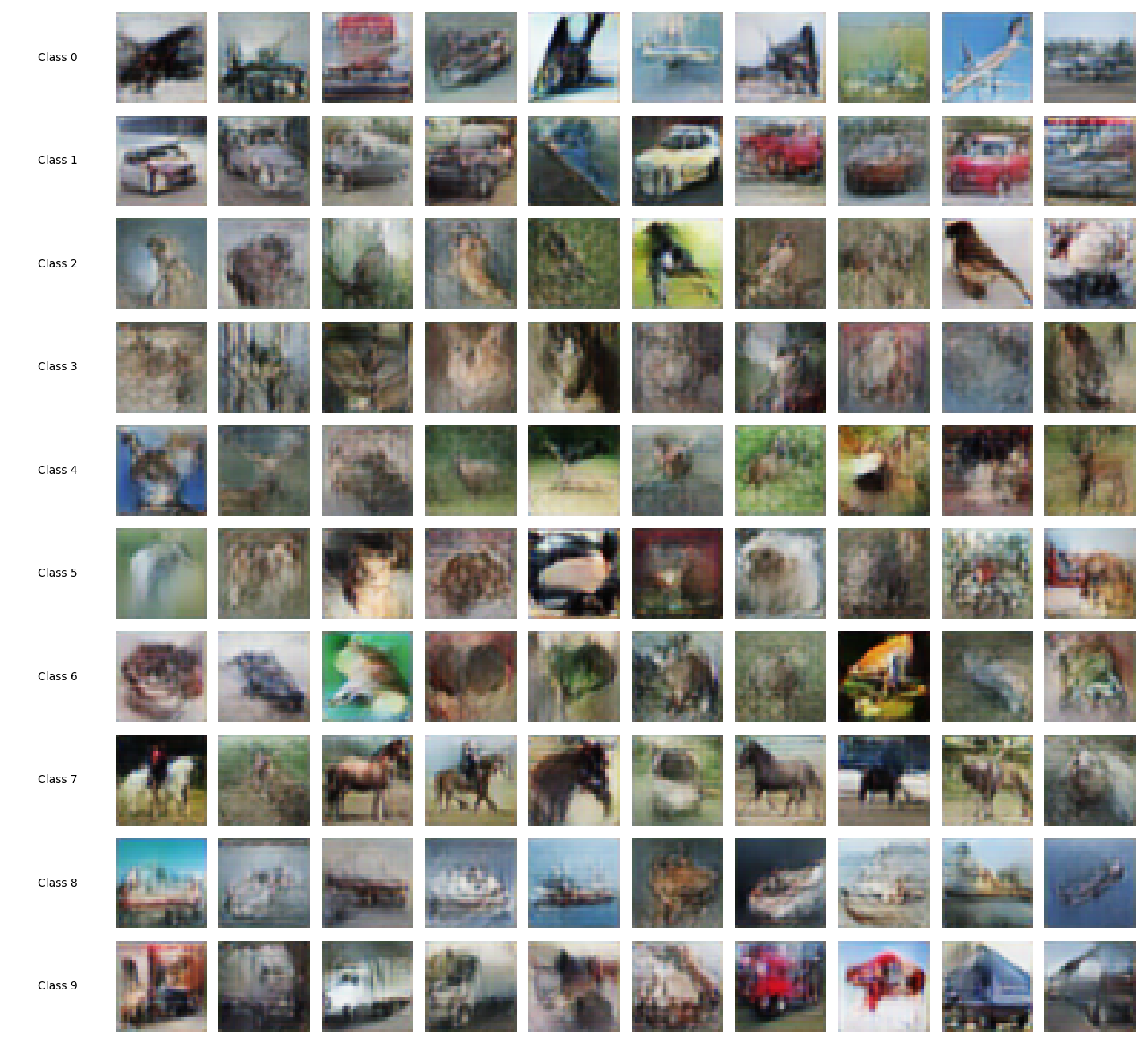}
    \caption{
        Synthetic CIFAR-10 samples generated by sampling latent vectors from a Self-Organizing Map (SOM) trained on VAE latent space. Each latent vector was sampled from the full-covariance Gaussian distribution of a Best Matching Unit (BMU) labeled with the target class. The sampled latents were then decoded into images using VAE. Each row corresponds to one class (0–9).
    }
    \label{fig:som_synthetic_cifar10}
\end{figure}

For CIFAR-10 and CIFAR-100, we combine a VAE and the SOM trained on the VAE's latent space, where, after training, the SOM statistics store the mean, variance, and full covariance matrix of the latent samples. Synthetic latent samples are then generated using eigen-decomposition of the covariance and then decoded using the VAE decoder to reconstruct the images. Figure~\ref{fig:som_synthetic_cifar10} and ~\ref{fig:som_synthetic_cifar100} illustrate some of the samples generated for each class using this method.

\section*{SOM Representations During Continual Learning}
\label{app:som_representations}

\begin{figure*}[t]
    \centering

    \begin{minipage}[t]{0.45\textwidth}
        \centering
        \includegraphics[height=0.23\textheight,keepaspectratio]{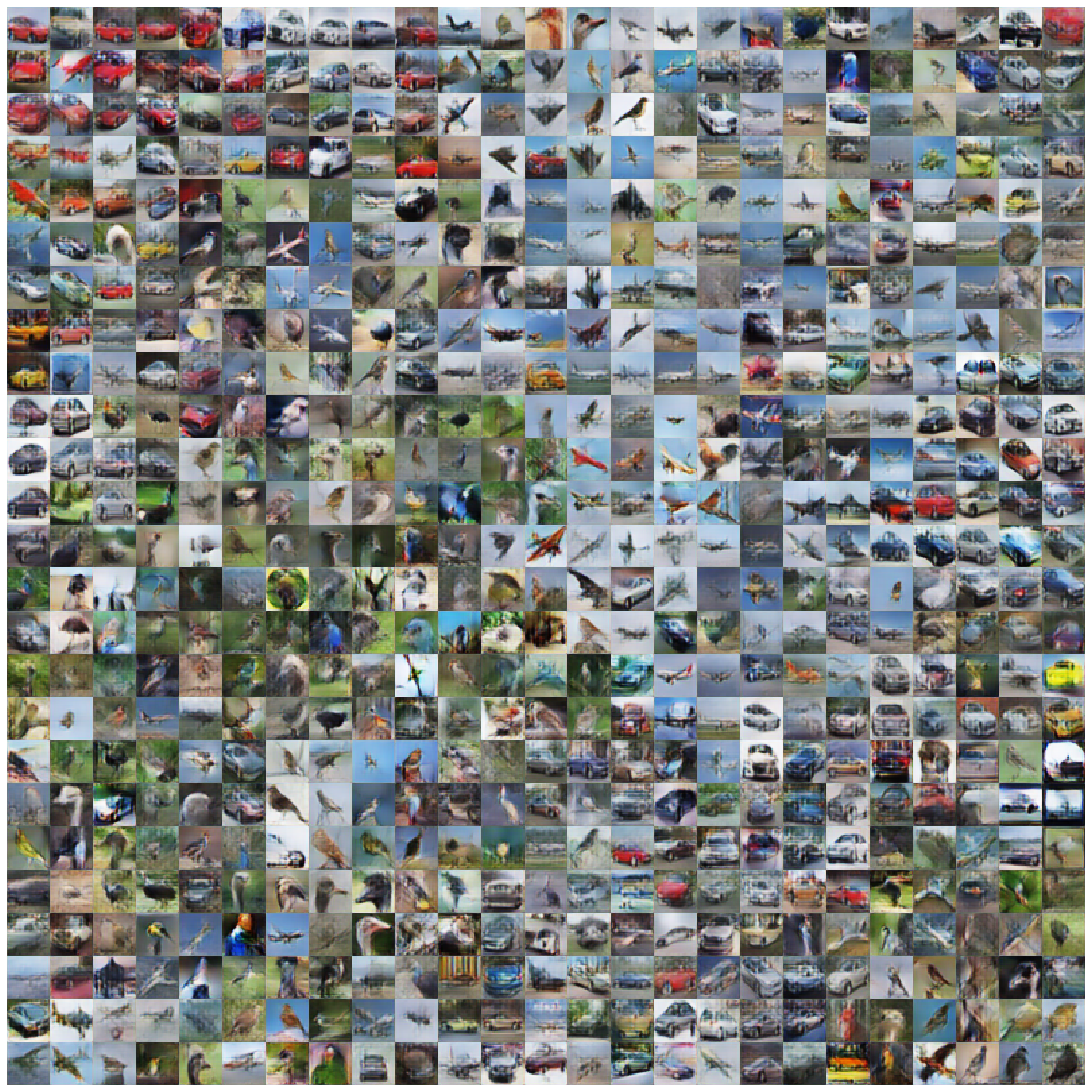}
        \vspace{-1mm}
        \centerline{\footnotesize (a) VAE-FT global}
    \end{minipage}
    \hfill
    \begin{minipage}[t]{0.45\textwidth}
        \centering
        \includegraphics[height=0.23\textheight,keepaspectratio]{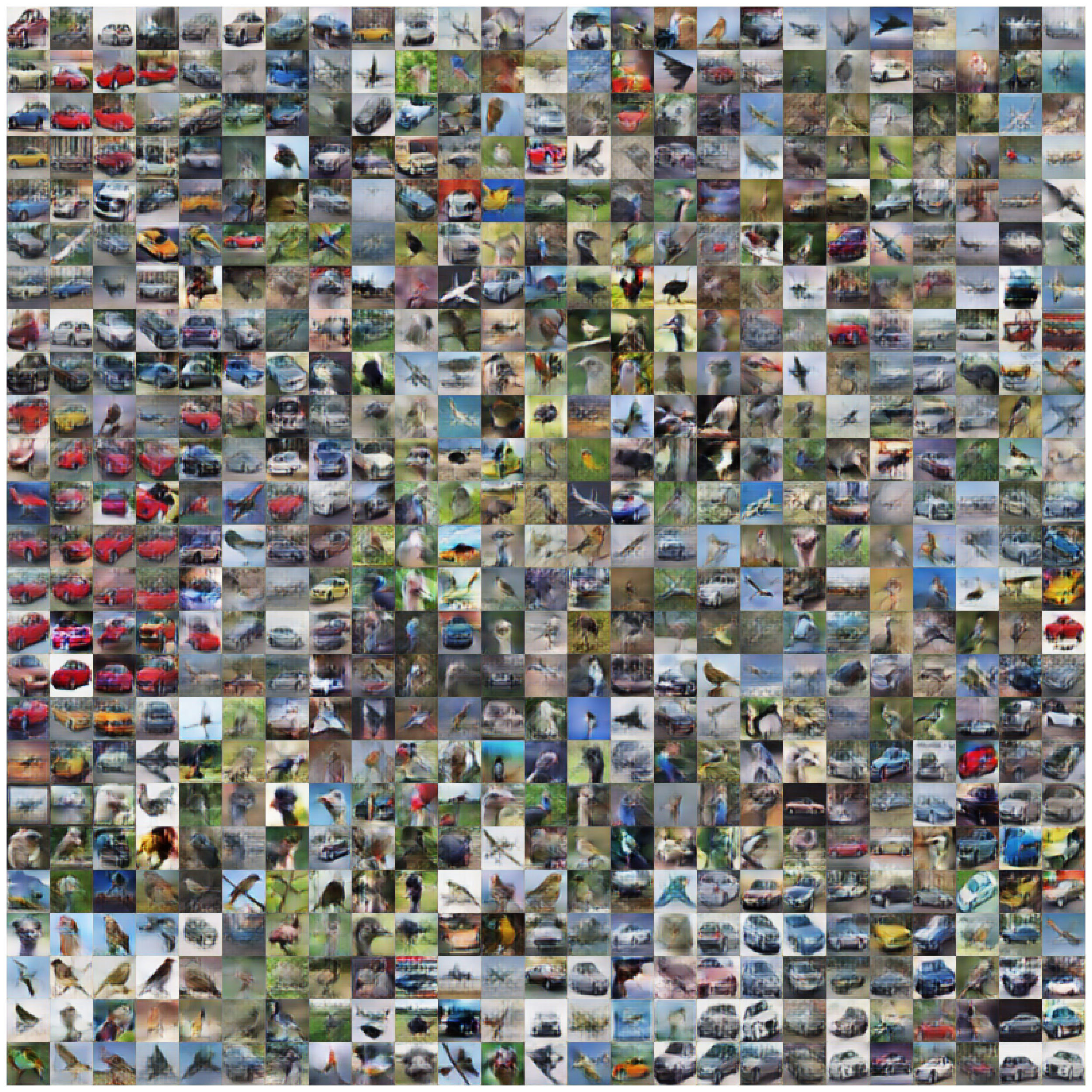}
        \vspace{-1mm}
        \centerline{\footnotesize (b) CLIP-SOM}
    \end{minipage}

    \caption{Visualization of decoded SOM unit vectors from the VAE-FT global model and the CLIP-SOM model after learning the third class on CIFAR-10 using a $25 \times 25$ SOM.}
    \label{fig:som_unit_vectors}
\end{figure*}

Figure~\ref{fig:som_unit_vectors} visualizes the CIFAR-10 SOM unit vectors after data was fed through the decoders of the VAE and CLIP after the third classes. This highlights a strong feature of this methodology -- it is easy to visualize the progress of the algorithm to determine if it is working properly or not, which is beneficial for optimizing hyperparameters or performing early stopping. Appendix~\ref{app:som_representations} provides a full presentation of SOM unit vector visualizations after each class (or for every 10 classes, in the case of CIFAR-100) for each dataset.

Across datasets, we observe that the SOMs initially form a well-distributed representation of the initial class. After training on the next class, part of the map is overwritten by the new class, while some regions preserve some information from previous classes, which were synthetically generated from the prior SOM iteration. As new tasks are learned, this process continues and representations for prior classes become more compressed in the map; by the time the final class is introduced, the representation is significantly shifted. However, distinct clusters of previously-learned classes remain in various regions for all datasets, showing that our method maintains earlier class representations.

\begin{figure*}[!t]
    \centering
    \includegraphics[width=0.32\textwidth]{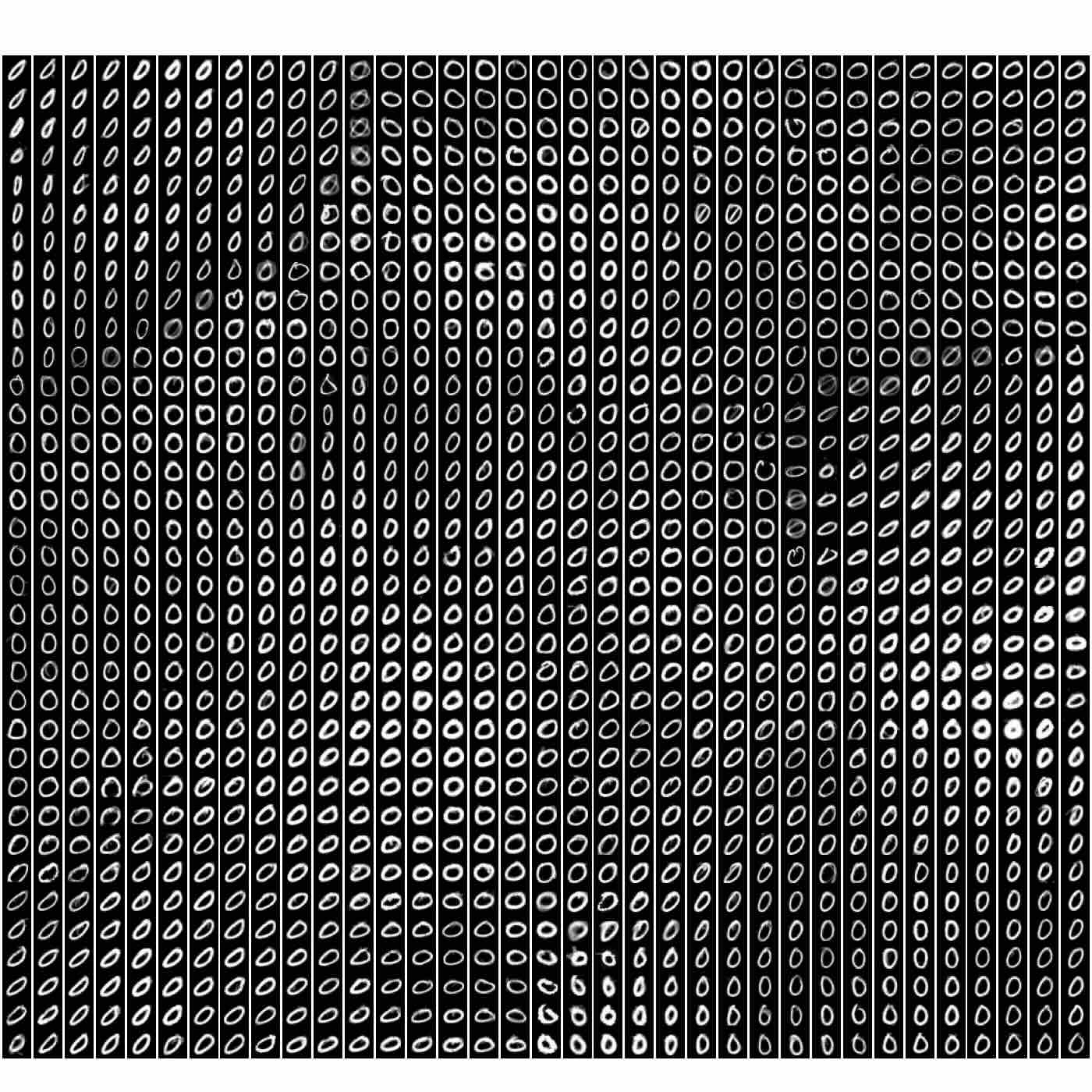}
    \includegraphics[width=0.32\textwidth]{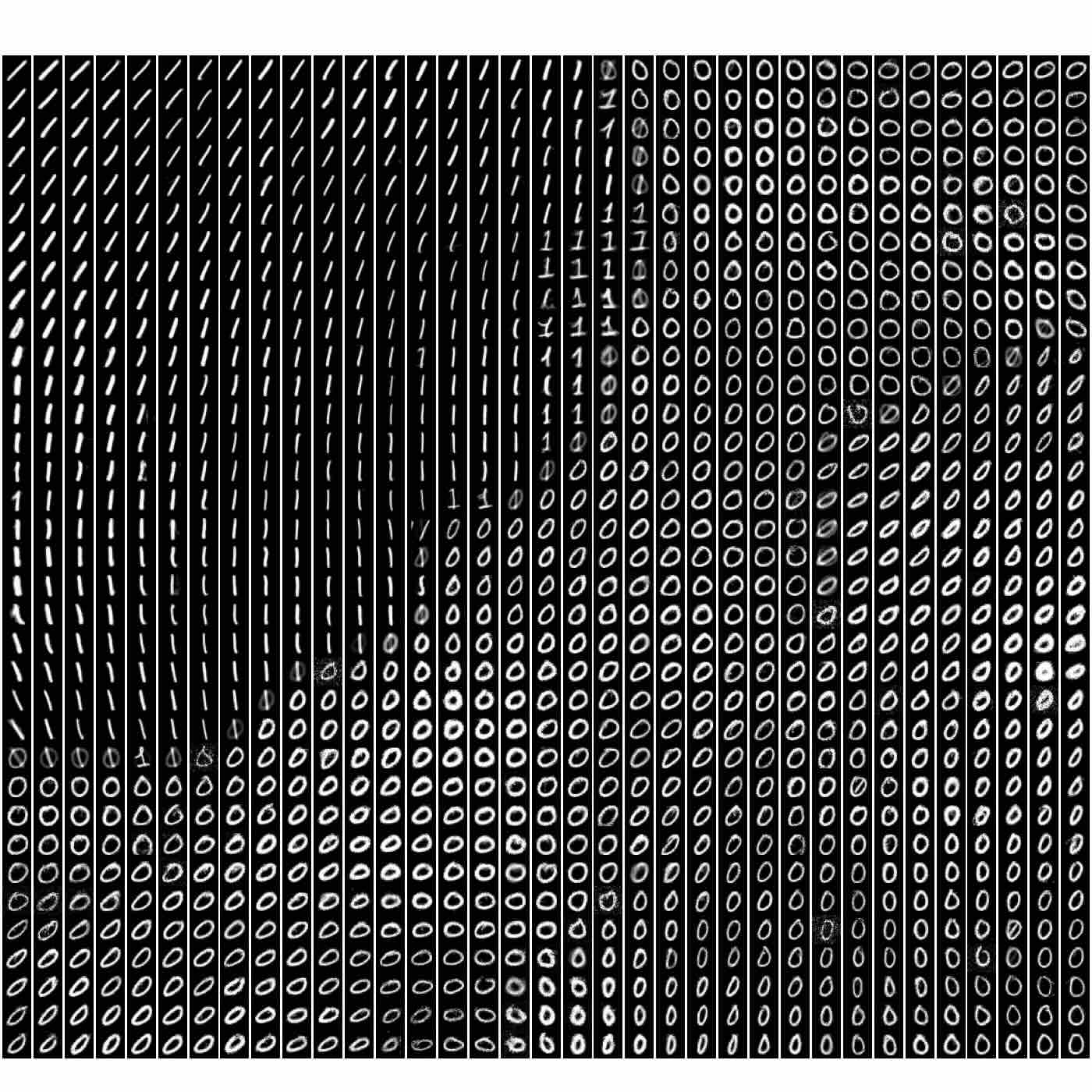}
    \includegraphics[width=0.32\textwidth]{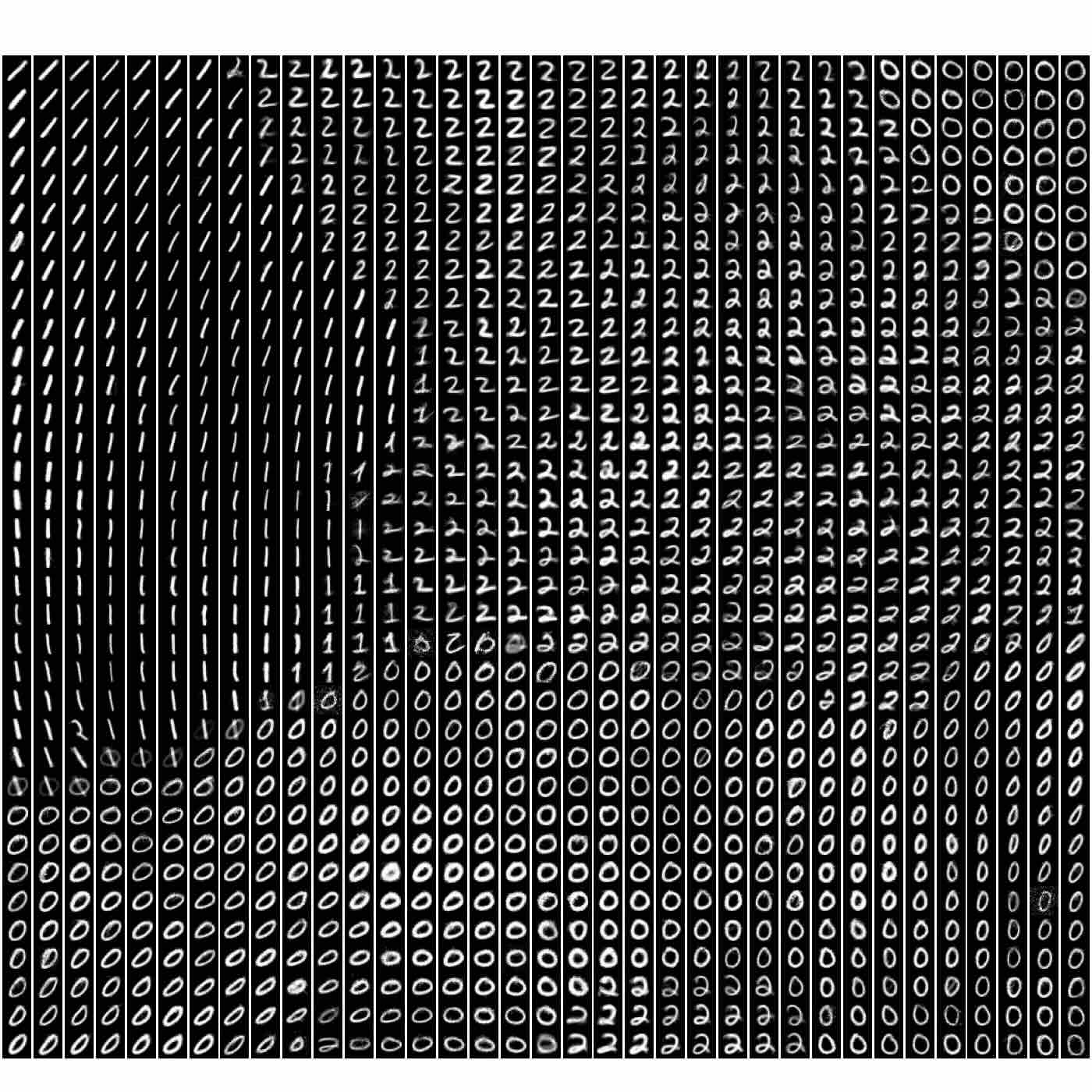} \\[1mm]

    \includegraphics[width=0.32\textwidth]{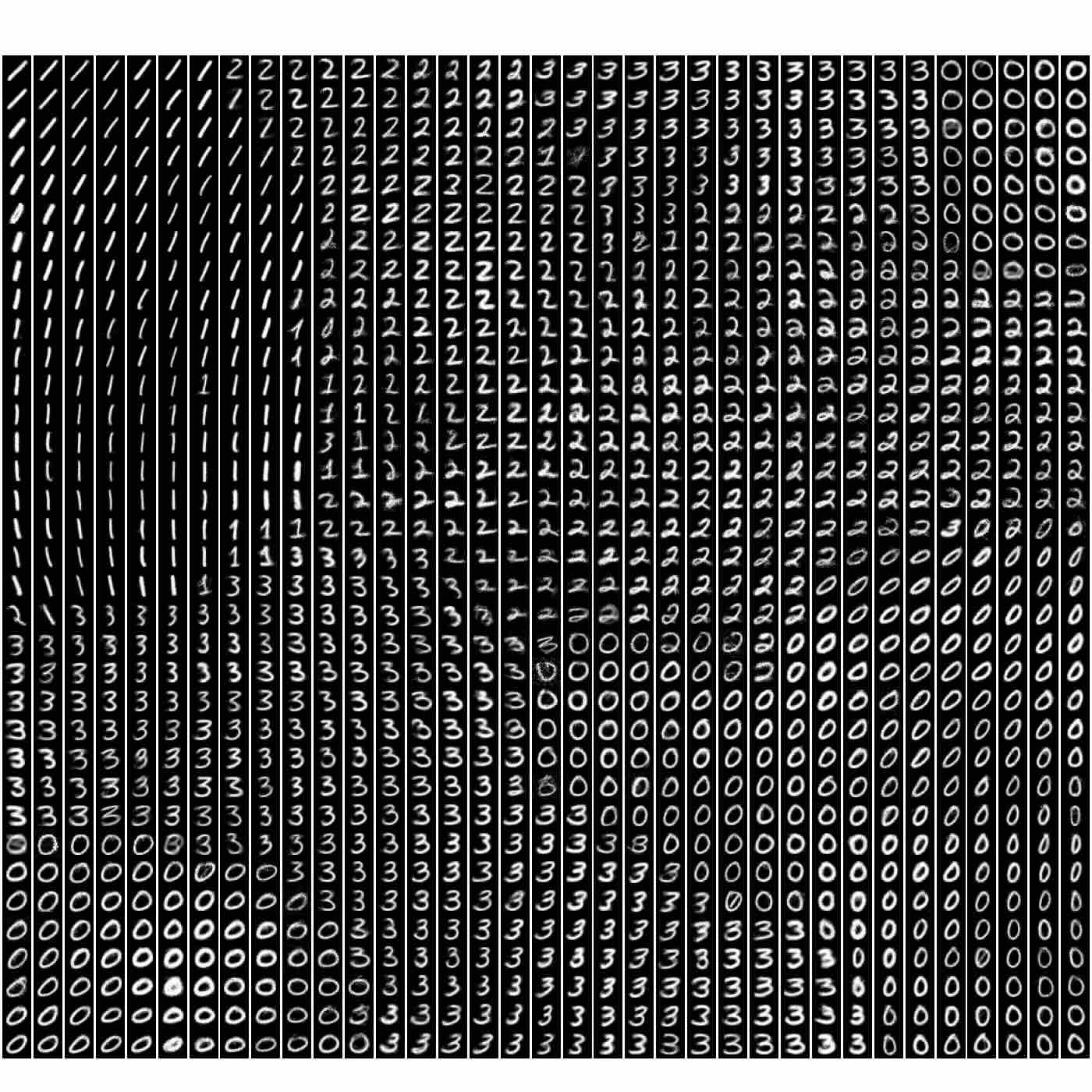}
    \includegraphics[width=0.32\textwidth]{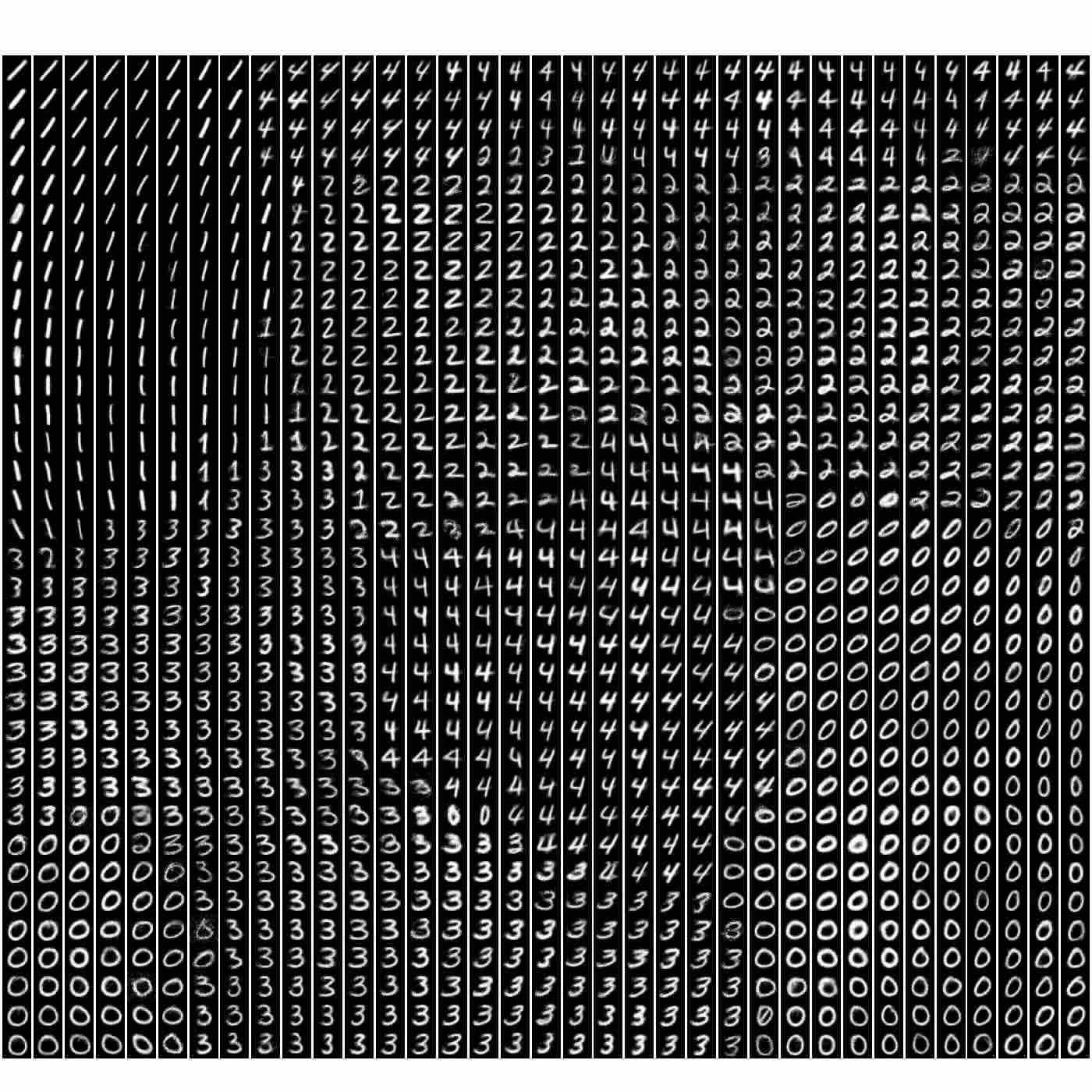}
    \includegraphics[width=0.32\textwidth]{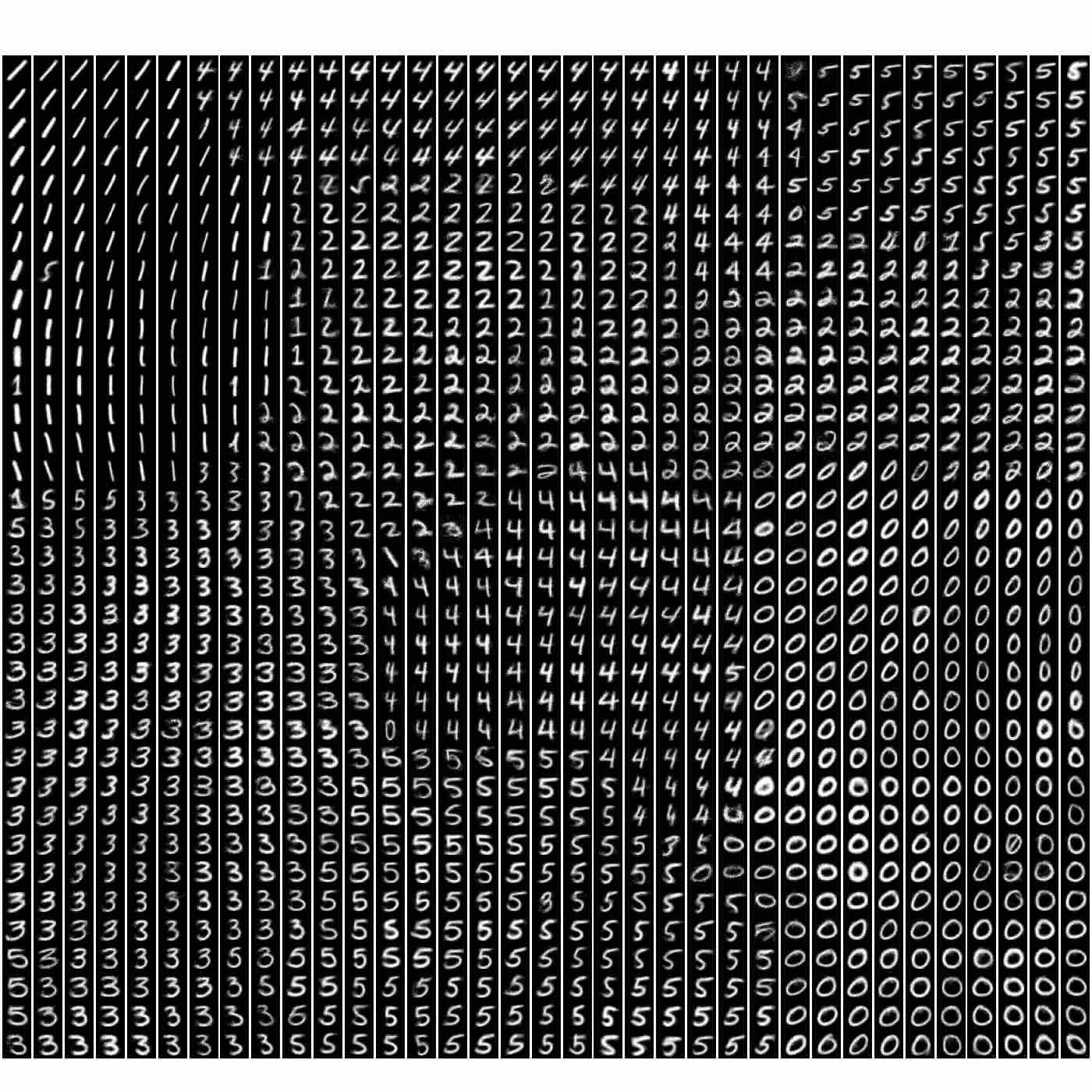} \\[1mm]

    \includegraphics[width=0.32\textwidth]{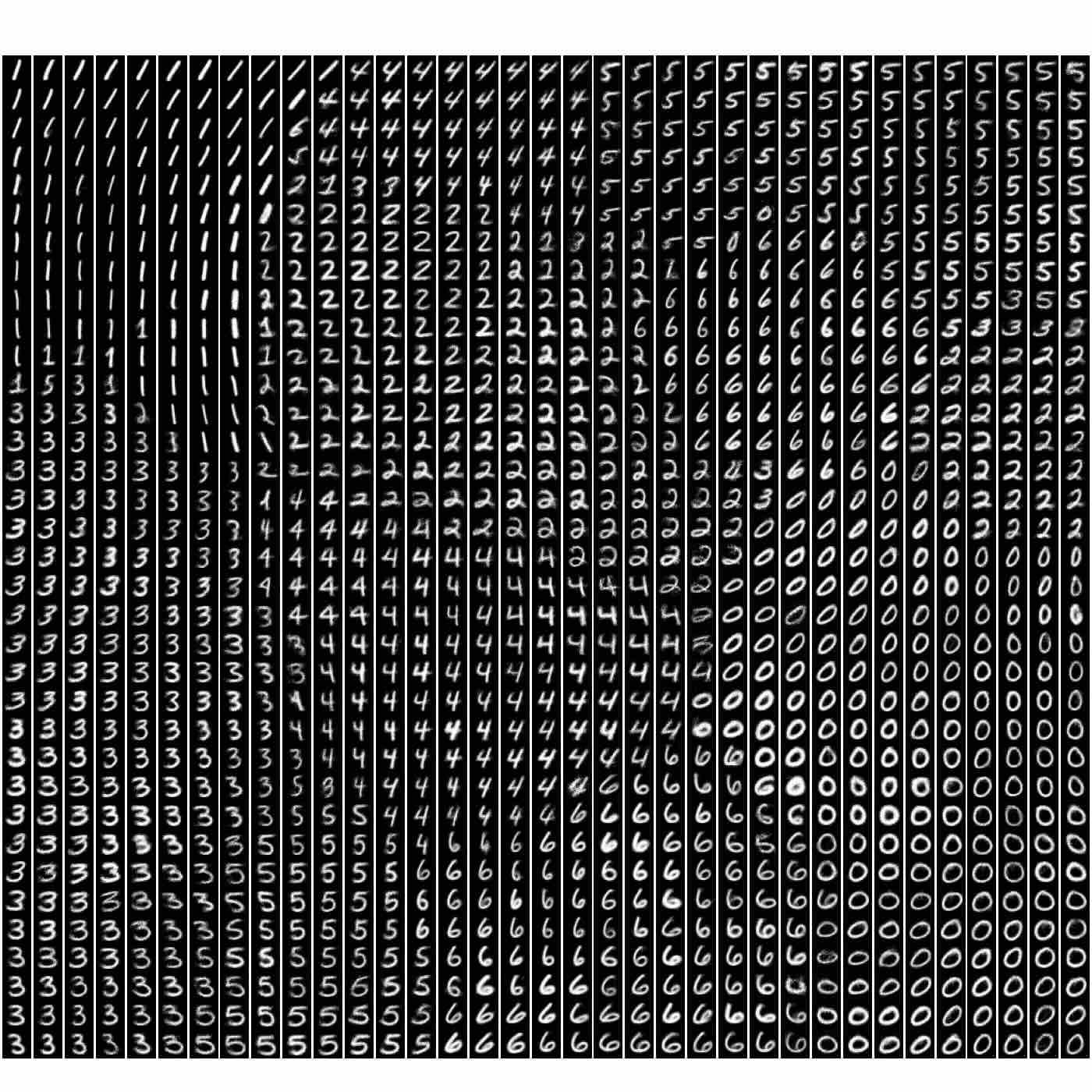}
    \includegraphics[width=0.32\textwidth]{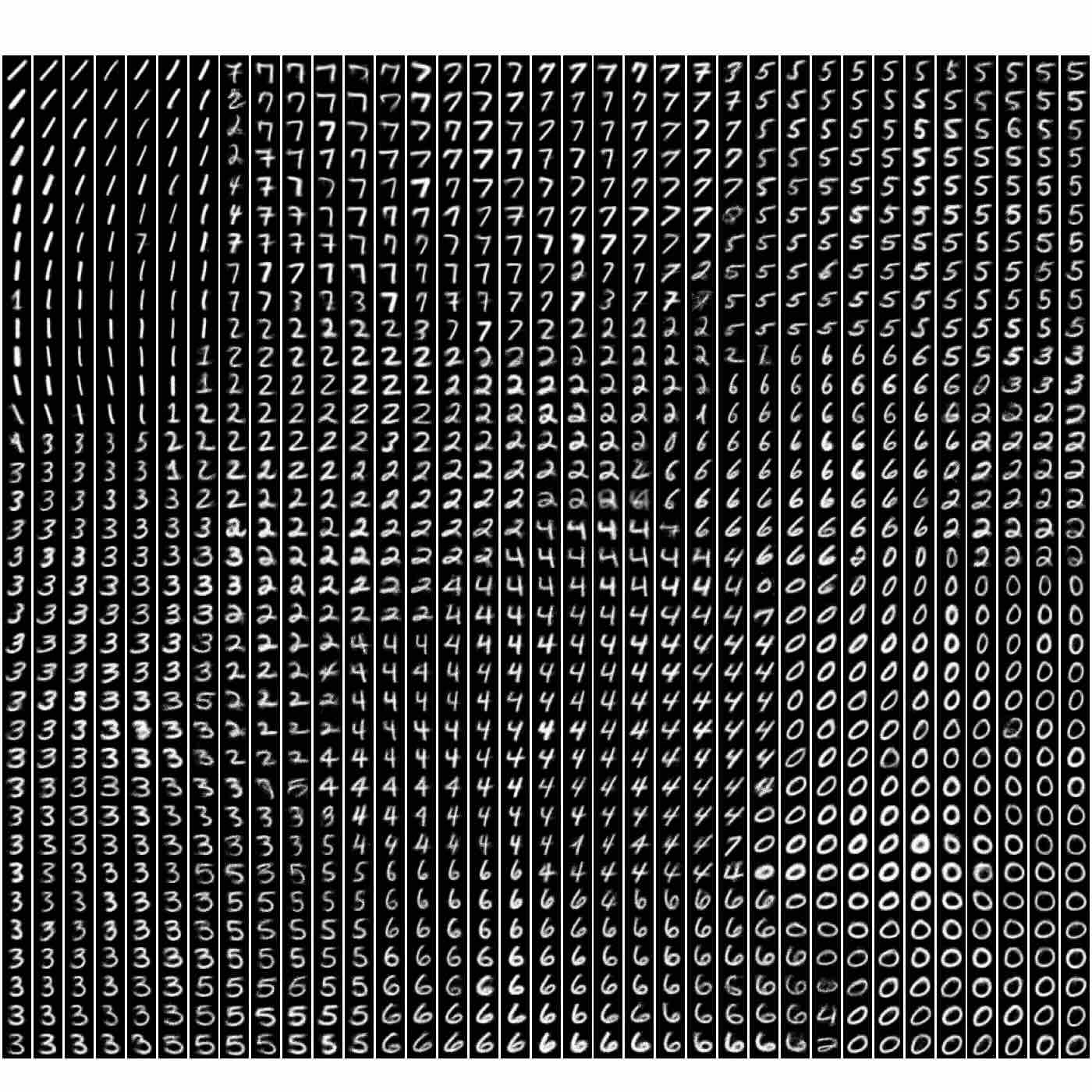}
    \includegraphics[width=0.32\textwidth]{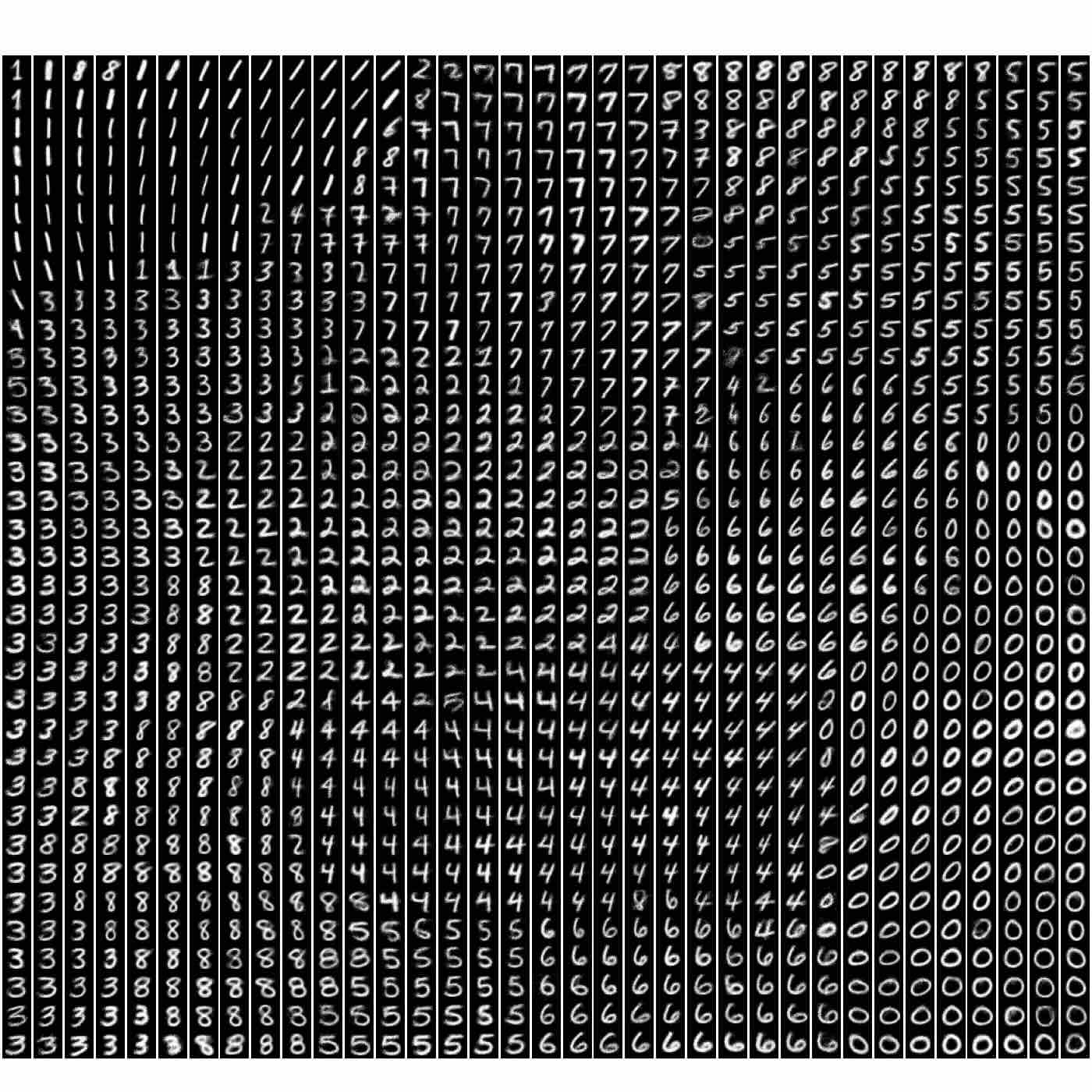} \\[1mm]

    \includegraphics[width=0.32\textwidth]{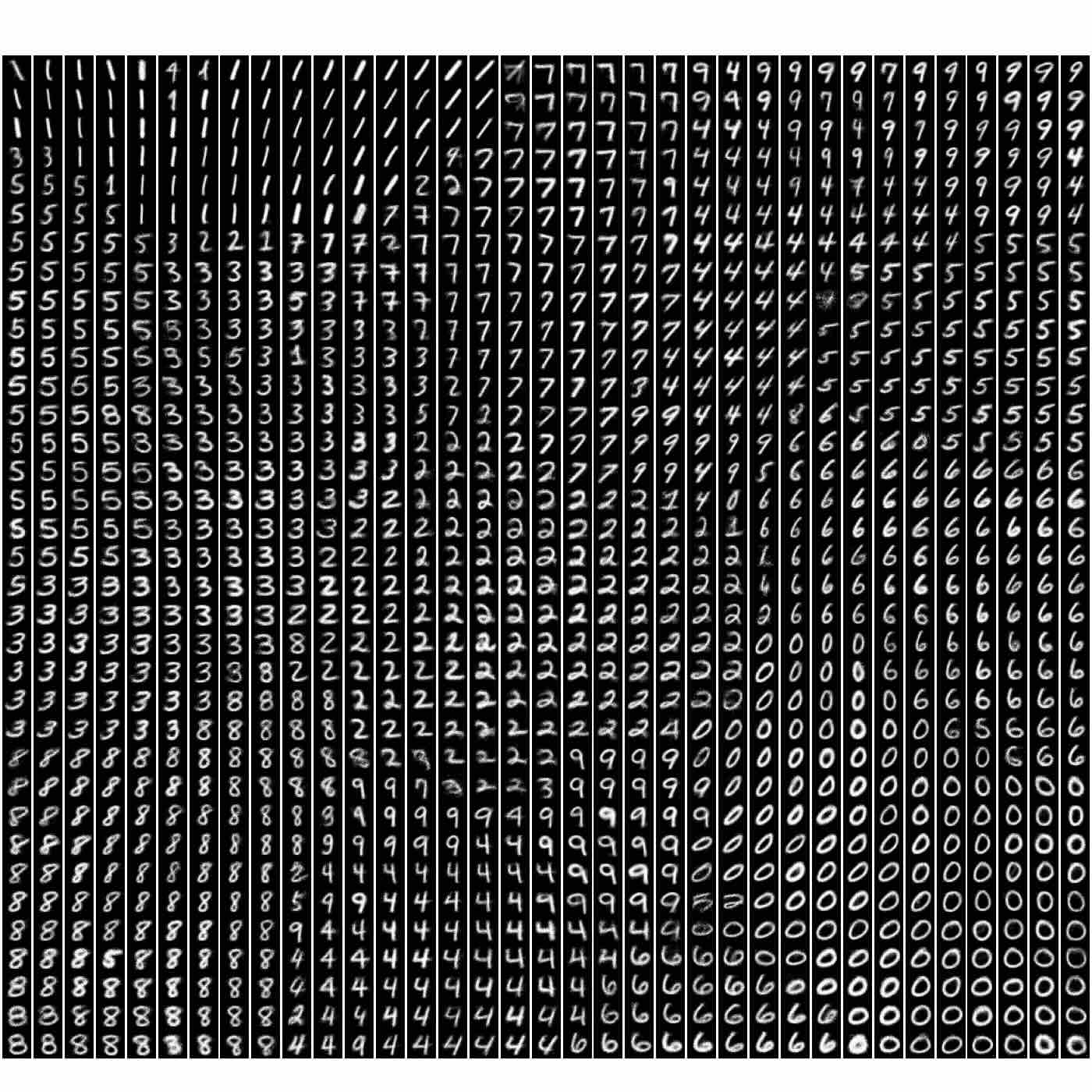}

    \caption{Visualization of the SOM unit vectors after each task for MNIST (35×35 SOM).}
    \label{fig:appendix_mnist_gen}
\end{figure*}

\begin{figure*}[!t]
    \centering
    \includegraphics[width=0.32\textwidth]{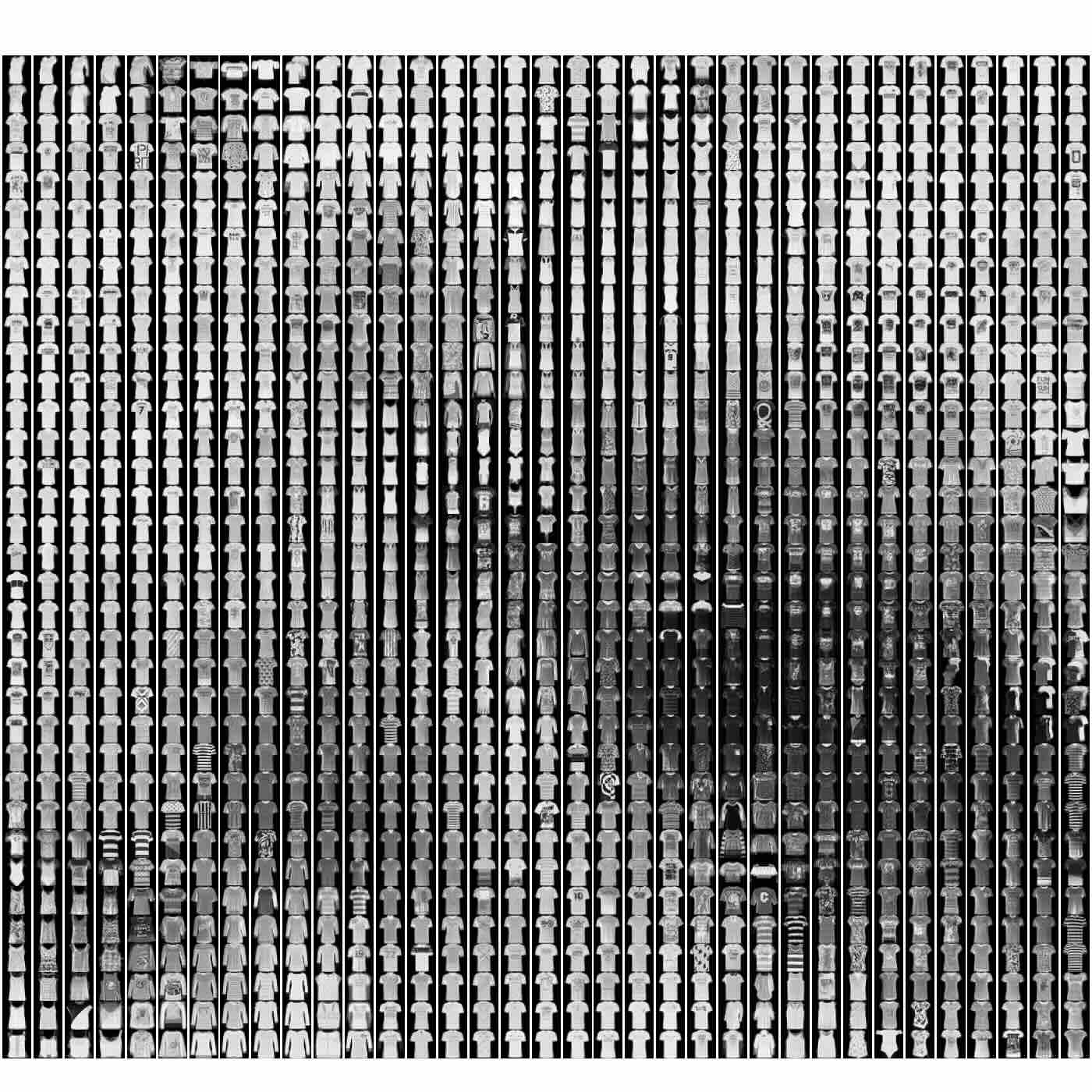}
    \includegraphics[width=0.32\textwidth]{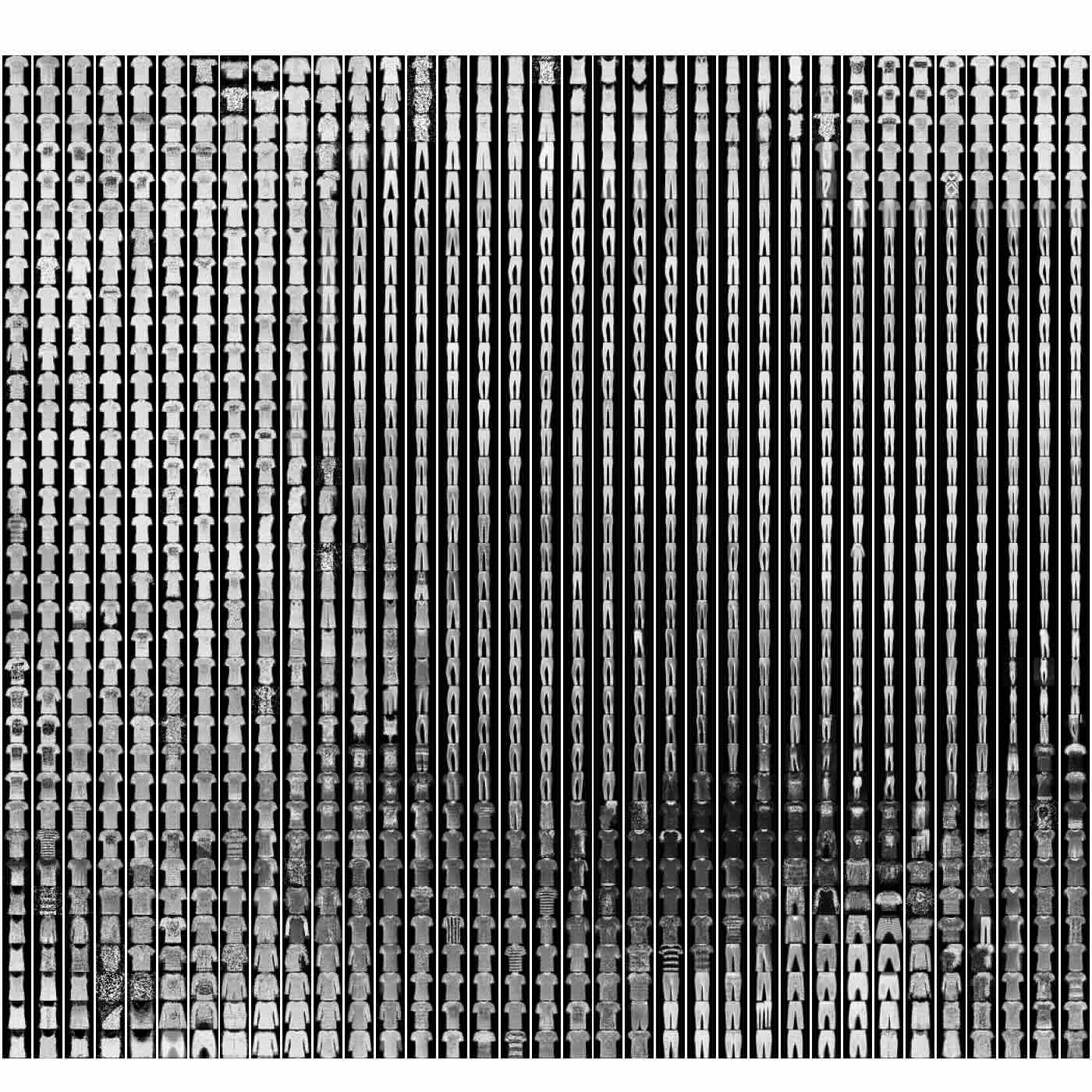}
    \includegraphics[width=0.32\textwidth]{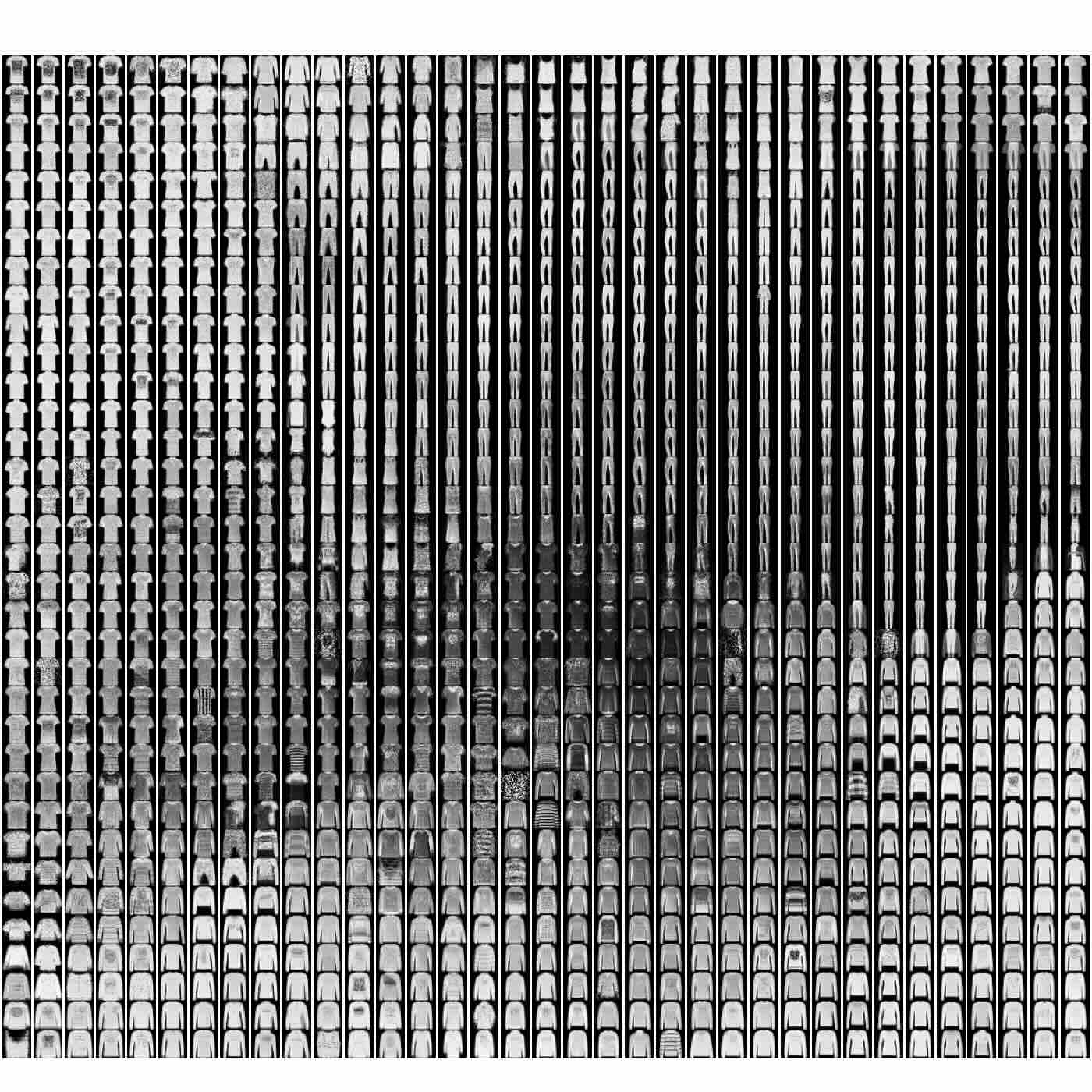} \\[1mm]

    \includegraphics[width=0.32\textwidth]{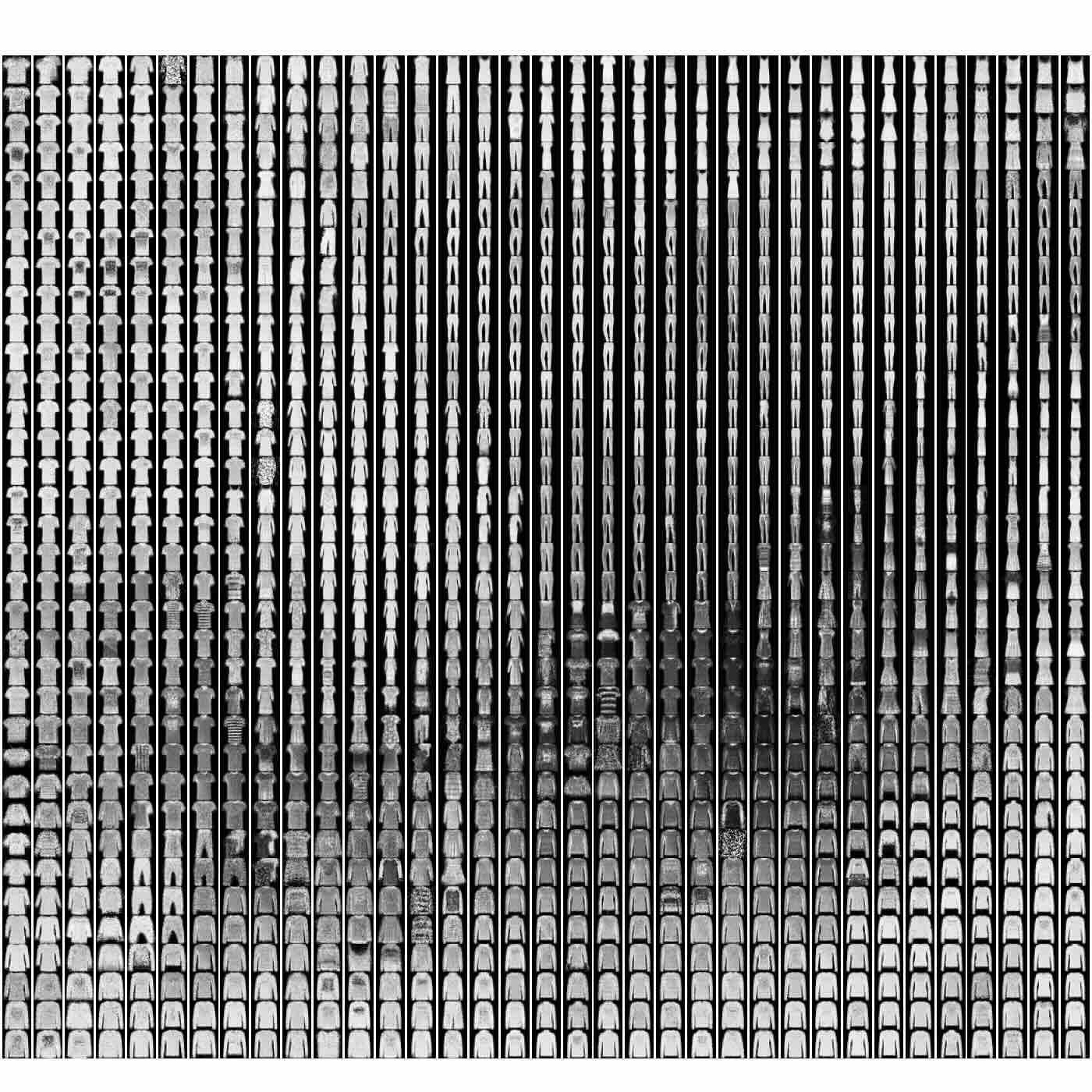}
    \includegraphics[width=0.32\textwidth]{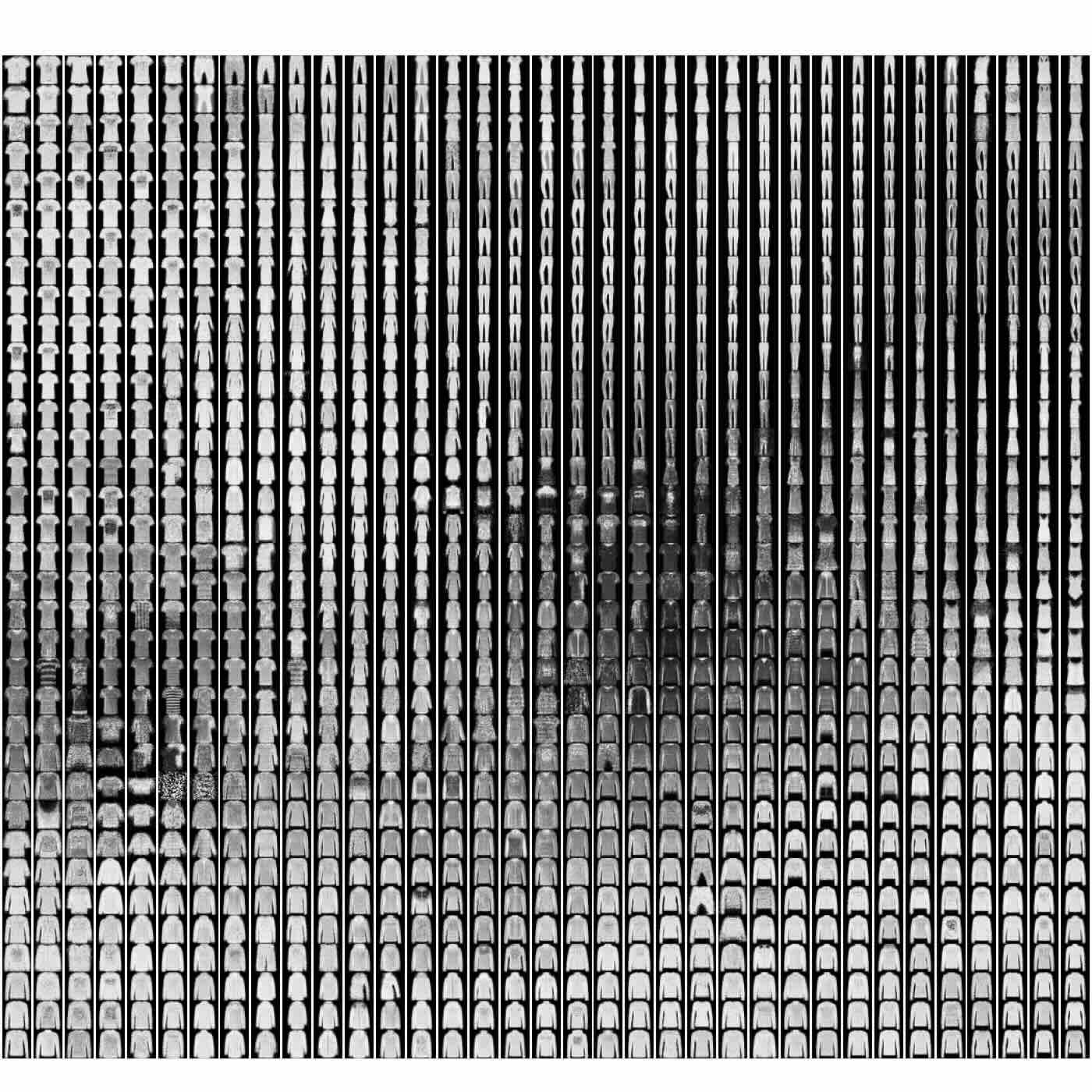}
    \includegraphics[width=0.32\textwidth]{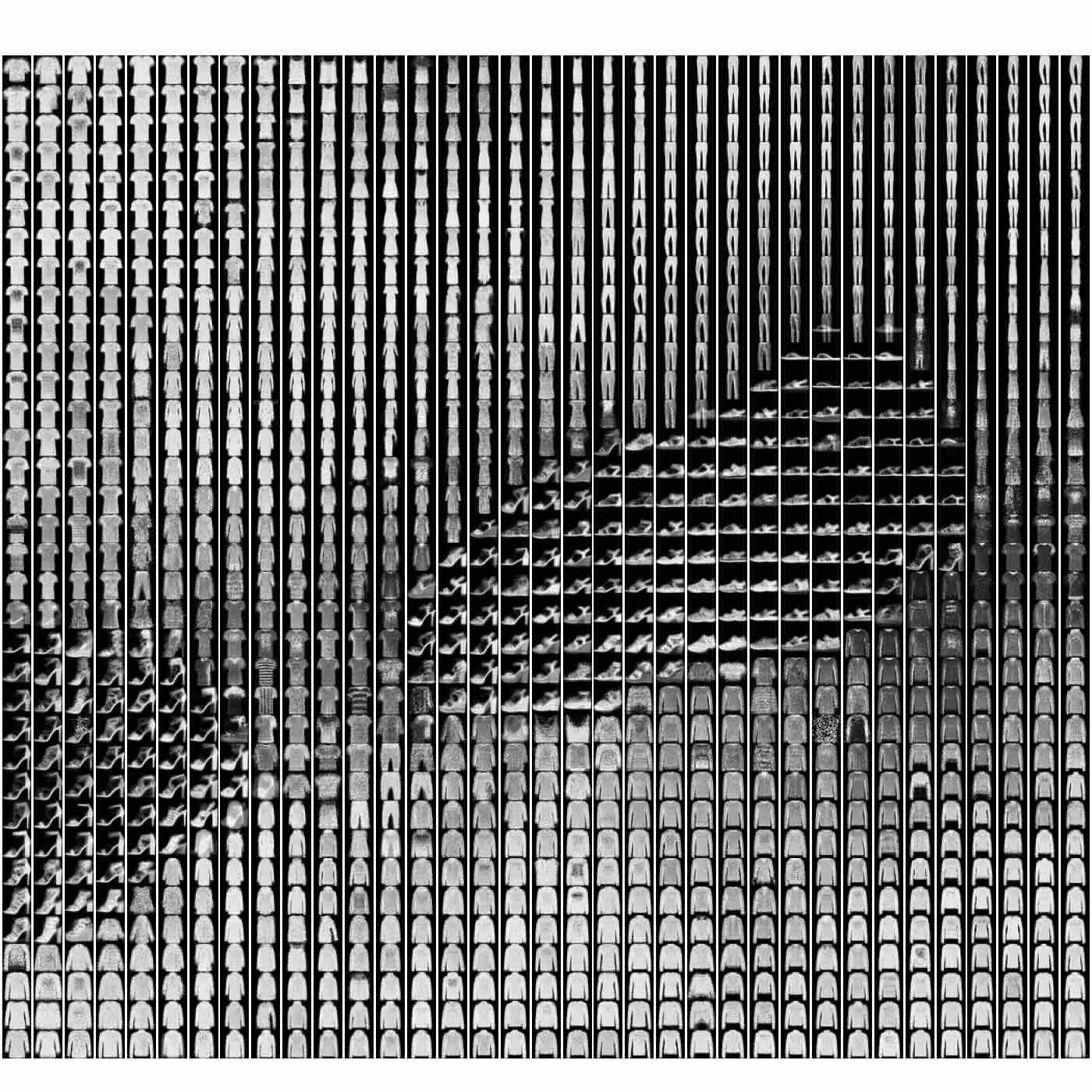} \\[1mm]

    \includegraphics[width=0.32\textwidth]{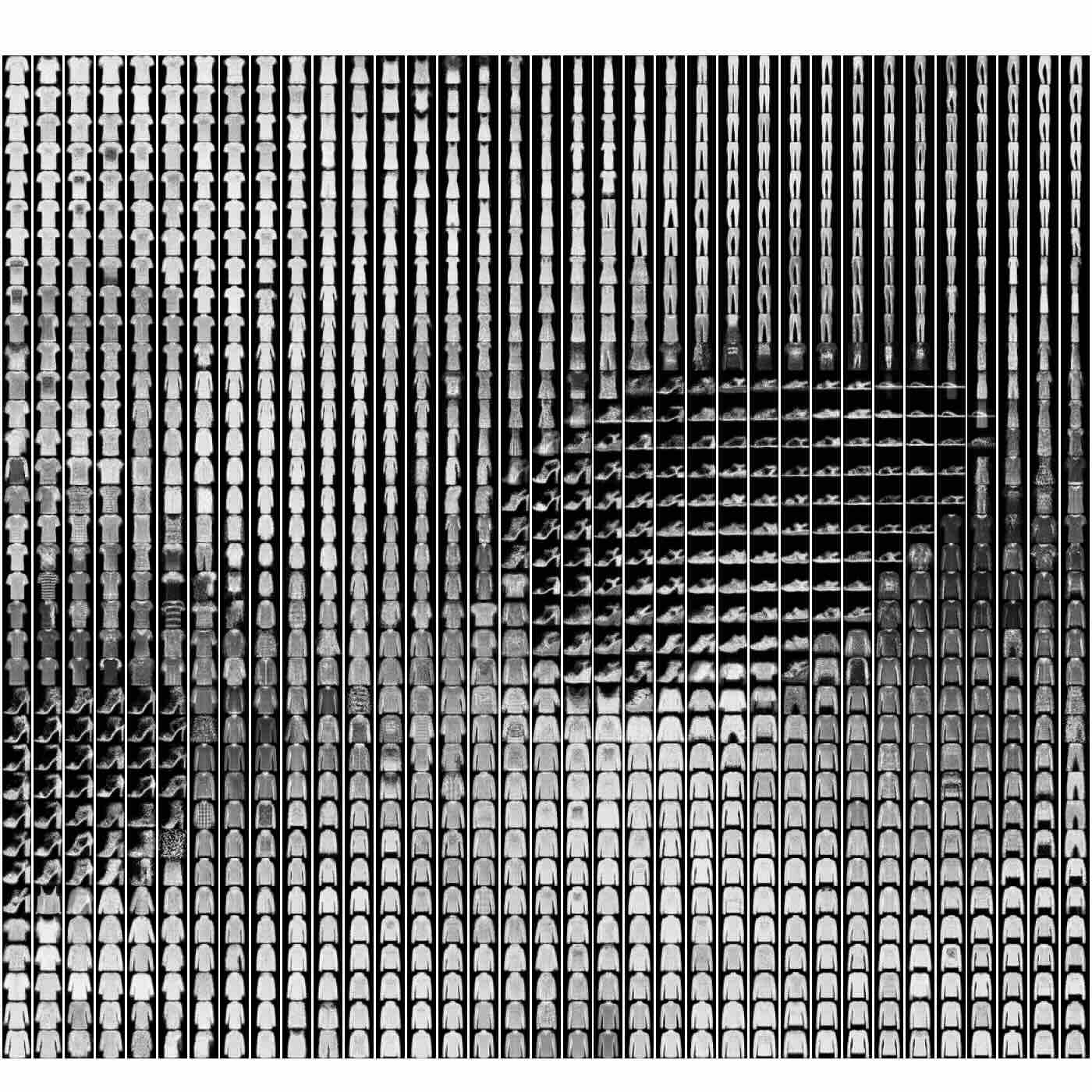}
    \includegraphics[width=0.32\textwidth]{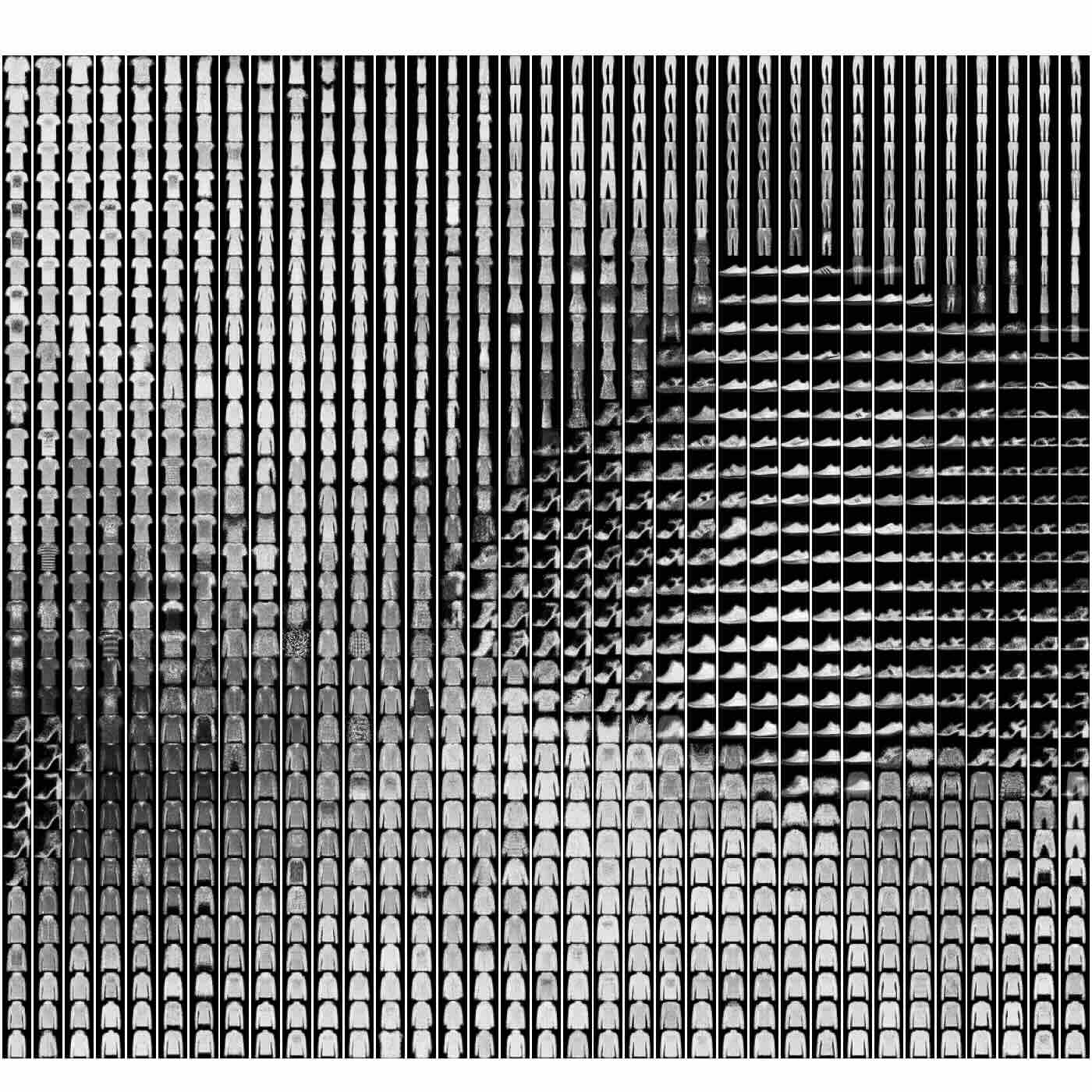}
    \includegraphics[width=0.32\textwidth]{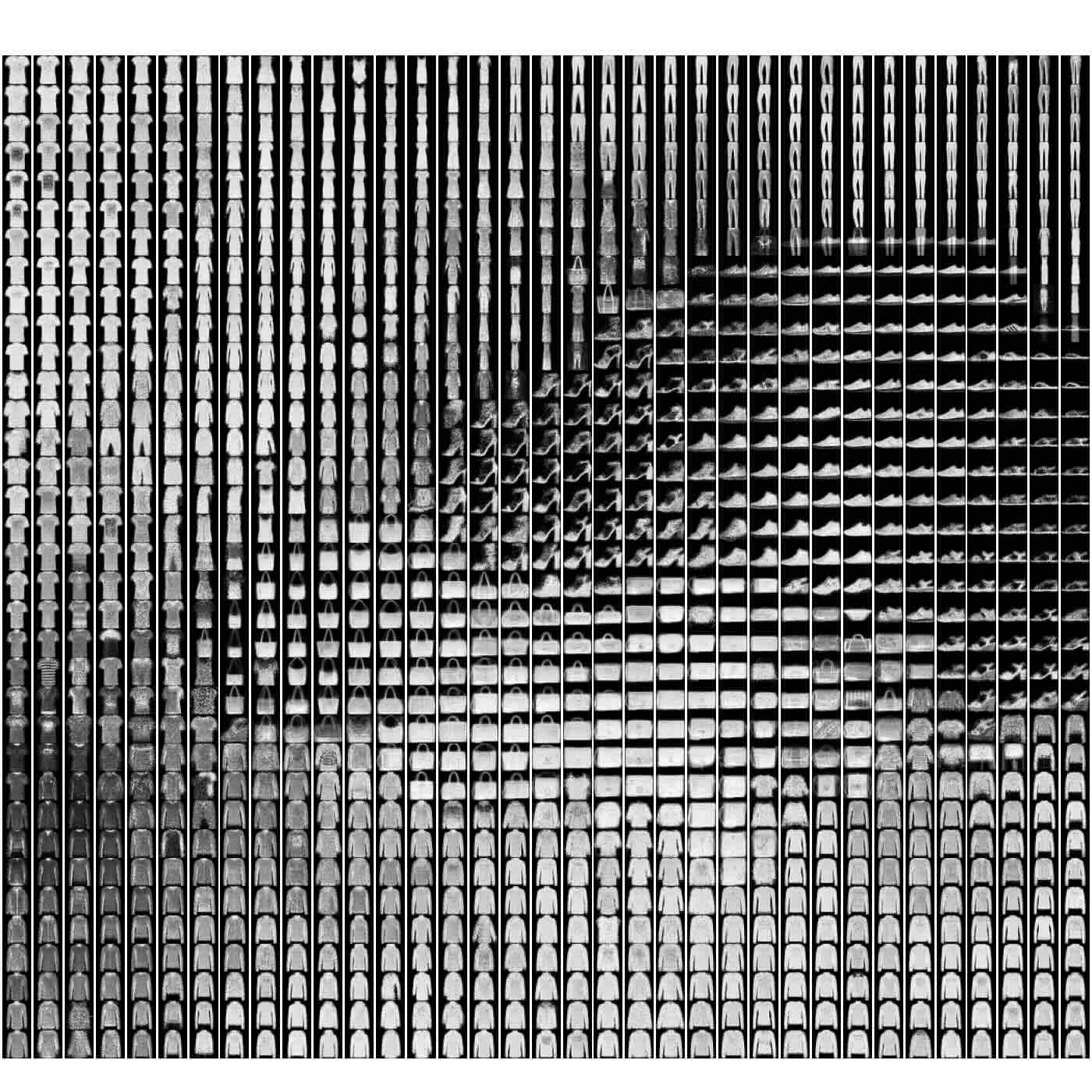} \\[1mm]

    \includegraphics[width=0.32\textwidth]{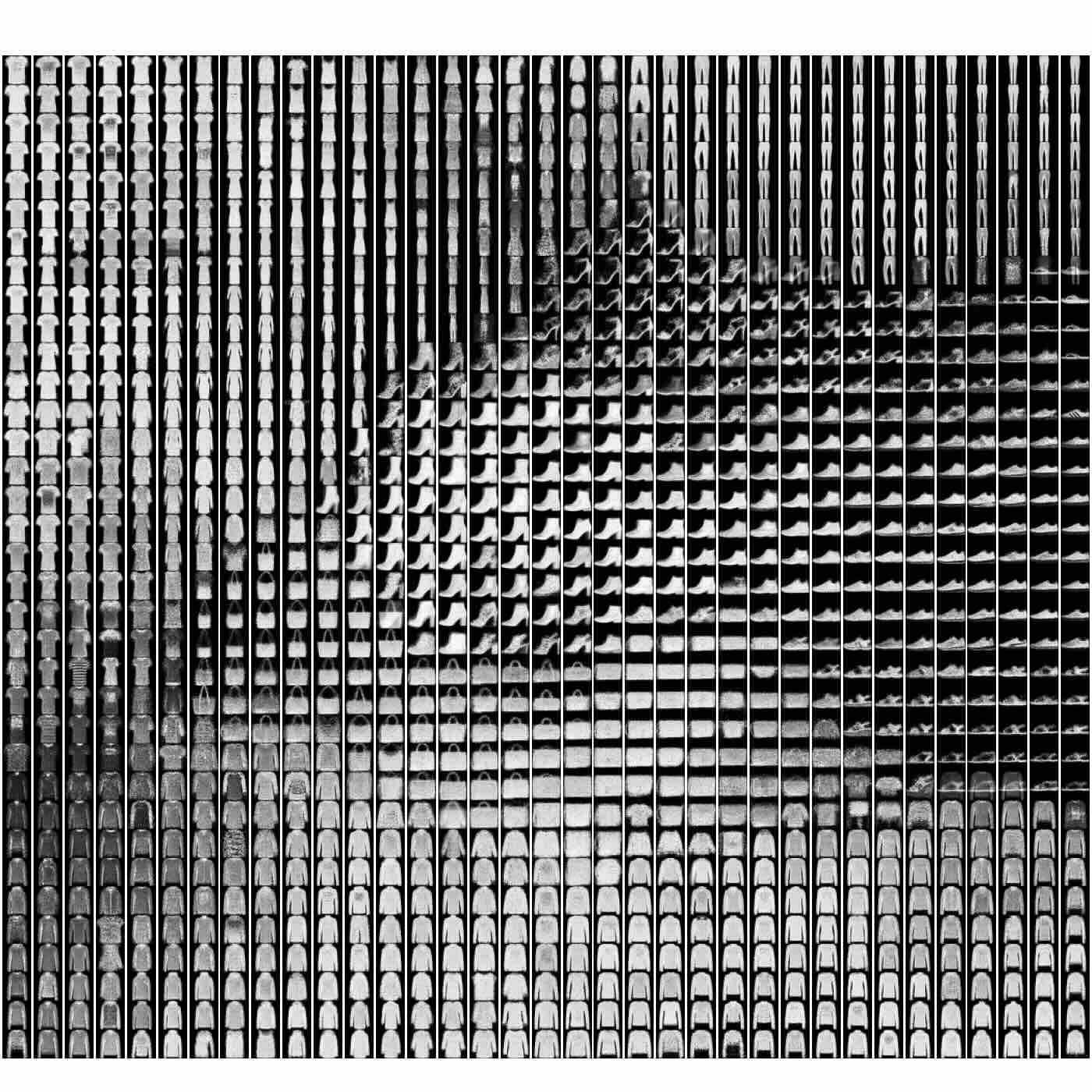}

    \caption{Visualization of SOM unit vectors after each task for Fashion-MNIST (35×35 SOM).}
    \label{fig:appendix_fmnist_gen}
\end{figure*}

\begin{figure*}[!t]
    \centering
    \includegraphics[width=0.32\textwidth]{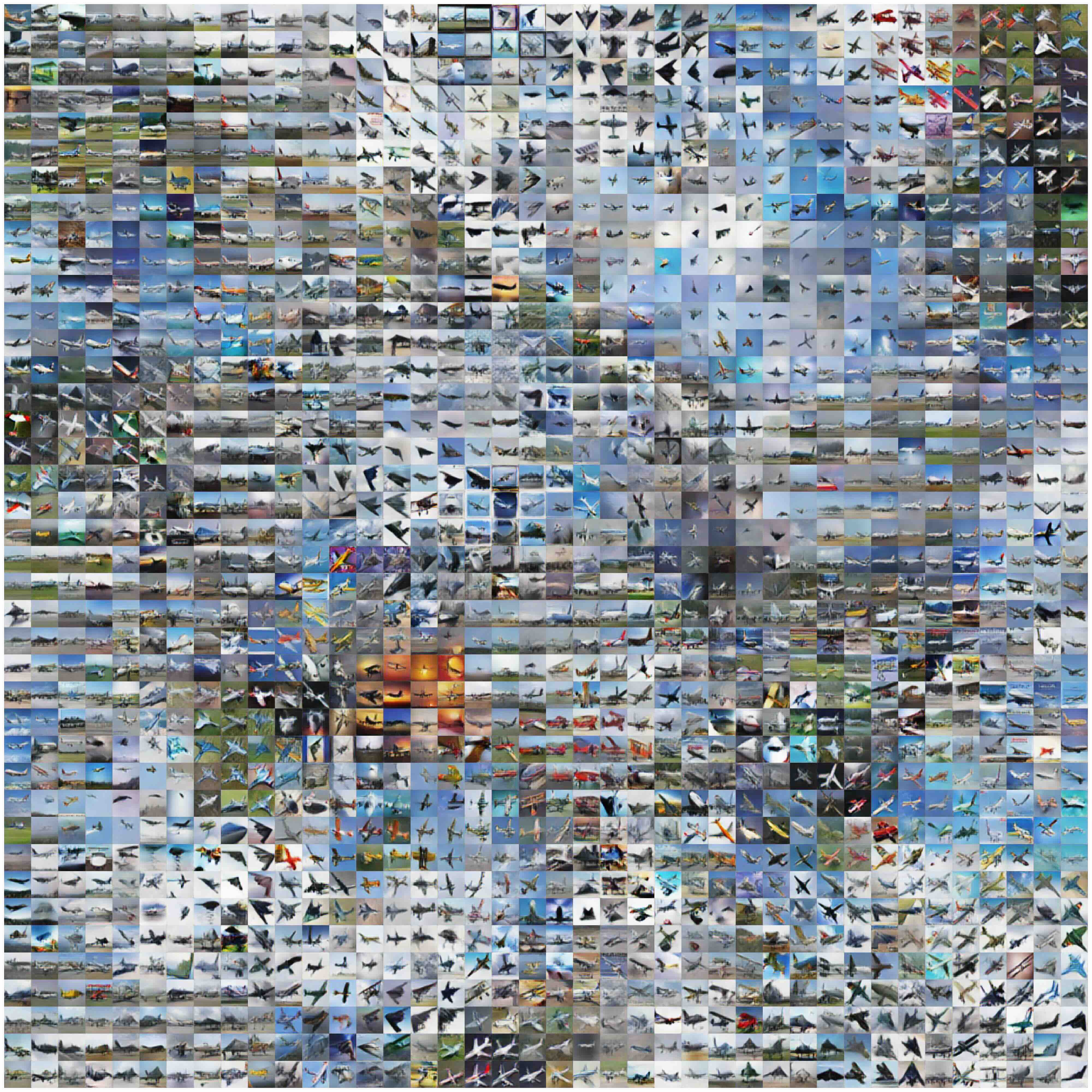}
    \includegraphics[width=0.32\textwidth]{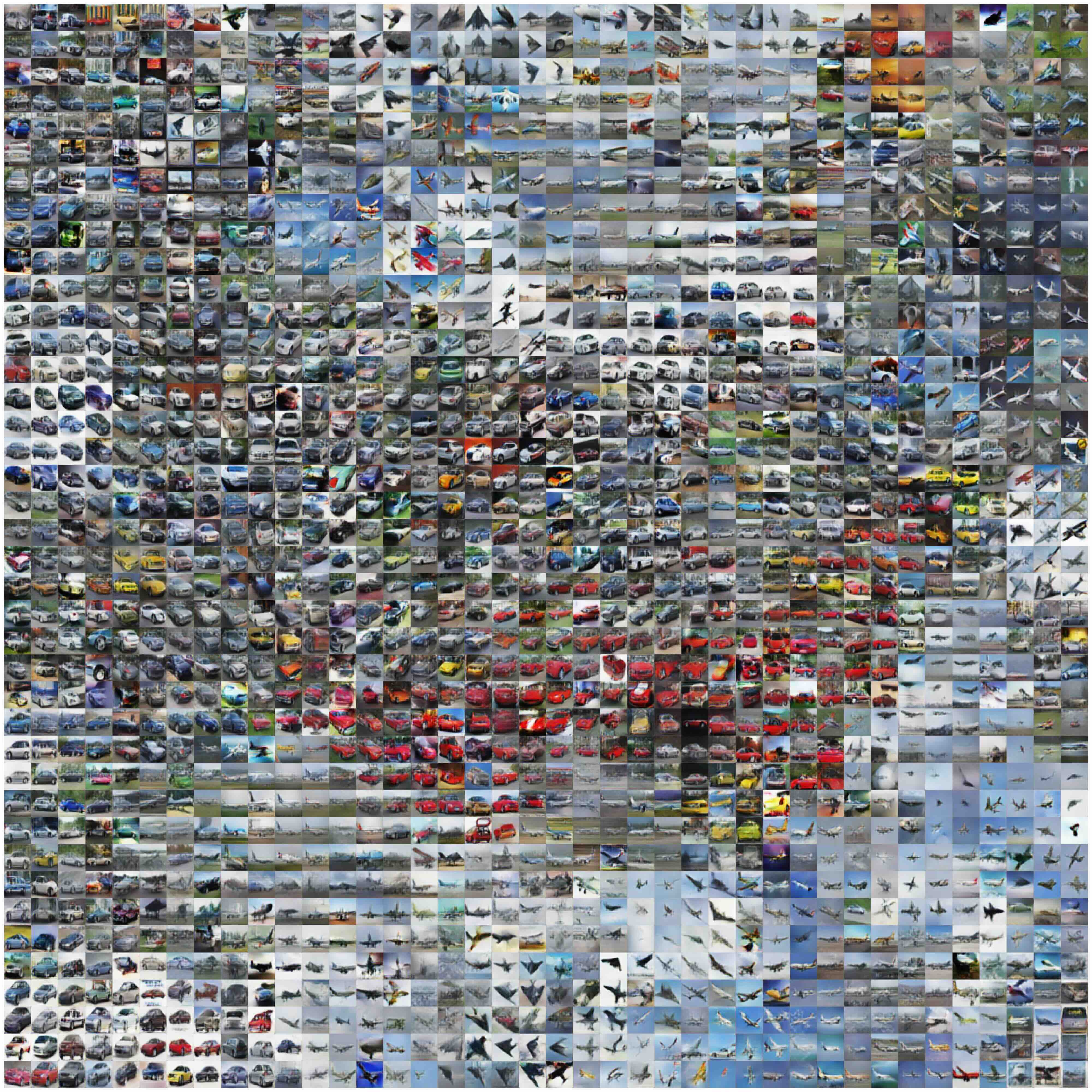}
    \includegraphics[width=0.32\textwidth]{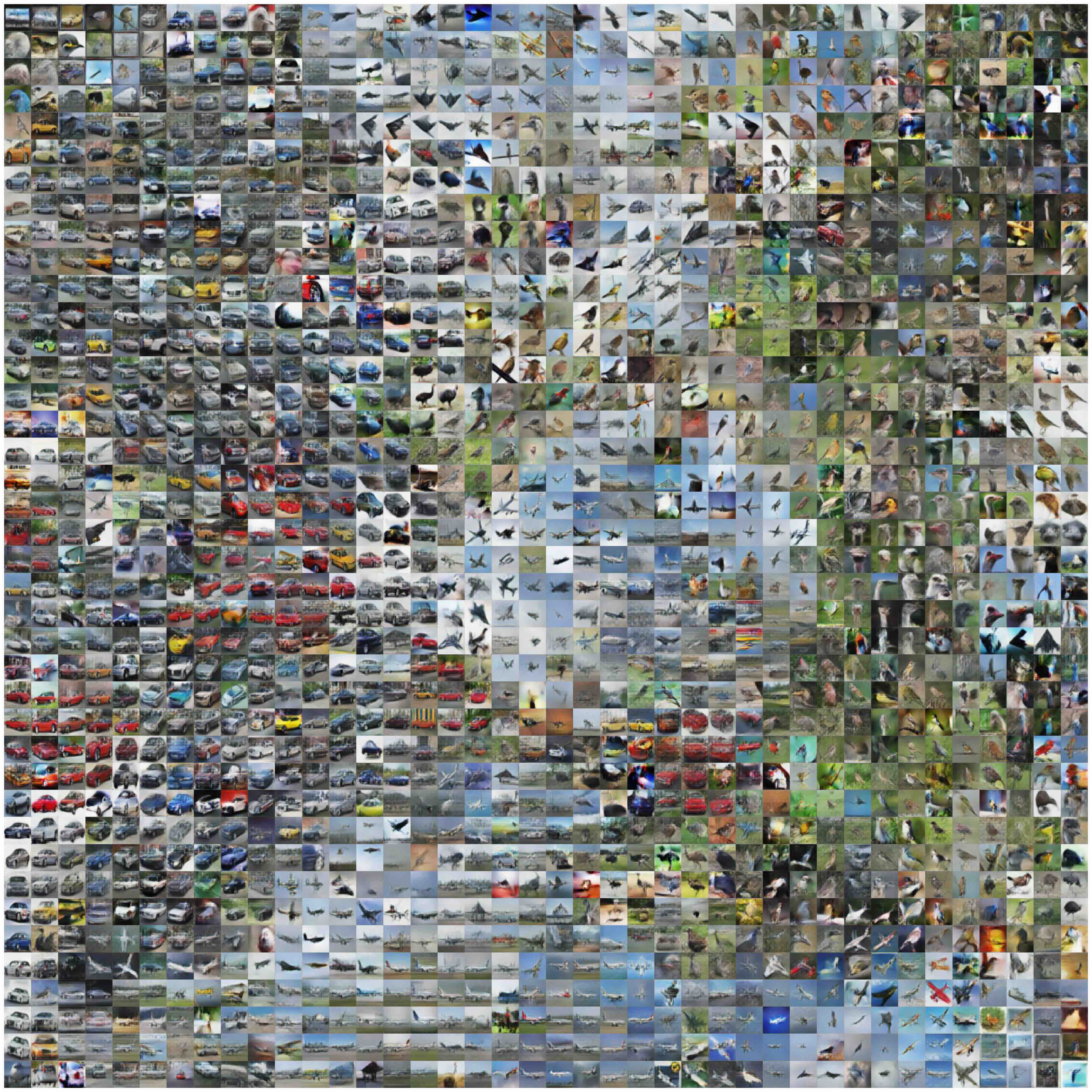} \\[1mm]

    \includegraphics[width=0.32\textwidth]{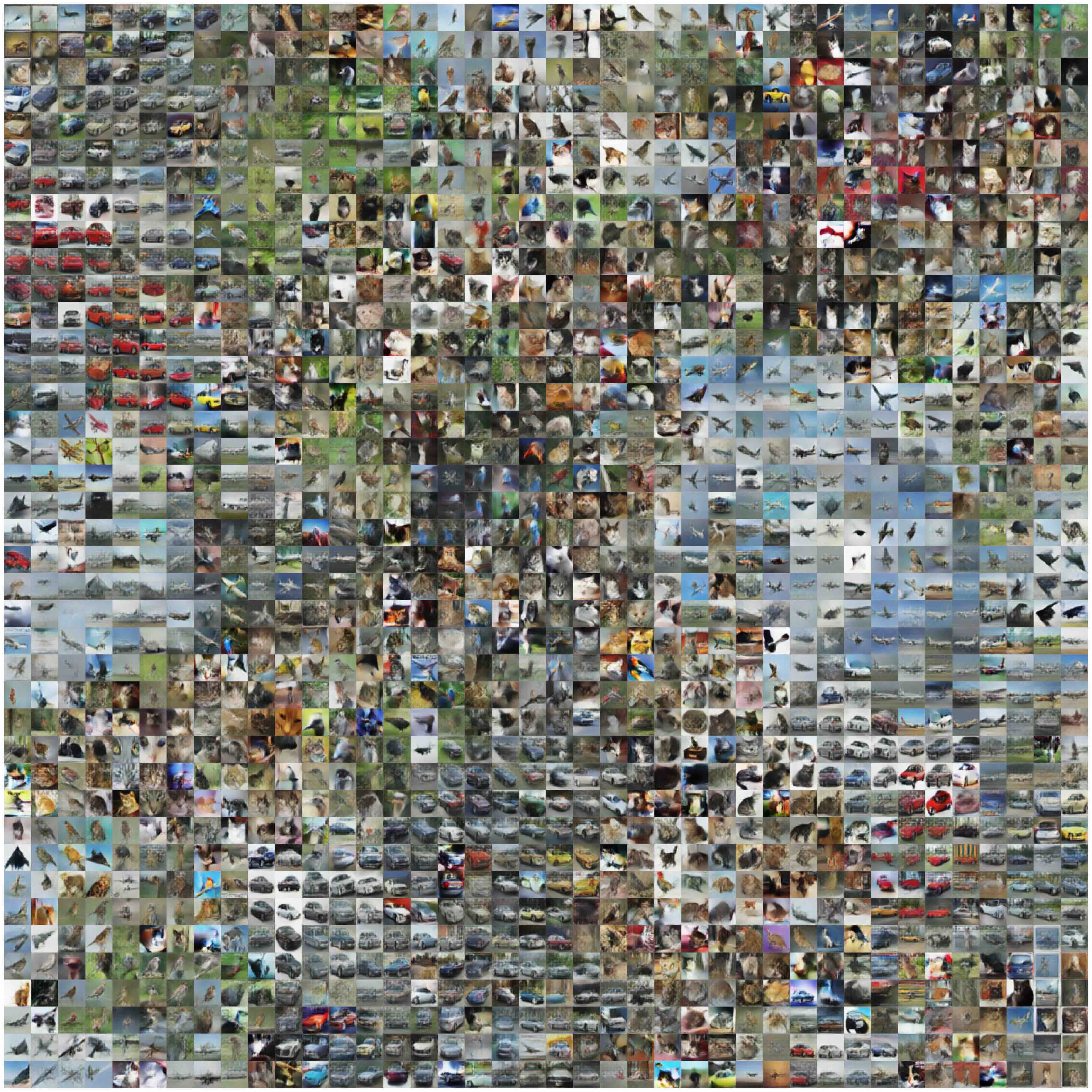}
    \includegraphics[width=0.32\textwidth]{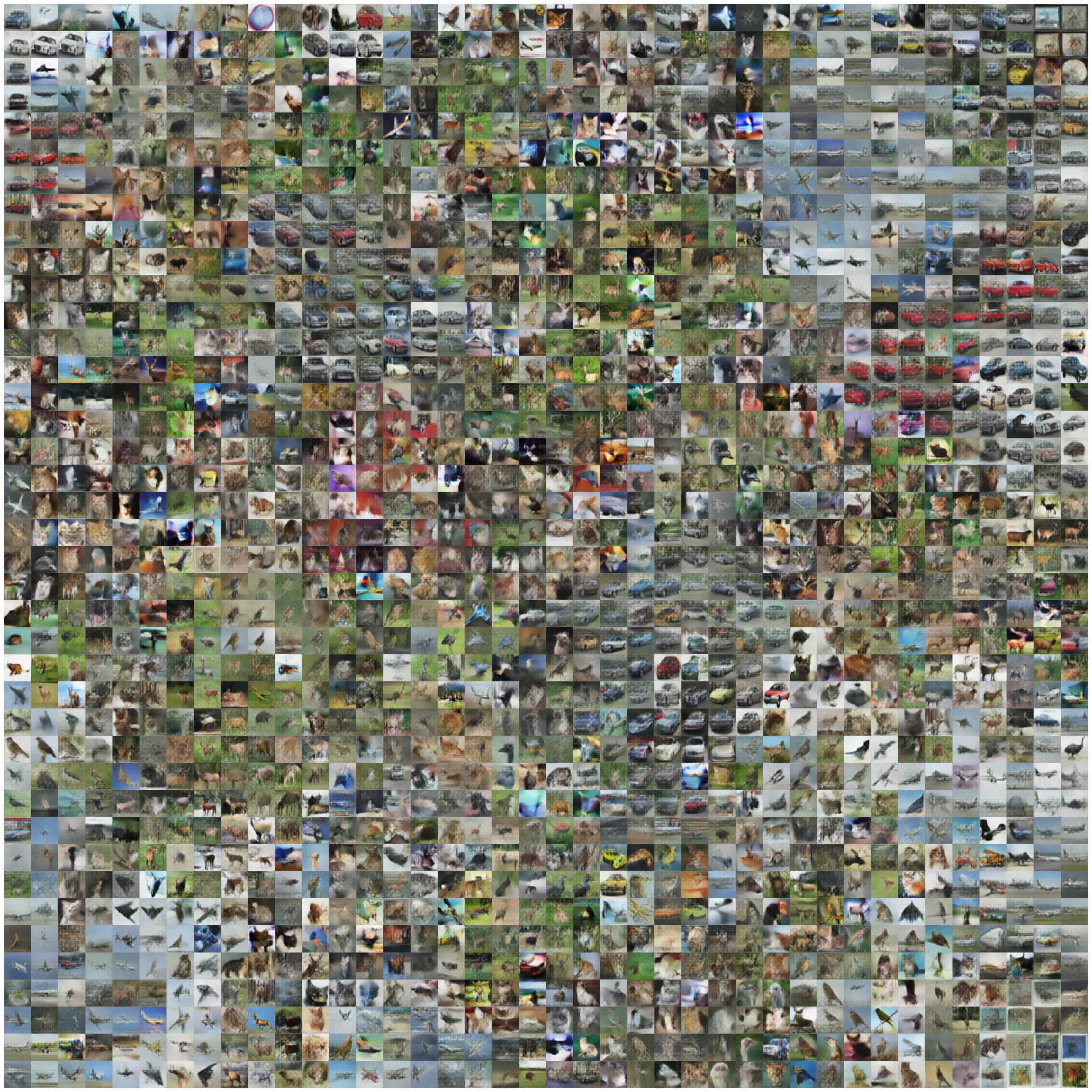}
    \includegraphics[width=0.32\textwidth]{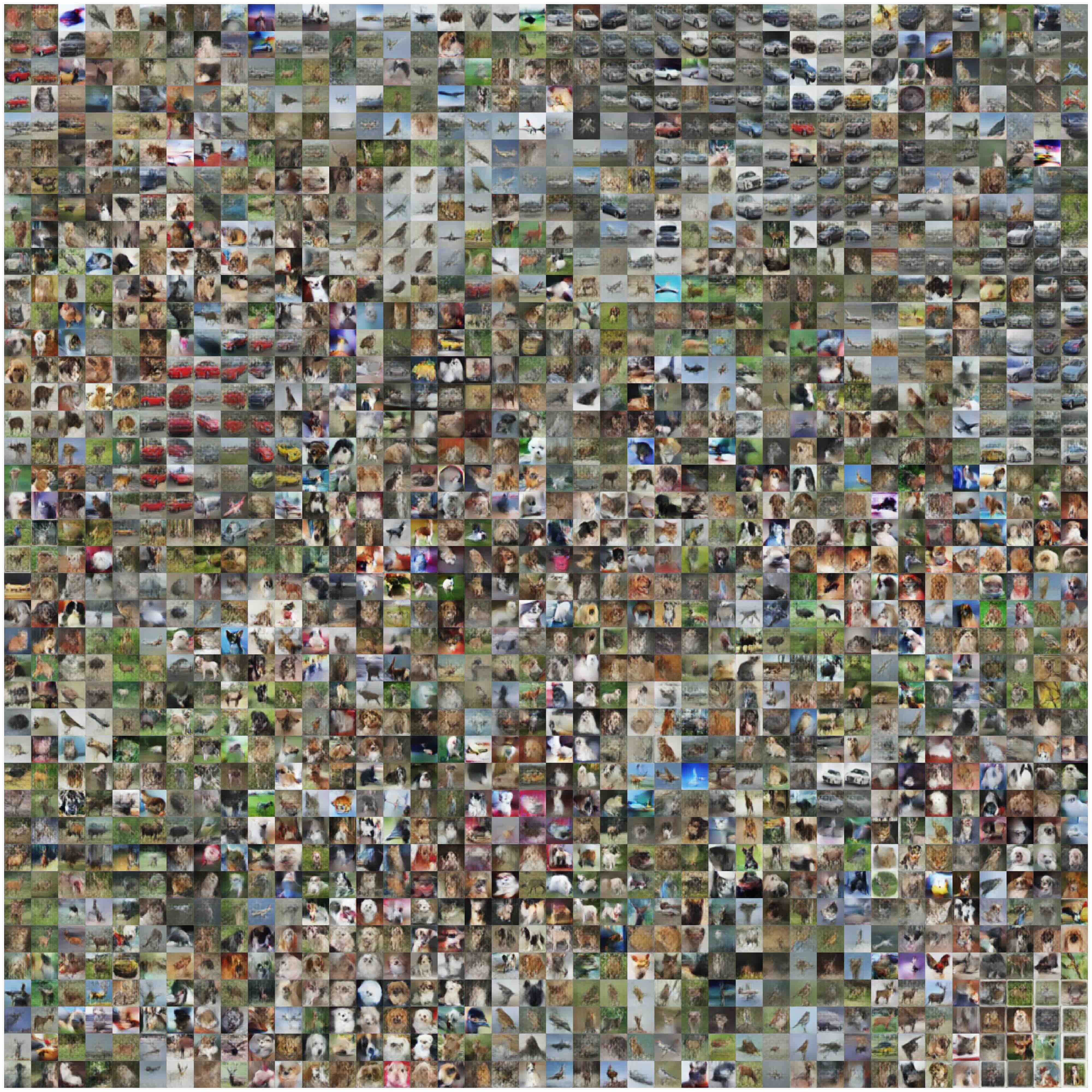} \\[1mm]

    \includegraphics[width=0.32\textwidth]{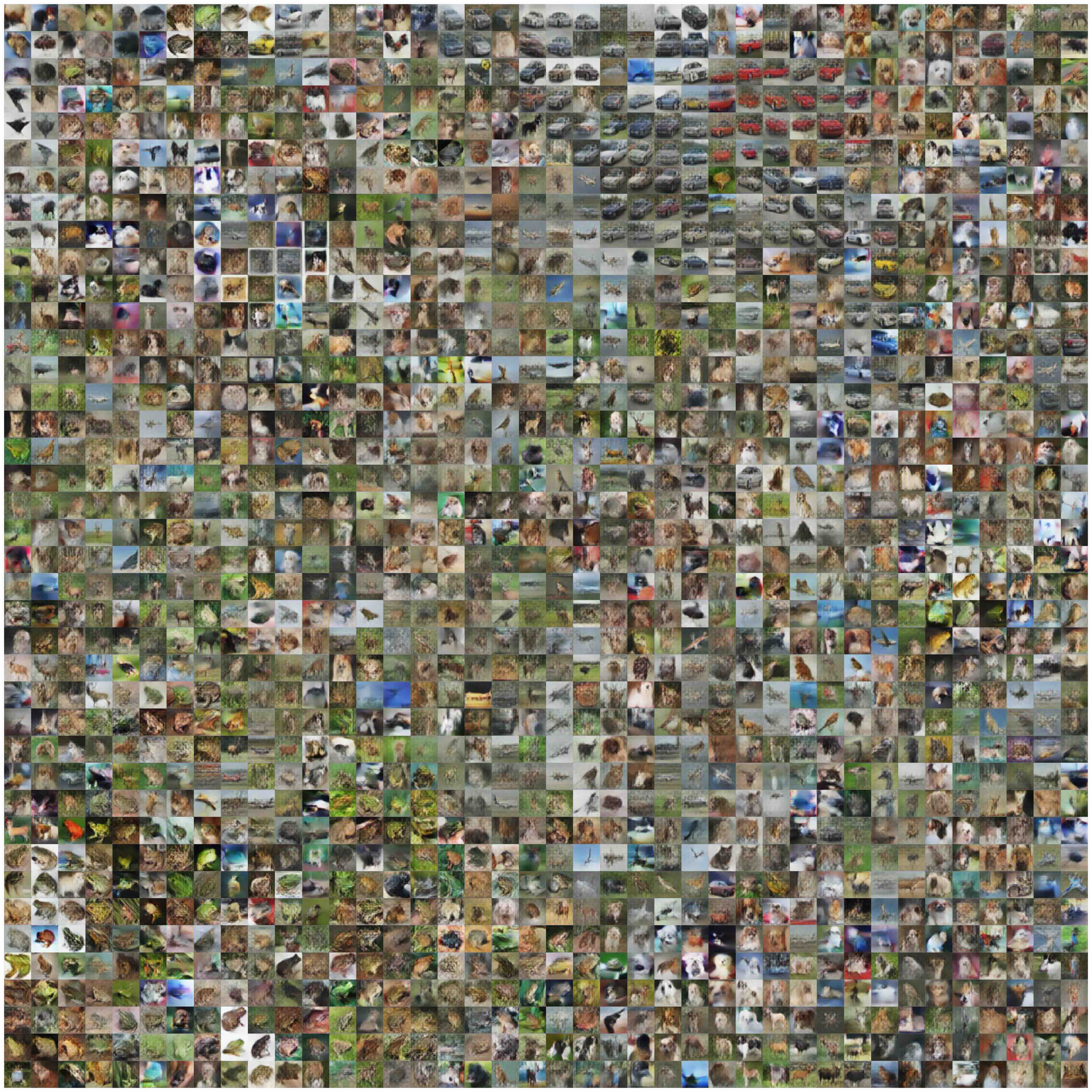}
    \includegraphics[width=0.32\textwidth]{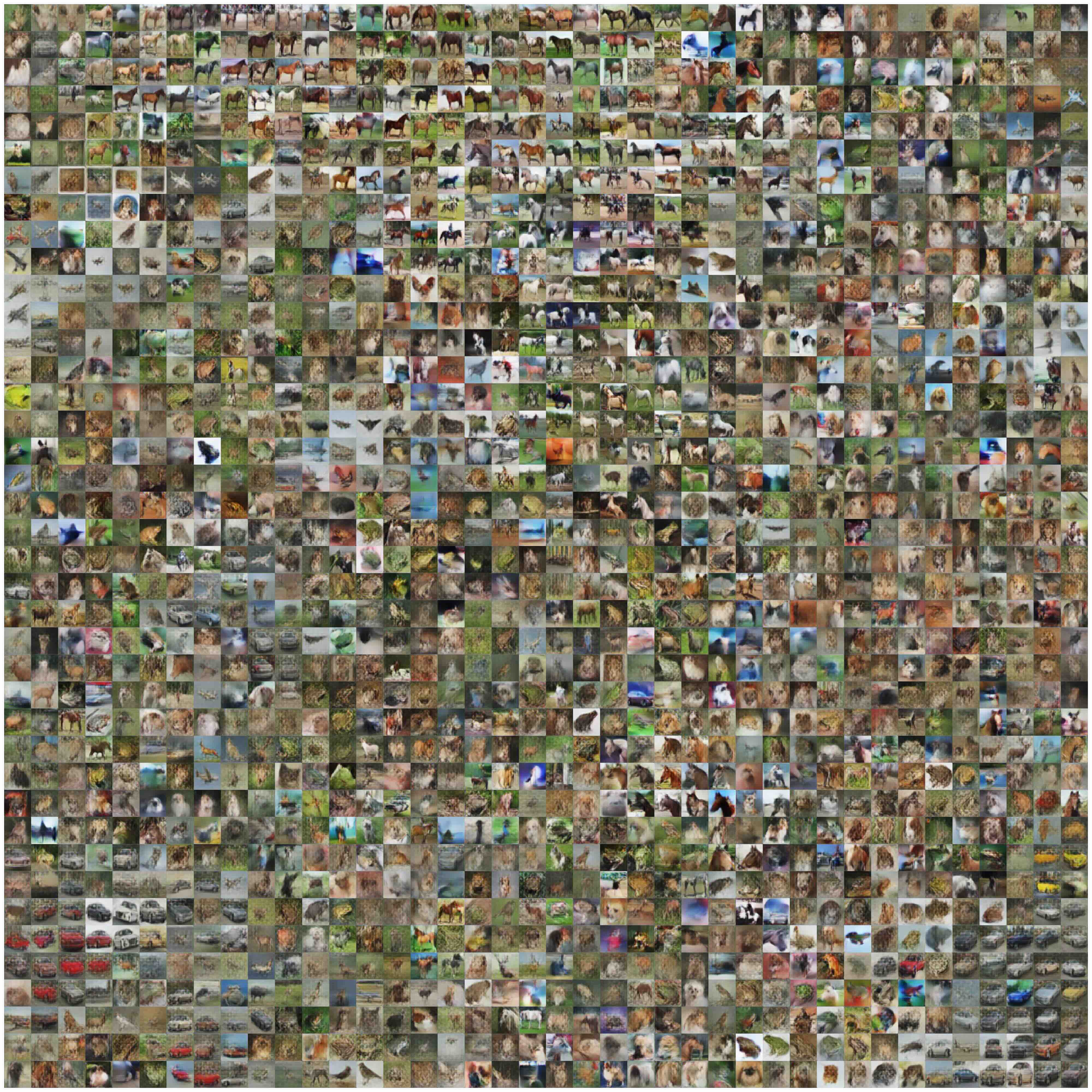}
    \includegraphics[width=0.32\textwidth]{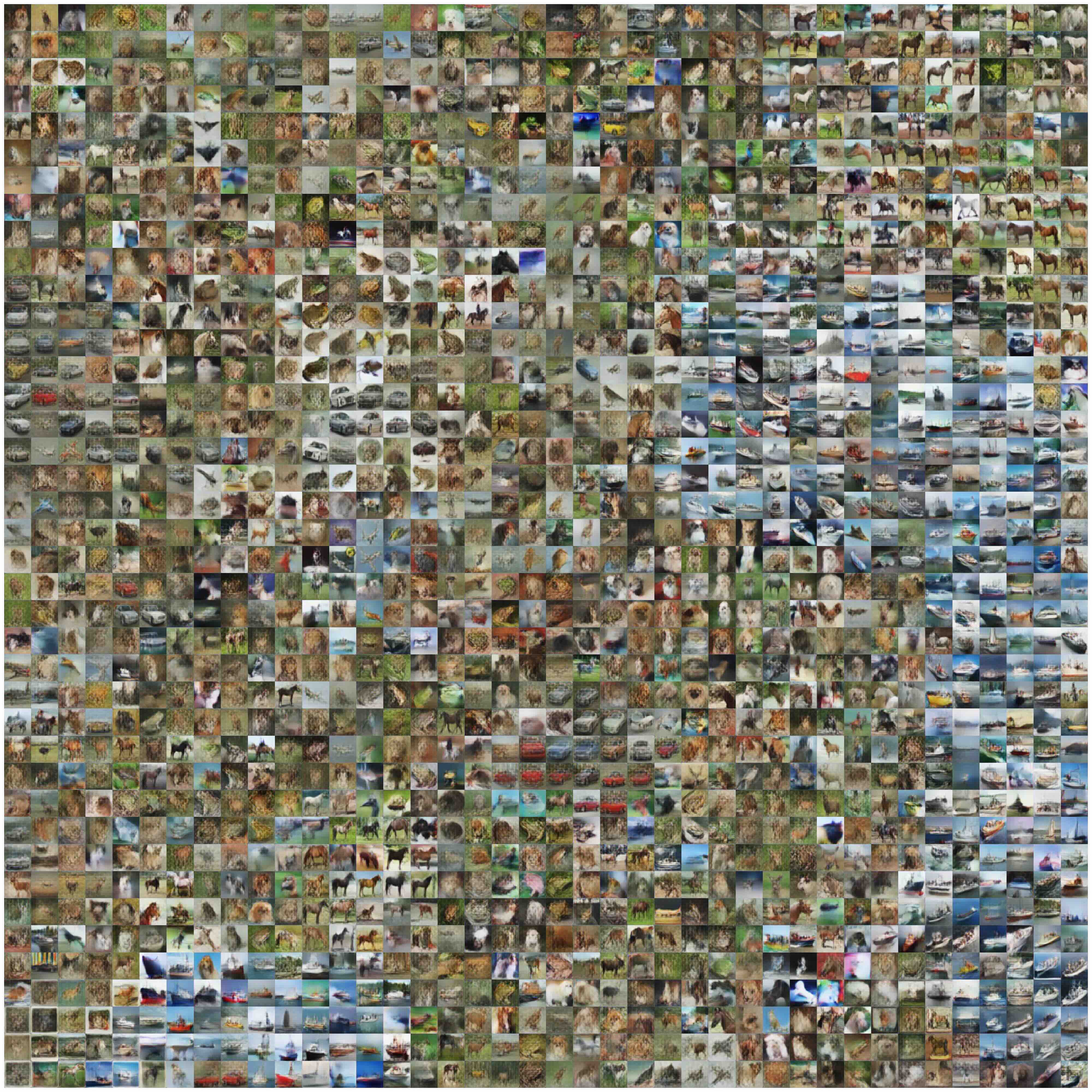} \\[1mm]

    \includegraphics[width=0.32\textwidth]{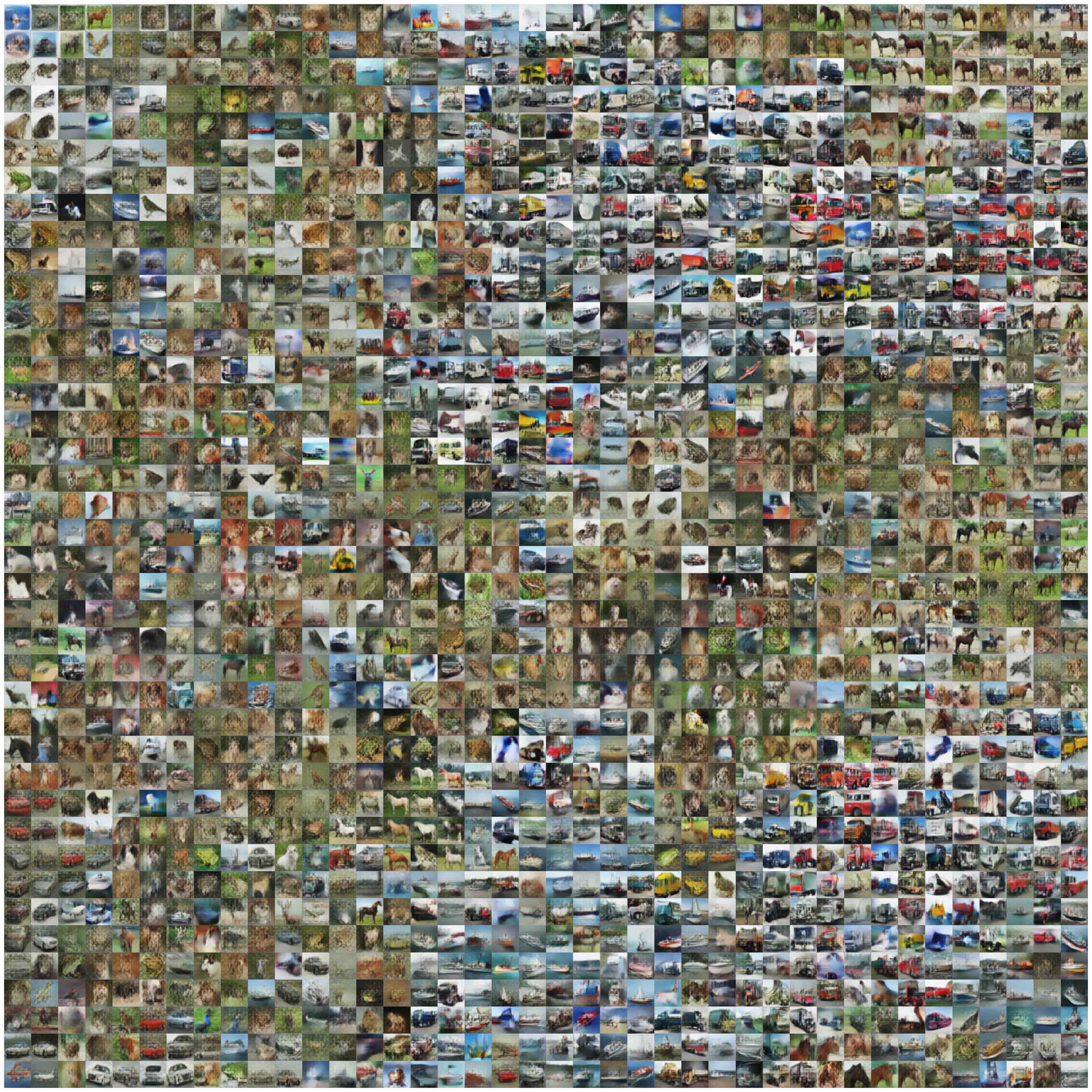}

    \caption{Visualization of decoded SOM unit vectors after each task on CIFAR-10 (40x40 SOM).}
    \label{fig:appendix_cifar_10_gen}
\end{figure*}

\begin{figure*}[!t]
    \centering
    \includegraphics[width=0.32\textwidth]{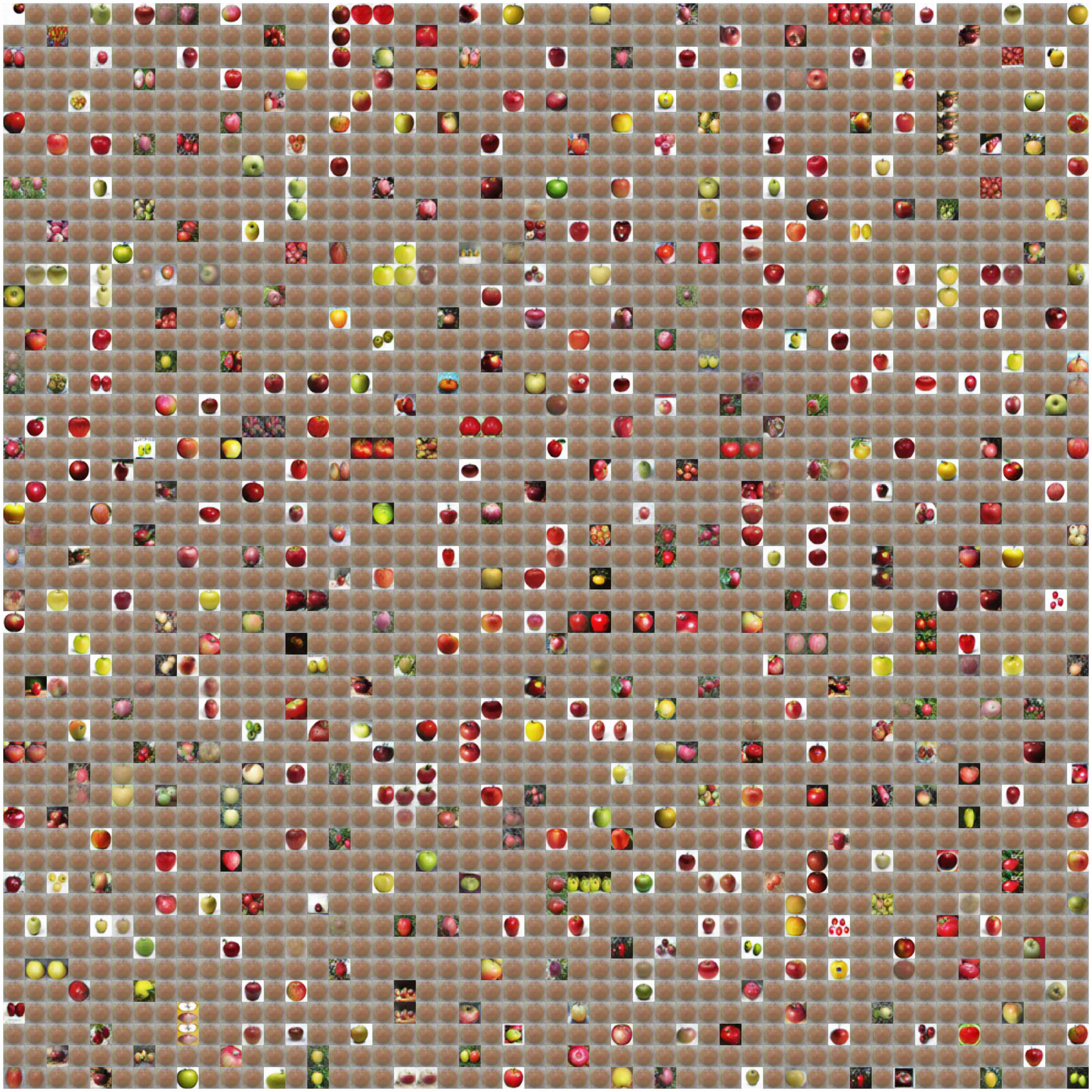}
    \includegraphics[width=0.32\textwidth]{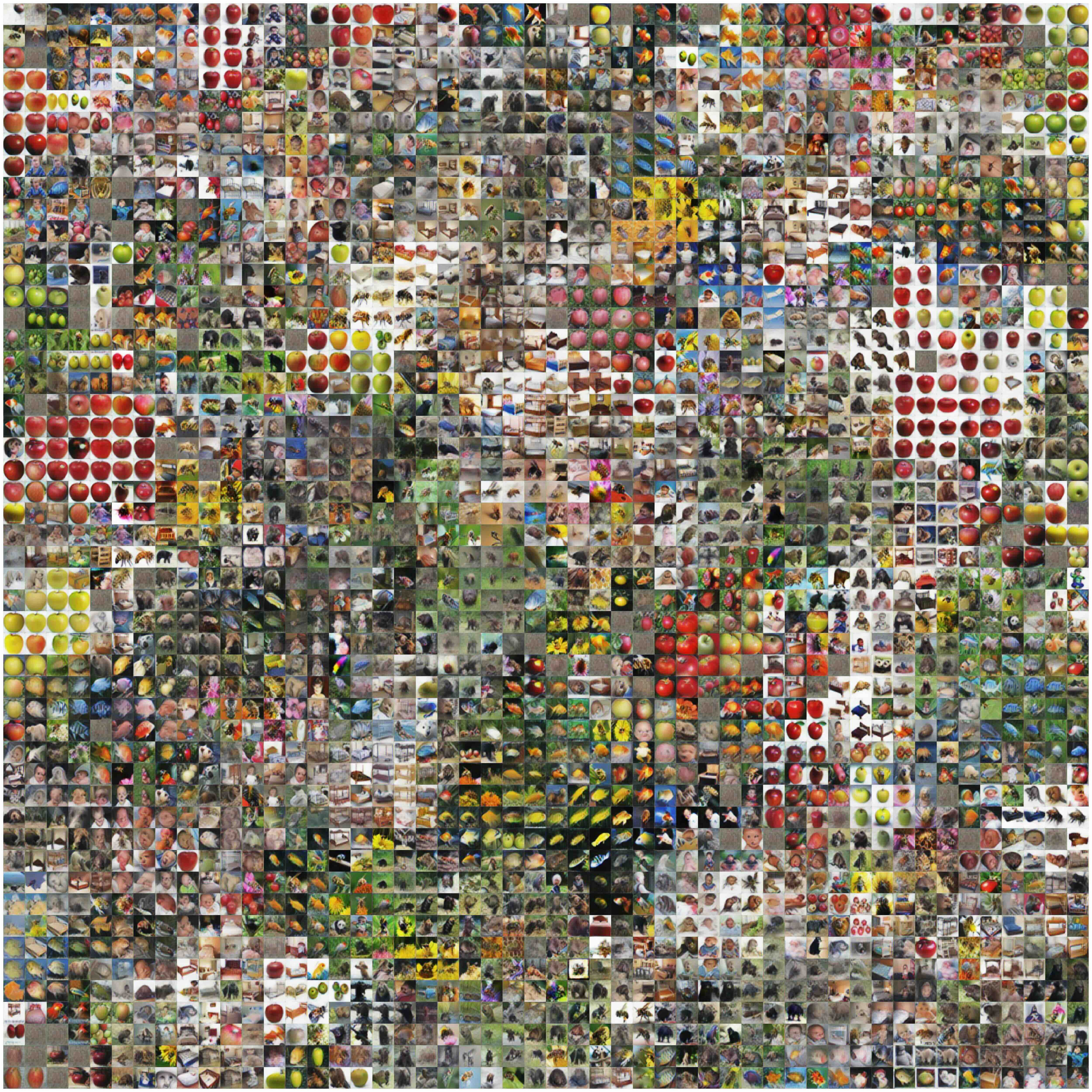}
    \includegraphics[width=0.32\textwidth]{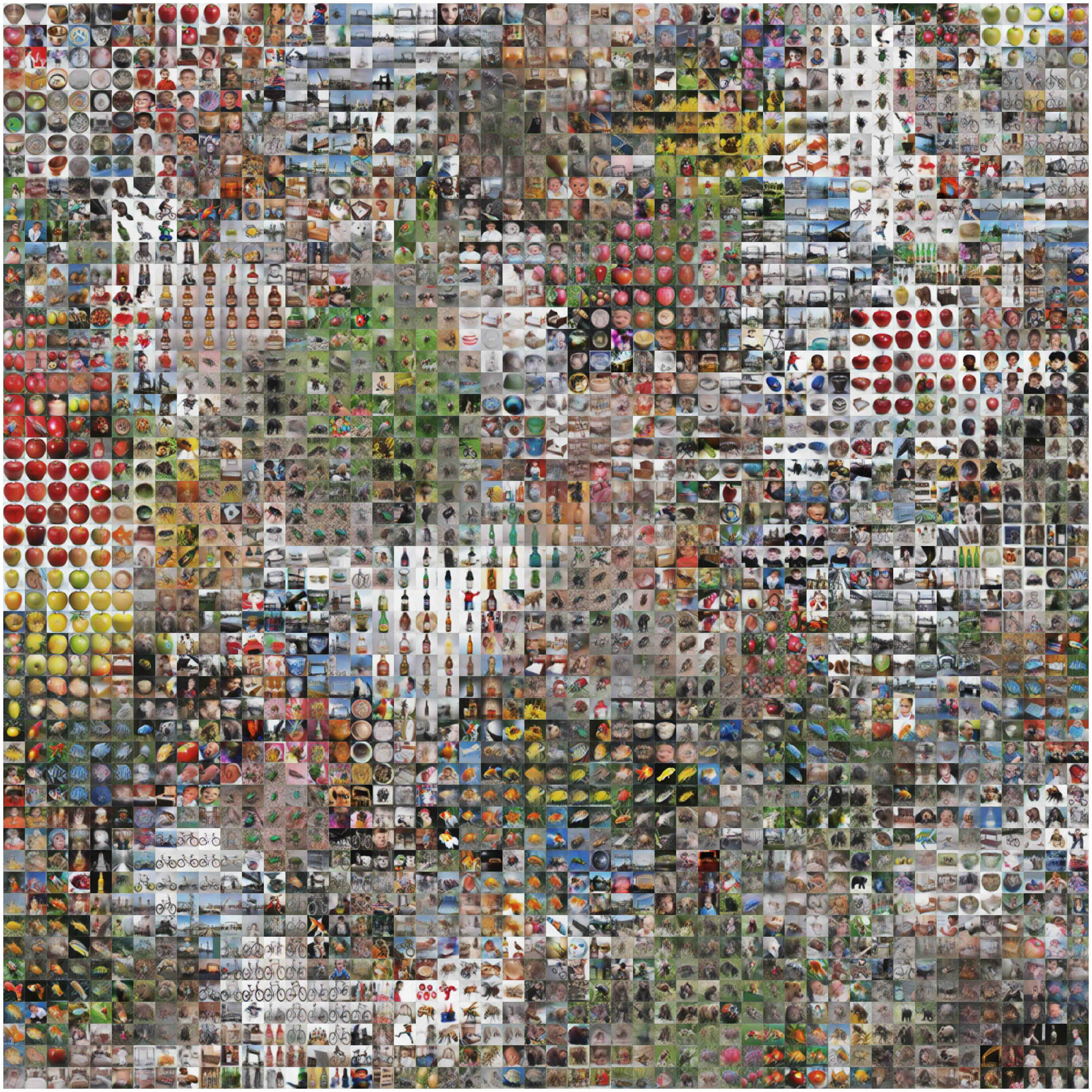} \\[1mm]

    \includegraphics[width=0.32\textwidth]{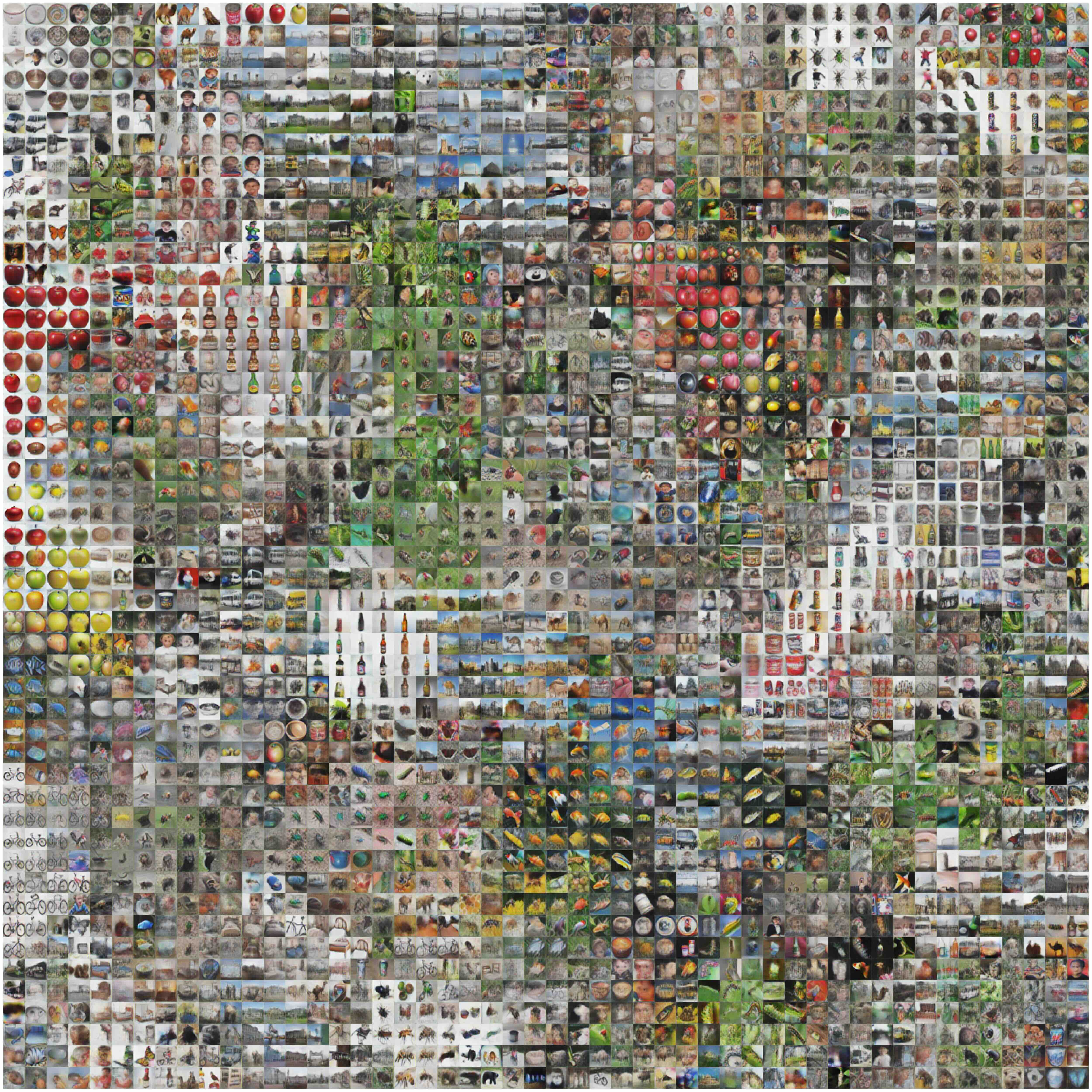}
    \includegraphics[width=0.32\textwidth]{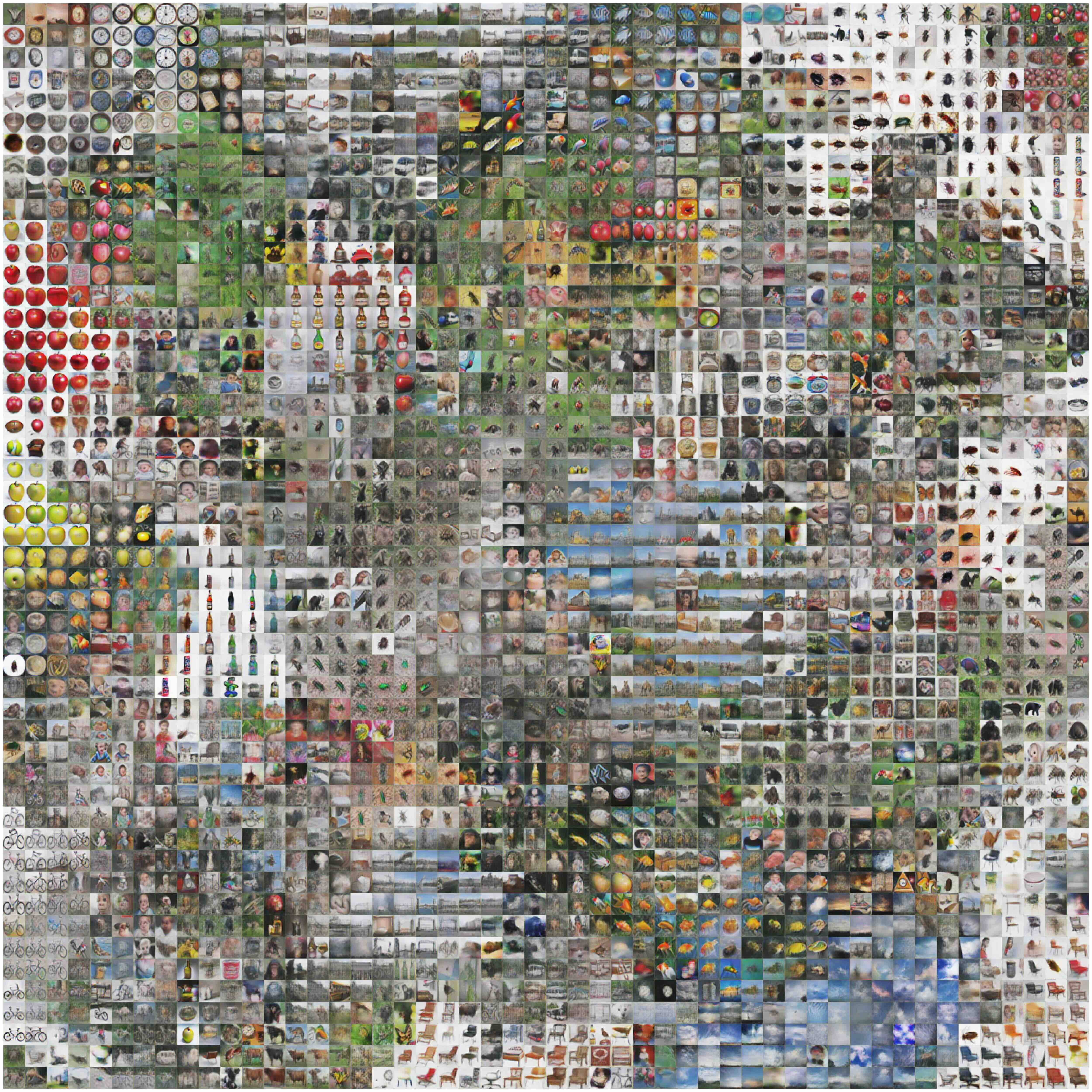}
    \includegraphics[width=0.32\textwidth]{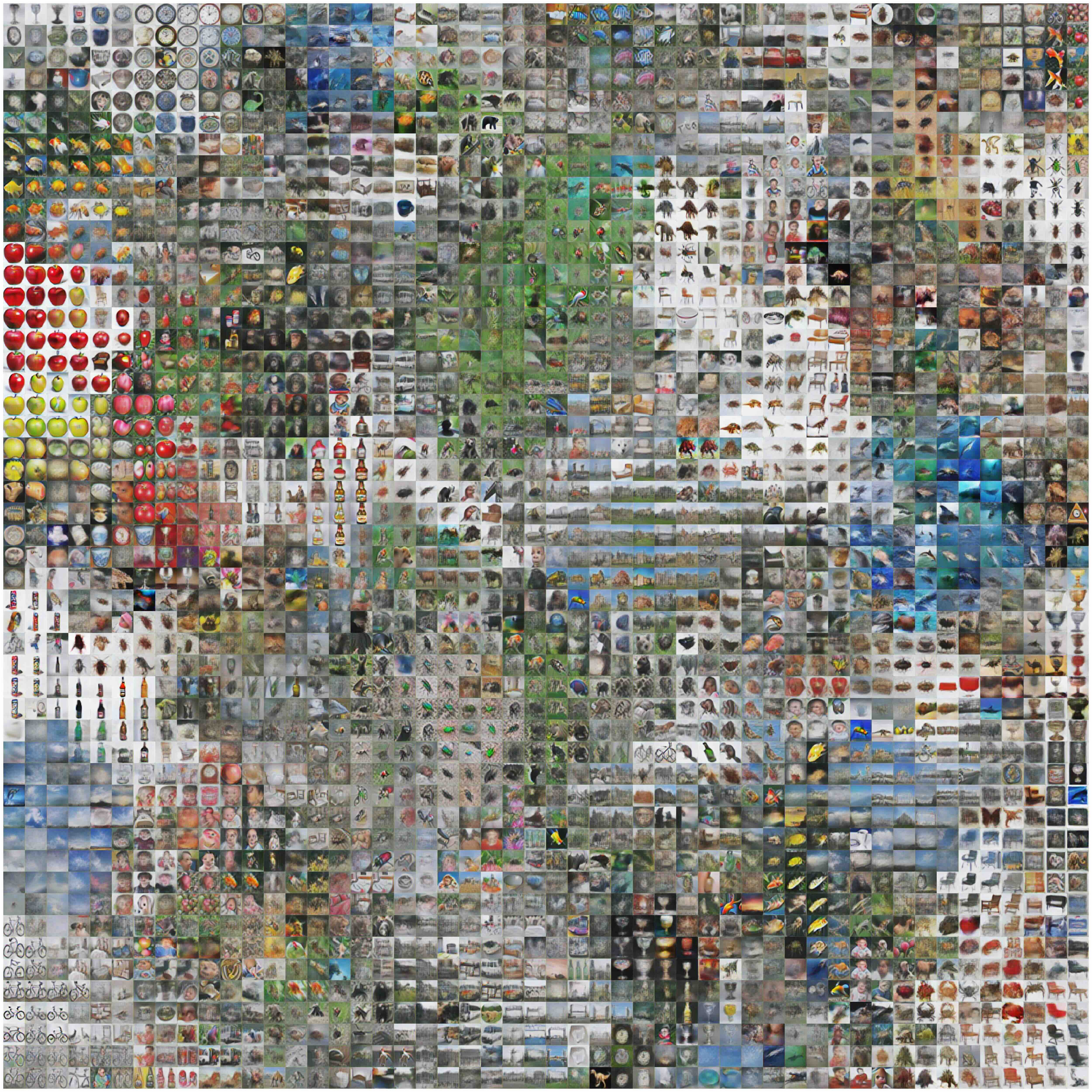} \\[1mm]

    \includegraphics[width=0.32\textwidth]{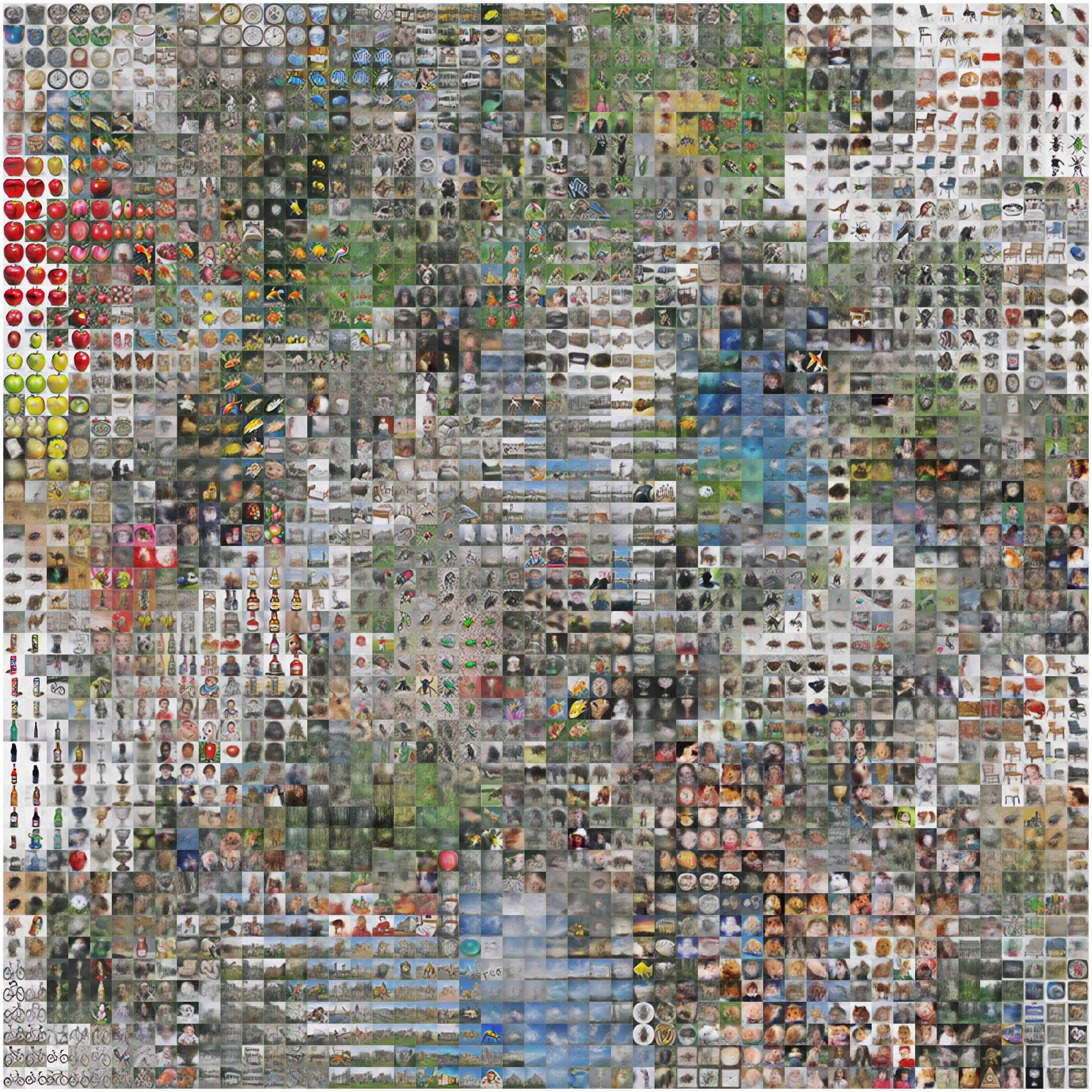}
    \includegraphics[width=0.32\textwidth]{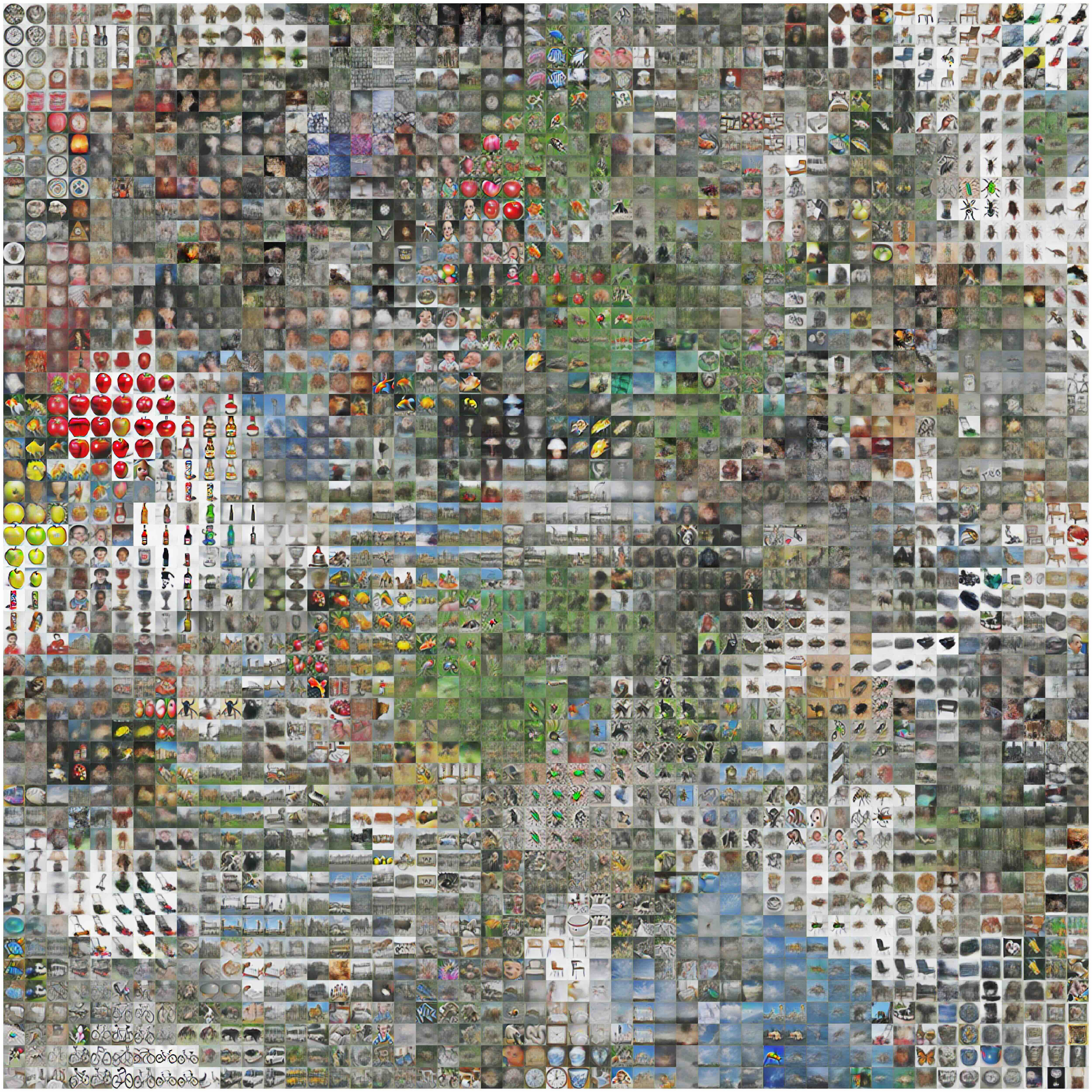}
    \includegraphics[width=0.32\textwidth]{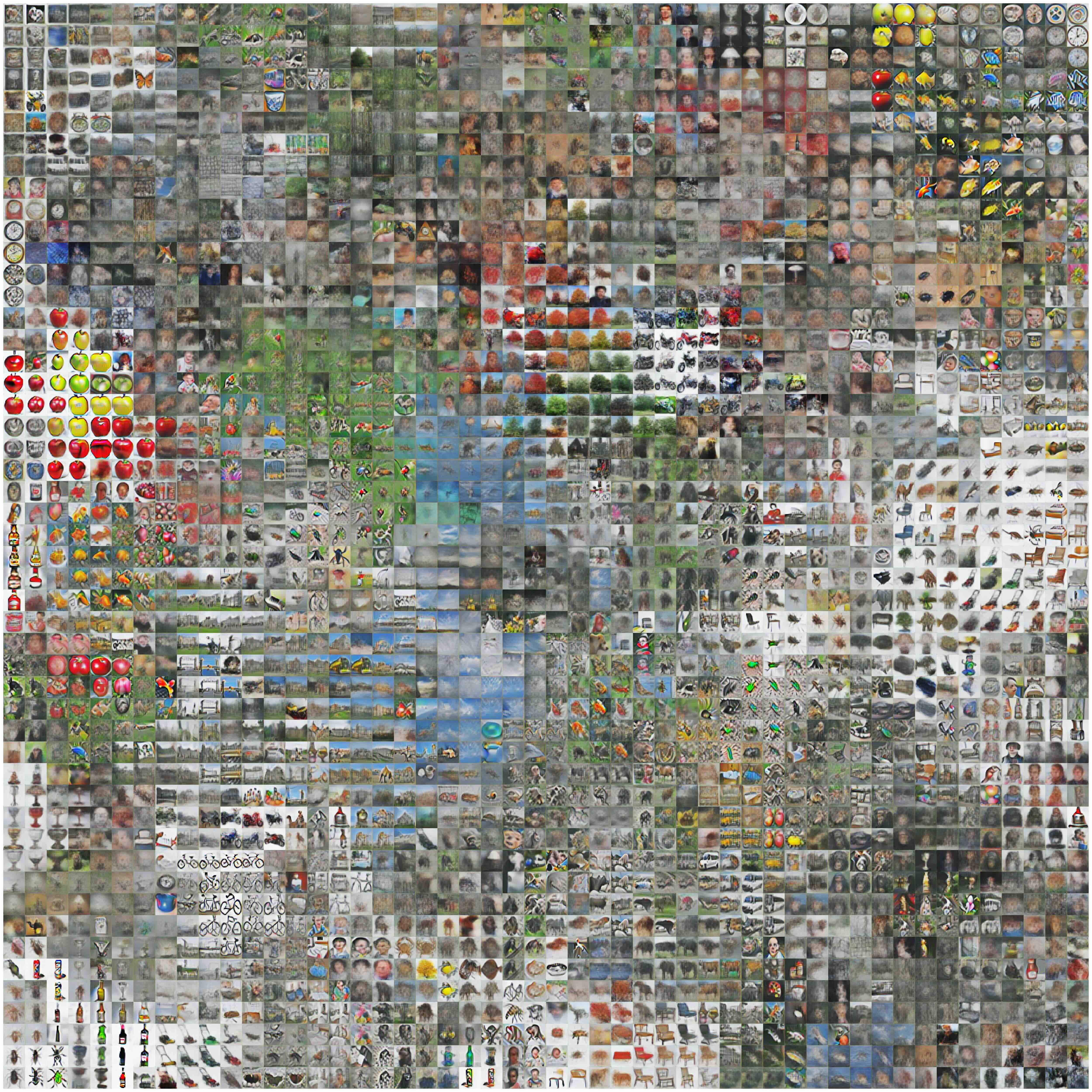} \\[1mm]

    \includegraphics[width=0.32\textwidth]{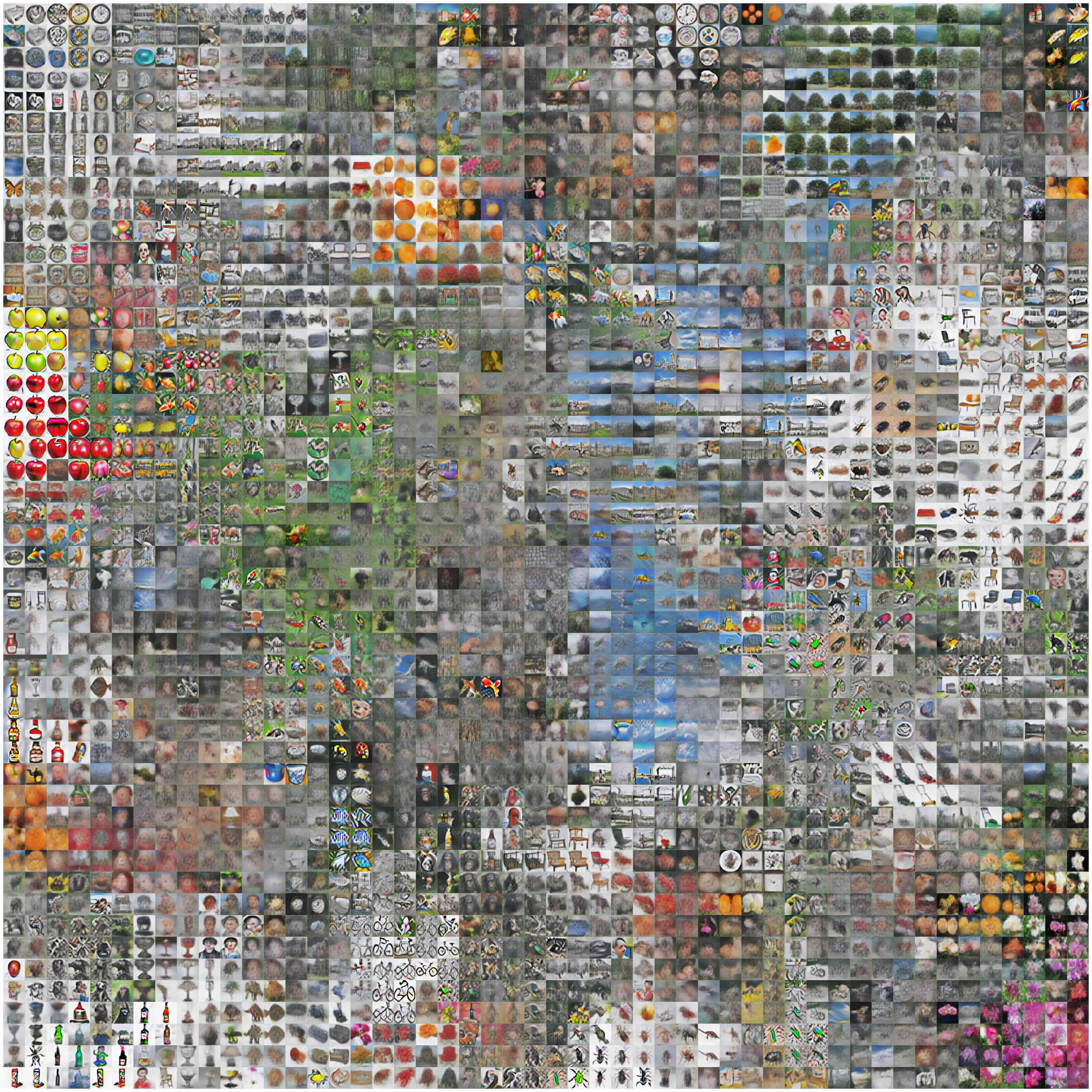}
    \includegraphics[width=0.32\textwidth]{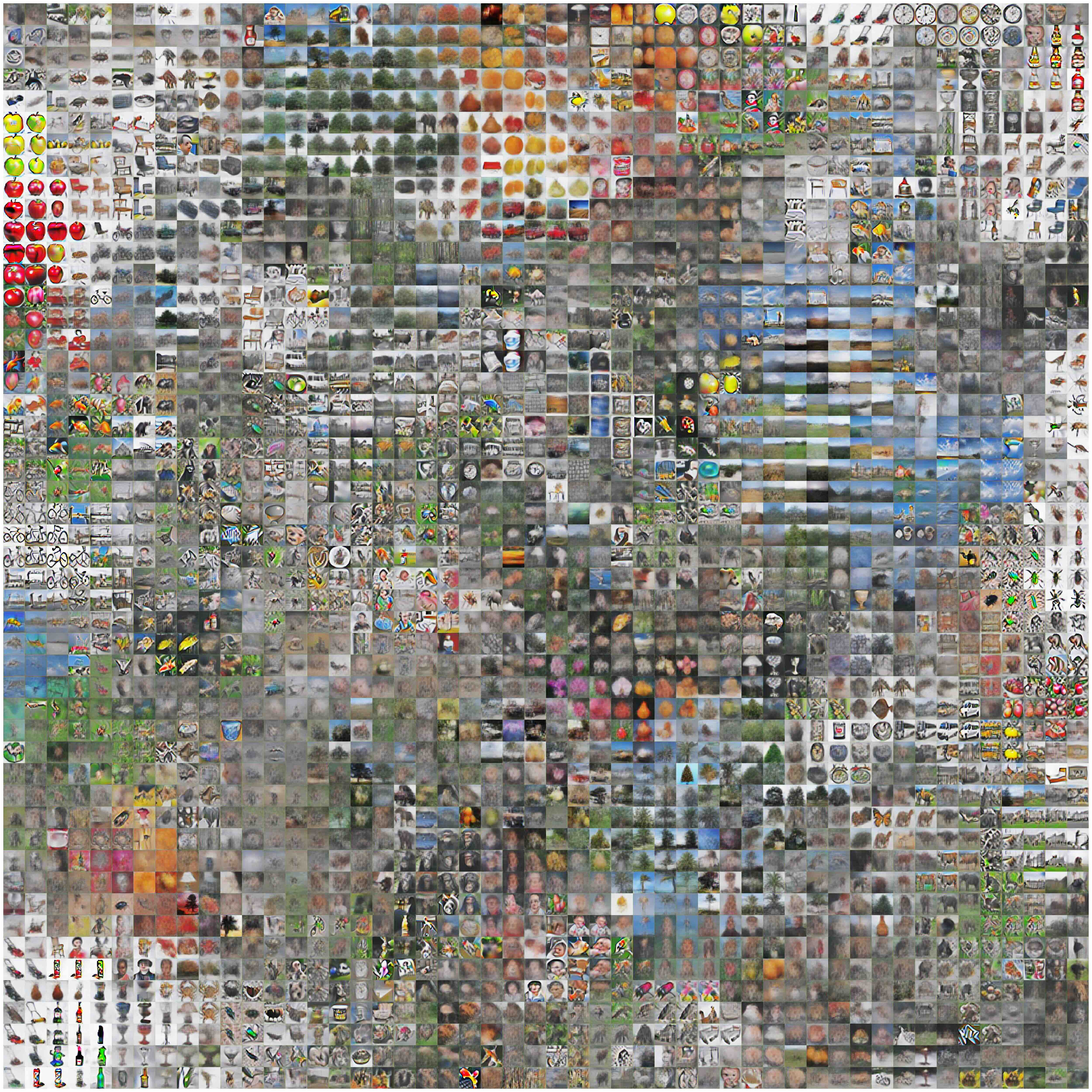}

    \caption{Visualization of decoded SOM unit vectors after every 10 tasks on CIFAR-100 (50x50 SOM).}
    \label{fig:appendix_cifar_100_gen}
\end{figure*}

Figures~\ref{fig:appendix_mnist_gen} and~\ref{fig:appendix_fmnist_gen} show a visualization of the unit vectors after each task for the SOMs generated for MNIST and Fashion-MNIST, respectively.  Each image in the grid represents the SOM BMU unit vector after observing class 0, 1, and up to class 9 sequentially in the one-class-at-a-time learning setup. The 784 dimension unit vectors are converted back to 28x28 tensors of the original image sizes, and then scaled back to pixel space. Similarly, Figures~\ref{fig:appendix_cifar_10_gen} and~\ref{fig:appendix_cifar_100_gen} show the SOM BMU unit vectors after being decoded back to images by the VAE. CIFAR-10 shows the SOM state after each class, and CIFAR-100 shows the SOM after every 10 classes. These images show the methodology retains regions for each learned class within the SOM.

\section*{Memory Footprint}
The memory required by the SOM grows only with the number of BMUs (i.e., the SOM
grid size) and the latent dimensionality used for its per-unit statistics as seen in Table \ref{tab:memory_footprint}. Each SOM unit stores a running mean and variance, which scale linearly with the latent dimensionality, as well as a symmetric covariance matrix for which only the $d(d+1)/2$ unique entries are retained. As a result, the statistical memory exhibits an overall quadratic $O(d^2)$ dependence on the latent dimensionality. This memory footprint is fixed once the SOM resolution and latent dimensionality are chosen, and remains stable and predictable across tasks, independent of the number of samples observed. 

When a single global model is used (global VAE-SOM or global CLIP-SOM), the total memory therefore increases linearly with SOM size. The proposed BMU-specific variants retain this shared backbone while introducing only lightweight BMU-specific encoder/projection heads and decoder output heads. Consequently, the model memory scales as \(O(P_{\mathrm{shared}} + |\mathcal{B}|P_{\mathrm{head}})\), where \(|\mathcal{B}|\) is the number of active BMUs, \(P_{\mathrm{shared}}\) is the shared backbone size, and \(P_{\mathrm{head}} \ll P_{\mathrm{shared}}\) is the parameter count of a BMU-specific head. This is substantially more memory-efficient than replicating a complete encoder--decoder for every BMU, whose memory scales as \(O(|\mathcal{B}|P_{\mathrm{model}})\).



Importantly, these changes affect only the model component as the SOM itself remains compact, with memory determined solely by the grid size and latent dimensionality. This highlights that the SOM continues to act as a lightweight and scalable structural backbone. Meanwhile, ongoing work is focused on improving the efficiency of models (such as VAE, CLIP) built on top of it.

\begin{table*}[t]
\centering
\footnotesize
\caption{Memory footprint (MB) of SOM-only, VAE-SOM, Per-BMU VAE-SOM, and CLIP-SOM across different SOM sizes and latent dimensions. All values include SOM weights + SOM running statistics (mean, variance, covariance). The replay buffer is not used since each replay is generated at each iteration. VAE parameters are fixed at 12.6M (50.4 MB). CLIP encoder and the decoder trained with CLIP-embeddings contribute 348 MB and 460 MB, respectively; a total of 808 MB per CLIP unit.}
\label{tab:memory_footprint}
\begin{tabular}{ccccc}
\toprule
\makecell{\textbf{Latent}\\\textbf{(Dim)}} &
\textbf{SOM Size} &
\makecell{\textbf{SOM Memory}\\\textbf{(MB)}} &
\makecell{\textbf{Extra Model}\\\textbf{Memory (MB)}} &
\makecell{\textbf{Total}\\\textbf{(MB)}} \\
\midrule
\midrule
\multicolumn{5}{c}{On MNIST and FMNIST (mean-variance method)} \\
\midrule
\multicolumn{5}{c}{\textbf{SOM-only}} \\
\midrule
-- & 10$\times$10 & 0.60 & --   & 0.60 \\
-- & 12$\times$12 & 0.86 & --   & 0.86 \\
-- & 20$\times$20 & 2.39 & --   & 2.39 \\
-- & 30$\times$30 & 5.38 & --   & 5.38 \\
-- & 35$\times$35 & 7.33 & --   & 7.33 \\
\midrule
\midrule
\multicolumn{5}{c}{On CIFAR10 and CIFAR100 (full-covariance)} \\
\midrule
\multicolumn{5}{c}{\textbf{VAE-SOM (Global VAE, 12.6M params = 50.4 MB)}} \\
\midrule
32x2x2 & 25$\times$25 & 20.31 & 50.4 & 70.71 \\
32x2x2 & 30$\times$30 & 28.54 & 50.4 & 78.94 \\
32x2x2 & 35$\times$35 & 38.36 & 50.4 & 88.76 \\
32x2x2 & 40$\times$40 & 50.74 & 50.4 & 101.14 \\
\midrule
\multicolumn{5}{c}{\textbf{CLIP-SOM (Global CLIP)}} \\
\midrule
512 & 25$\times$25 & 316.69 & 808 & 1{,}124.69 \\
512 & 30$\times$30 & 456.03 & 808 & 1{,}264.03 \\
512 & 35$\times$35 & 620.70 & 808 & 1{,}428.70 \\
512 & 40$\times$40 & 810.72 & 808 & 1{,}618.72 \\
\midrule
\multicolumn{5}{c}{\textbf{BMU-Specific VAE-SOM (Shared VAE trunk + BMU-specific heads)}} \\
\midrule
32$\times$2$\times$2 & 25$\times$25 & 20.31 & 138.26 & 158.57 \\
32$\times$2$\times$2 & 30$\times$30 & 28.54 & 176.98 & 205.52 \\
32$\times$2$\times$2 & 35$\times$35 & 38.36 & 222.74 & 261.10 \\
32$\times$2$\times$2 & 40$\times$40 & 50.74 & 275.54 & 326.28 \\

\midrule
\multicolumn{5}{c}{\textbf{BMU-Specific CLIP-SOM (Shared CLIP/decoder trunk + BMU-specific heads)}} \\
\midrule
512 & 25$\times$25 & 316.69 & 1{,}473.91 & 1{,}790.60 \\
512 & 30$\times$30 & 456.03 & 1{,}766.92 & 2{,}222.95 \\
512 & 35$\times$35 & 620.70 & 2{,}113.20 & 2{,}733.90 \\
512 & 40$\times$40 & 810.72 & 2{,}512.76 & 3{,}323.48 \\
\bottomrule
\end{tabular}
\end{table*}

\begin{figure}
\centering
\includegraphics[width=\linewidth]{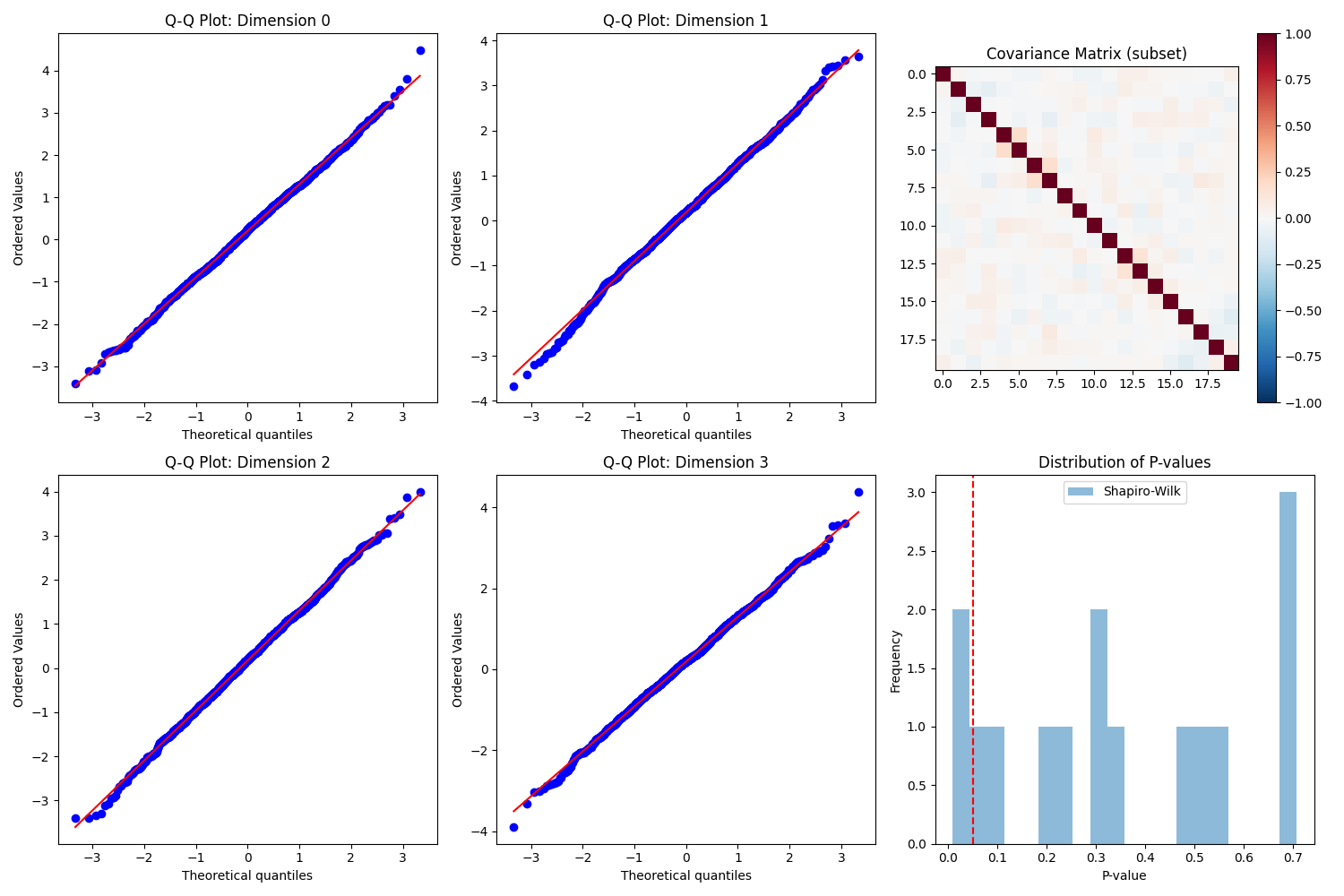}
\caption{
\textbf{Gaussianity analysis of encoder latent representations on CIFAR-10.}
Panels (a–d) show Q–Q plots for four representative standardized latent dimensions, comparing empirical quantiles (y-axis) against a standard normal distribution (x-axis). The close alignment with the diagonal indicates approximate marginal Gaussianity, with only minor deviations in the tails.
Panel (e) visualizes a subset of the covariance matrix computed from standardized latents of a 20 x 20 covariance subset, revealing a strong diagonal structure and weak inter-dimensional correlations.
Panel (f) shows the distribution of Shapiro–Wilk p-values across tested latent dimensions (red dashed line denotes $\alpha=0.05$).
All results are computed using 5{,}000 real and 5{,}000 generated samples.
}
\label{fig:gaussianity}
\end{figure}

\section*{Gaussianity Analysis of Latent Representations and SOM Replay}
\label{app:gaussianity}

We examine whether the encoder’s latent representations exhibit systematic deviations from Gaussianity that could undermine the validity of Gaussian replay at the level of individual SOM units using the Shapiro--Wilk \cite{shapiro} test, complemented by visual diagnostics based on quantile--quantile (Q--Q) plots. All experiments in this section are conducted on CIFAR-10, using 5{,}000 real images and 5{,}000 generated images along with the 35x35 trained SOM model. Latent codes are extracted from the encoder trained from scratch, without external pretraining.
For all statistical tests, latent dimensions are standardized to a zero mean and unit variance prior to analysis.

\paragraph{Q--Q analysis of latent dimensions:}

Figure~\ref{fig:gaussianity} (panels a--d) presents Q--Q plots for four representative latent dimensions. These dimensions correspond to individual coordinates of the flattened latent vector and are selected for visualization only where each plot compares empirical quantiles of standardized latent values to those of a standard normal distribution. Across all dimensions tested, the empirical quantiles closely follow the theoretical diagonal, with only minor deviations observed in the extreme tails, as seen in Figure \ref{fig:gaussianity}.

\paragraph{Correlation structure of standardized latents:}

Figure~\ref{fig:gaussianity} (panel e) visualizes a $20 \times 20$ subset of the covariance matrix computed from standardized latent representations. The matrix is strongly diagonally dominant, indicating that most latent dimensions are only weakly correlated. While these correlations are generally small, we retain the full covariance structure during replay to capture any remaining dependencies and to avoid assumptions of independence between latent dimensions.

\paragraph{Shapiro--Wilk test results:}

Figure~\ref{fig:gaussianity} (panel f) reports the distribution of Shapiro--Wilk \cite{shapiro} p-values computed independently for 15 latent dimensions using globally pooled samples. While some dimensions fall below the conventional $\alpha = 0.05$ threshold, a substantial fraction do not reject Gaussianity.The Shapiro--Wilk test is known to be highly sensitive in large-sample regimes, where even mild tail deviations can lead to statistical rejection. Accordingly, these results are interpreted jointly with the Q--Q diagnostics, which consistently indicate approximate normality of marginal latent distributions despite occasional low p-values.

Overall, these findings indicate that the learned latent representations are approximately Gaussian at both the global level and within individual SOM units. This empirical evidence supports the use of per-unit Gaussian modeling for synthetic replay in our continual learning framework.

\end{document}